# Topology Optimization via Machine Learning and Deep Learning: A Review

Seungyeon Shin[1,a], Dongju Shin[1,2,a], and Namwoo Kang[1,2,*]

[1] Cho Chun Shik Graduate School of Mobility,
Korea Advanced Institute of Science and Technology

[2] Narnia Labs

*Corresponding author: nwkang@kaist.ac.kr

[a] Contributed equally to this work.

## Abstract

Topology optimization (TO) is a method of deriving an optimal design that satisfies a given load and boundary conditions within a design domain. This method enables effective design without initial design, but has been limited in use due to high computational costs. At the same time, machine learning (ML) methodology including deep learning has made great progress in the 21st century, and accordingly, many studies have been conducted to enable effective and rapid optimization by applying ML to TO. Therefore, this study reviews and analyzes previous research on ML-based TO (MLTO). Two different perspectives of MLTO are used to review studies: (1) TO and (2) ML perspectives. The TO perspective addresses "why" to use ML for TO, while the ML perspective addresses "how" to apply ML to TO. In addition, the limitations of current MLTO research and future research directions are examined.

# 1. Introduction

Topology optimization (TO) is a field of design optimization that determines the optimal material layout under certain constraints on loads and boundaries within a given design space. This method allows the optimal distribution of materials with desired performance to be determined while meeting the design constraints of the structure (Bendsøe, 1989). TO is meaningful in that, compared with conventional optimization approaches, designing is possible without meaningful initial design. Due to these advantages, various TO methodologies have been studied to date (Bendsøe, 1989; Rozvany et al., 1992; Mlejnek, 1992; Allaire et al., 2002; Wang et al., 2003; Xie & Steven, 1993). The following four TO methodologies are described in detail in *Appendix A:* density-based method (i.e., the solid isotropic material with penalization (SIMP) method), evolutionary structural optimization (ESO) method, level-set method (LSM), and moving morphable component (MMC) method.

Recent TO methods aim to solve various industrial applications. Examples include TO for patient-specific osteosynthesis plates (Park et al., 2021), microscale lattice parameter (i.e., the strut diameter) optimization for TO (Cheng et al., 2019), homogenization of 3D TO with microscale lattices (Zhang et al., 2021b), and multiscale TO for additive manufacturing (AM) (Kim et al., 2022). Other notable works for more complex TO problems include a multilevel approach to large-scale TO accounting for linearized buckling criteria (Ferrari & Sigmund, 2020), the localized parametric level-set method applying a B-spline interpolation method (Wu et al., 2020), the systematic TO approach for simultaneously designing morphing functionality and actuation in three-dimensional wing structures (Jensen et al., 2021), the parametrized level-set method combined with the MMA algorithm to solve nonlinear heat conduction problems with regional temperature constraints (Zhuang et al., 2021b), and the parametric level-set method for non-uniform mesh of fluid TO problems (Li et al., 2022).

However, although the aforementioned TO methodologies can produce good conceptual designs, one of the main challenges in performing TO is its high computational cost. The overall cost of the computational scheme is dominated by finite element analysis (FEA), which computes the sensitivity for each iteration of the optimization process. The required FEA time increases as the mesh size increases (e.g., when the mesh size is increased by 125 times, the required time increases by 4,137 times (Liu & Tovar, 2014)). Amid this computational challenge, performing TO for a fine (high-resolution) topological mesh can take a few hours to days (Rade et al., 2020). Furthermore, the 3D TO process requires much higher computational costs with increasing demands in the order SIMP, BESO, and level set method in terms of the number of iterations (Yago et al., 2022). Therefore, various studies have aimed to reduce the computation of solving these analysis equations in TO (Amir et al., 2009; Amir et al., 2010). For instance, by using an approximate approach to solve the nested analysis equation in the minimum compliance TO problem, Amir and Sigmund (2011) reduced the computational cost by one order of magnitude for an FE mesh with 40,500 elements.

Similarly, aiming to improve this computational challenge, various methods have been developed to accelerate TO (e.g., Limkilde et al. (2018) discussed the computational complexity of TO, while Ferrari and Sigmund (2020) conducted large-scale TO with reduced computational cost. Martínez-Frutos et al. (2017) performed efficient computation using GPU, while Borrvall and Petersson (2001) and Aage et al. (2015) attempted to accelerate TO by parallel computing. Despite the aforementioned efforts to reduce TO computing time, the computational costs remain high. This challenge has encouraged several researchers to develop ML-based TO to accelerate TO.

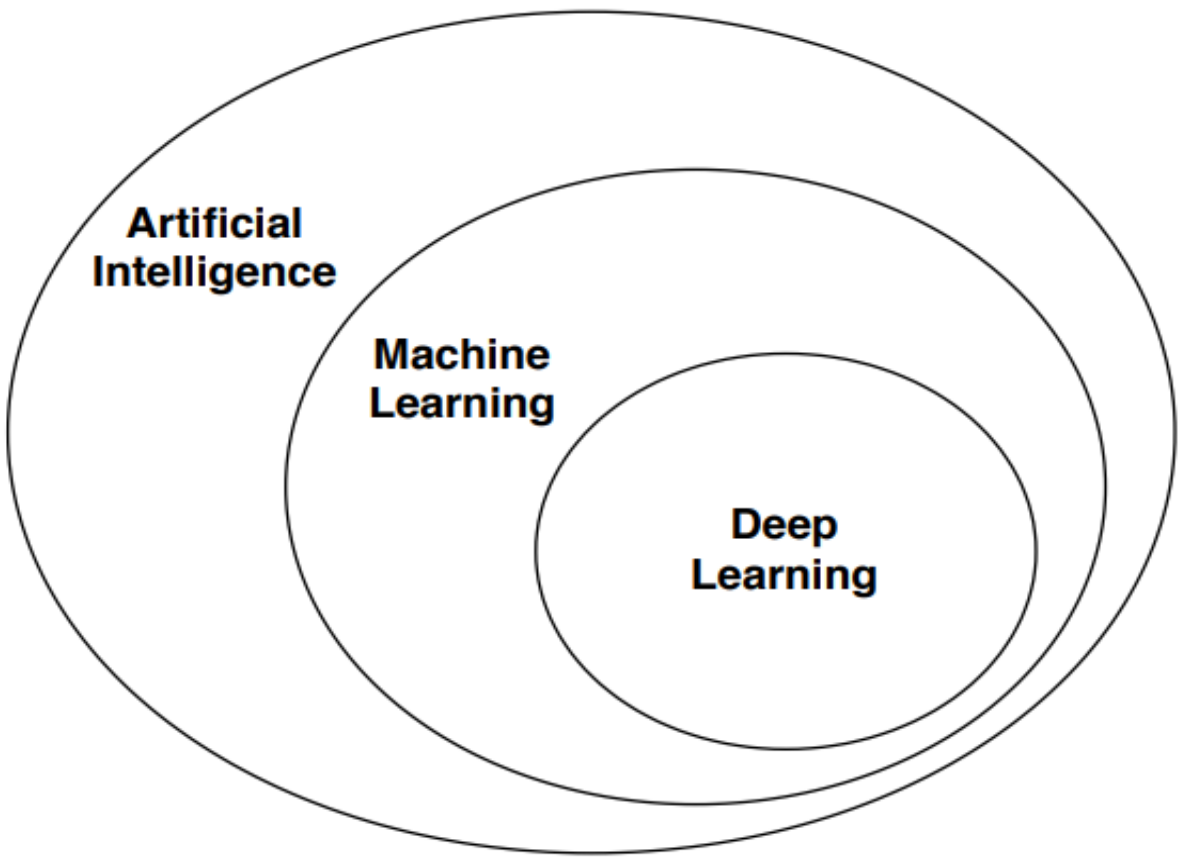

**Figure 1 Overview of AI, ML, and DL (Kaluarachchi et al., 2021)**

Artificial intelligence (AI) includes any technology that allows machines to emulate human behavior, and machine learning (ML) is a subset of AI, aimed at learning meaningful patterns from data by using statistical methods (Figure 1). Deep learning (DL) is a subset of ML, inspired by the neuron structure of the human brain, seeks to improve learning ability through methods of training hierarchical neural network (NN) structures consisting of multiple layers from the data itself (LeCun et al., 2015; Goodfellow et al., 2016).

Data-based TO methodologies complement the problem of conventional TO methodologies with computational cost problems due to design iterations over thousands of times, and many studies have been conducted to improve TO algorithms through AI (Sosnovik & Oseledets, 2019; Oh et al., 2019; Banga et al., 2018, Guo et al., 2018; Cang et al., 2019; Chandrasekhar & Suresh, 2021c). TO methods applying AI can quickly and effectively optimize the initial topology. In this study, ML-based TO (MLTO) is expressed throughout studies that incorporate ML or DL. In particular, we focus more on various methodologies by using DL.

This material intends to review various MLTO studies currently developed and analyze each characteristic. Section 2 and Section 3 present the existing studies from a TO point of view and an ML point of view, respectively, and analyzed the characteristics of the study. Section 2 groups and introduces studies focusing on the purpose, i.e., “why” ML is applied to TO from a TO perspective. Section 3 groups and introduces studies focusing on the ML method, i.e., “how” TO can be converted to ML problems from an ML perspective. In Section 4, the limitations of MLTO studies and the desirable future directions of the field are presented. *Appendixes A and B* introduce the fundamental background of TO and basic DL methodologies required to understand this study. Figure 2 presents various analytical perspectives on MLTO to be described in Sections 2 and 3.

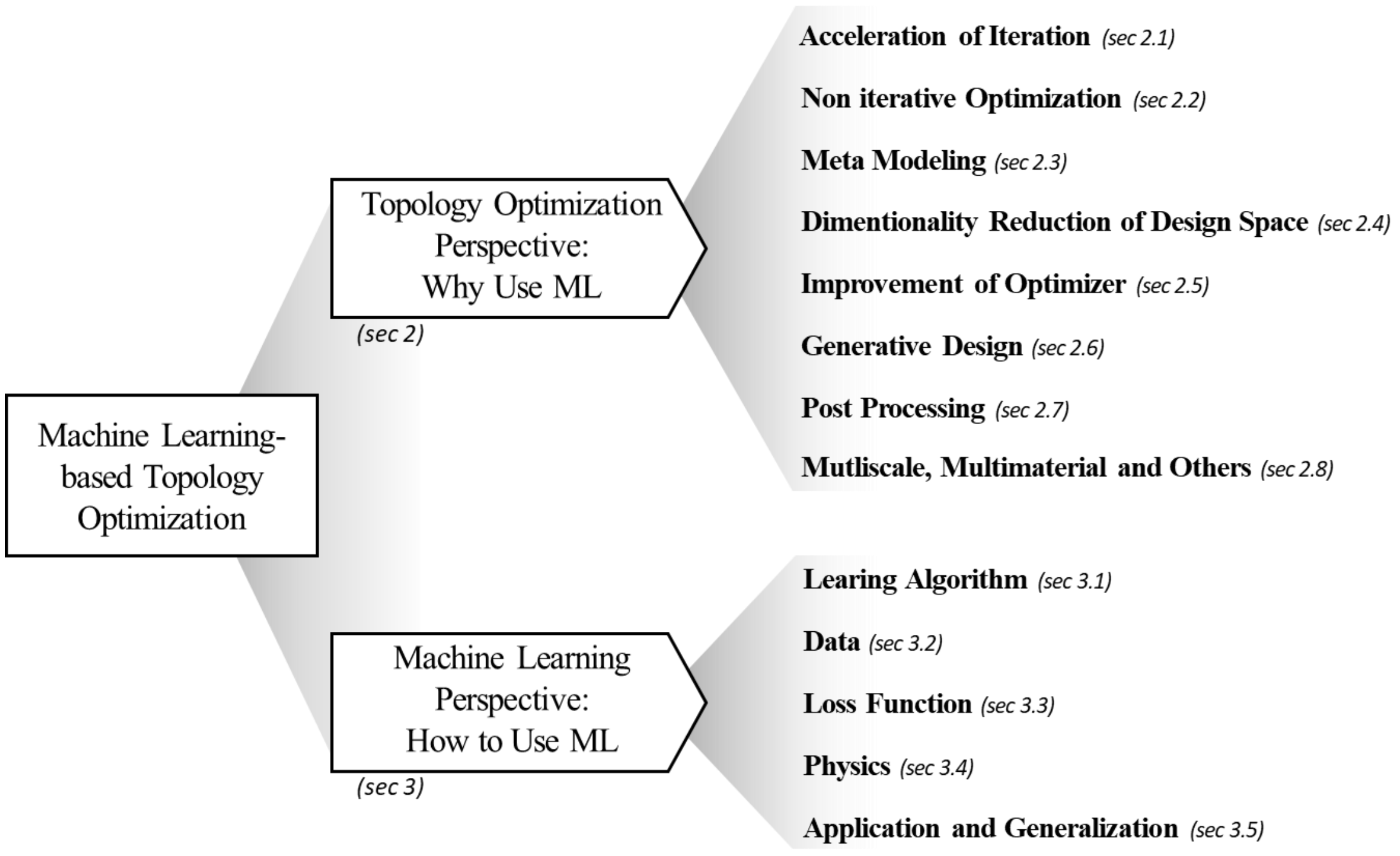

**Figure 2 MLTO Review Framework**

# 2. TO Perspective: Why Use ML

From the perspective of TO, studies that focus on ML/DL to improve the existing TO techniques have seven main purposes: acceleration of iteration, non-iterative optimization, meta-modeling, dimensionality reduction of design space, improvement of optimizer, generative design (design exploration), and postprocessing. Each purpose is shown in Figure 3, and the studies corresponding to each purpose are organized in Table 1.

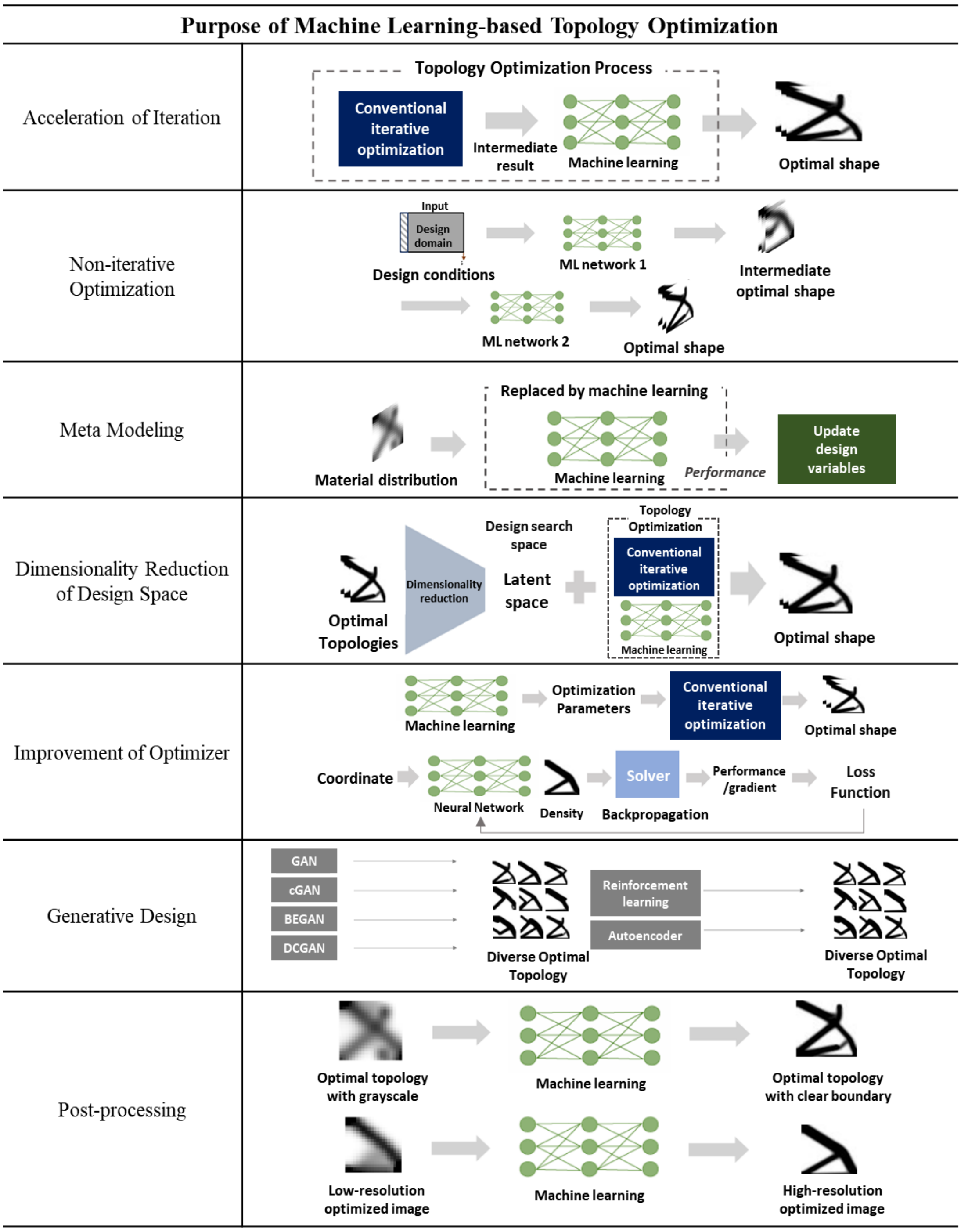

**Figure 3 Visualization of the methodology by MLTO purpose**

**Table 1 Classification of relevant studies by MLTO purpose**

| Purpose of MLTO in TO perspective | Corresponding Research |
|---|---|
| Acceleration of Iteration | Banga et al., 2018; Lin et al., 2018; Sosnovik & Oseledets, 2019; Bi et al., 2020; Kallioras et al., 2020; Kallioras & Lagaros, 2021a |
| Non-Iterative Optimization | Rawat & Shen, 2018; Cang et al., 2019; Li et al., 2019a; Rawat & Shen, 2019a; Rawat & Shen, 2019b; Sharpe & Seepersad, 2019; Yu et al., 2019; Zhang et al., 2019; Abueidda et al., 2020; Almasri et al., 2020; Kollmann et al., 2020; Rade et al., 2020; Keshavarzzadeh et al., 2021; Nie et al., 2021; Lew & Buehler, 2021; Qiu et al., 2021a |
| Meta-Modeling | Patel & Choi, 2012; Aulig & Olhofer, 2014; Xia et al., 2017; Zhou & Saitou, 2017; Doi et al., 2019; Li et al., 2019b; Sasaki & Igarashi, 2019a; Sasaki & Igarashi, 2019b; Takahashi et al., 2019; White et al., 2019; Asanuma et al., 2020; Deng et al., 2020; Keshavarzzadeh et al., 2020; Lee et al., 2020; Chi et al., 2021; Kim et al., 2021; Qian & Ye, 2021; Zheng et al., 2021 |
| Dimensionality Reduction of Design Space | Ulu et al., 2016; Guo et al., 2018; Li et al., 2019c |
| Improvement of Optimizer | Bujny et al., 2018; Hoyer et al., 2019; Lei et al., 2019; Lynch et al., 2019; Deng & To, 2020; Jiang et al., 2020; Chandrasekhar & Suresh, 2021a; Chandrasekhar & Suresh, 2021b; Chandrasekhar & Suresh, 2021c; Halle et al., 2021; Zehnder et al., 2021; Zhang et al., 2021a |
| Generative Design | Oh et al., 2018; Gillhofer et al., 2019; Jiang & Fan, 2019a; Jiang & Fan, 2019b; Jiang et al., 2019; Oh et al., 2019; Shen & Chen, 2019; Wen et al., 2019; Greminger, 2020; Ha et al., 2020; Kallioras & Lagaros, 2020; Malviya, 2020; Sun & Ma, 2020; Wang et al., 2020; Wen et al., 2020; Blanchard-Dionne & Martin, 2021; Kallioras & Lagaros, 2021b; Kudyshev et al., 2021; Li et al., 2021; Sim et al., 2021; Yamasaki et al., 2021; Jang et al., 2022 |
| Postprocessing | Yildiz et al., 2003; Chu et al., 2016; Strömberg, 2019; Karlsson et al., 2020; Napier et al., 2020; Strömberg, 2020; Wang et al., 2021; Xue et al., 2021 |
| Others | Liu et al., 2015; Gaymann & Montomoli, 2019; Hayashi & Ohsaki, 2020; Kumar & Suresh, 2020; Qiu et al., 2021b; Brown et al., 2022 |

## 2.1. Acceleration of Iteration

The optimization process of iterative TO can be divided into two stages: the early stage, in which the design change is significant, and the latter stage, in which the design changes slowly until it converges. Studies applying ML techniques supplement the latter step of the optimization process, which is especially time-consuming. Thus, studies that aim at accelerating iterative TO consist of two stages as follows:

(1) Until the intermediate stage of the iteration, the iterative TO is performed.
(2) In the intermediate stage, the optimal topology can be directly predicted by ML by using the intermediate topology obtained from (1).

This process can accelerate the time-consuming and costly latter stage in the conventional optimization process.

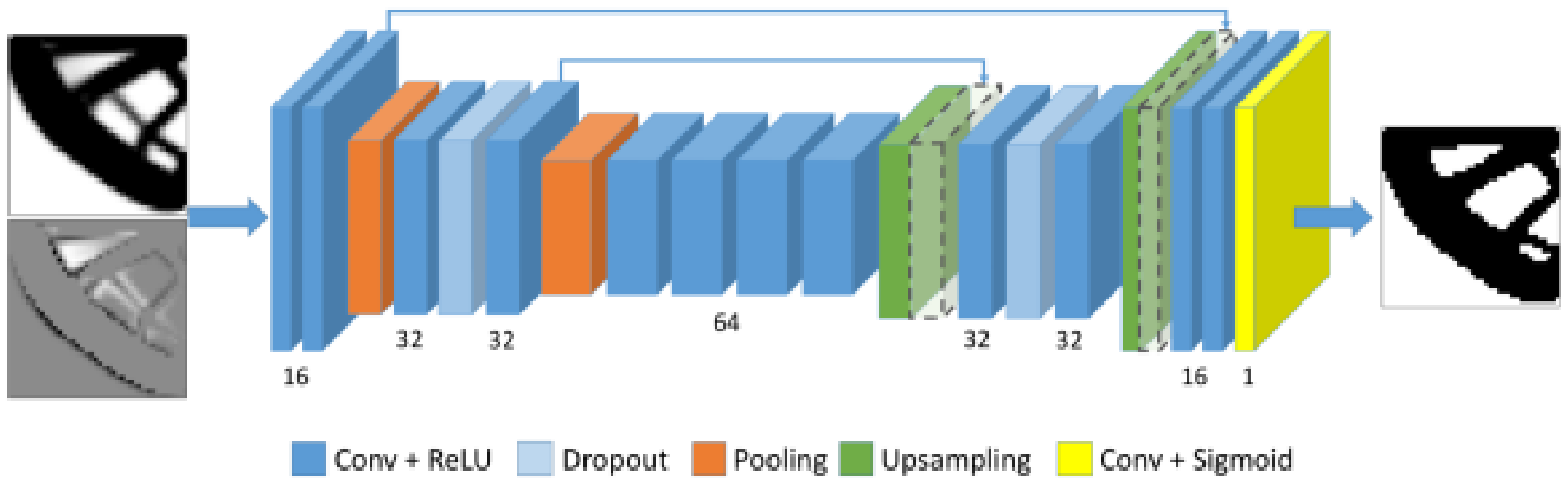

**Figure 4 Architecture of acceleration of iterative optimization from a representative paper (Sosnovik & Oseledets, 2019)**

As a representative study, Sosnovik and Oseledets (2019) proposed an optimization method that predicts the final optimized image through a fully connected network by using the density gradient between iterations obtained through the SIMP method, as shown in Figure 4, to reduce the total time consumption through the method of early stopping the SIMP method. Lin et al. (2018) also predicted the optimal shape of the conductive heat transfer by using two figures obtained by the SIMP method: the distribution of the design parameter and the gradient distribution of the design parameter between iterations. Extending this method to 3D, Banga et al. (2018) proposed the 3D MLTO by using three types of input, the 3D density distribution of voxels at iteration, the gradient of voxel densities between iterations, loading and boundary conditions along the x, y, and z directions, into a 3D convolutional NN (CNN) to predict optimized voxel topology. This study is similar to the work of Sosnovik and Oseledets (2019), except for adding loading and boundary conditions as input and applying them to the 3D domain.

In addition, to significantly reduce the computational cost of fine-discretized meshes, Kallioras et al. (2020) utilized deep belief networks (DBNs) with the SIMP method to perform optimization. The intermediate optimization results from the SIMP method are input into the DBN to predict the near-optimal element density and then the preliminary results were refined using the SIMP method. Similarly, Kallioras and Lagaros (2021a) utilized an iterative re-meshing and convolution process for the computational advantage of the fine-mesh domain. Similarly, Bi et al. (2020) used the gradient predicted by deep NN (DNN) instead of the gradient of every iteration during the optimization to accelerate the TO. They presented a method of predicting a given design and its corresponding gradient based on iteration history data through online updates by training DNN with gradients obtained through local sampling. With this approach, designs can be quickly updated using the predicted gradient with notably few training steps.

## 2.2. Non-Iterative Optimization

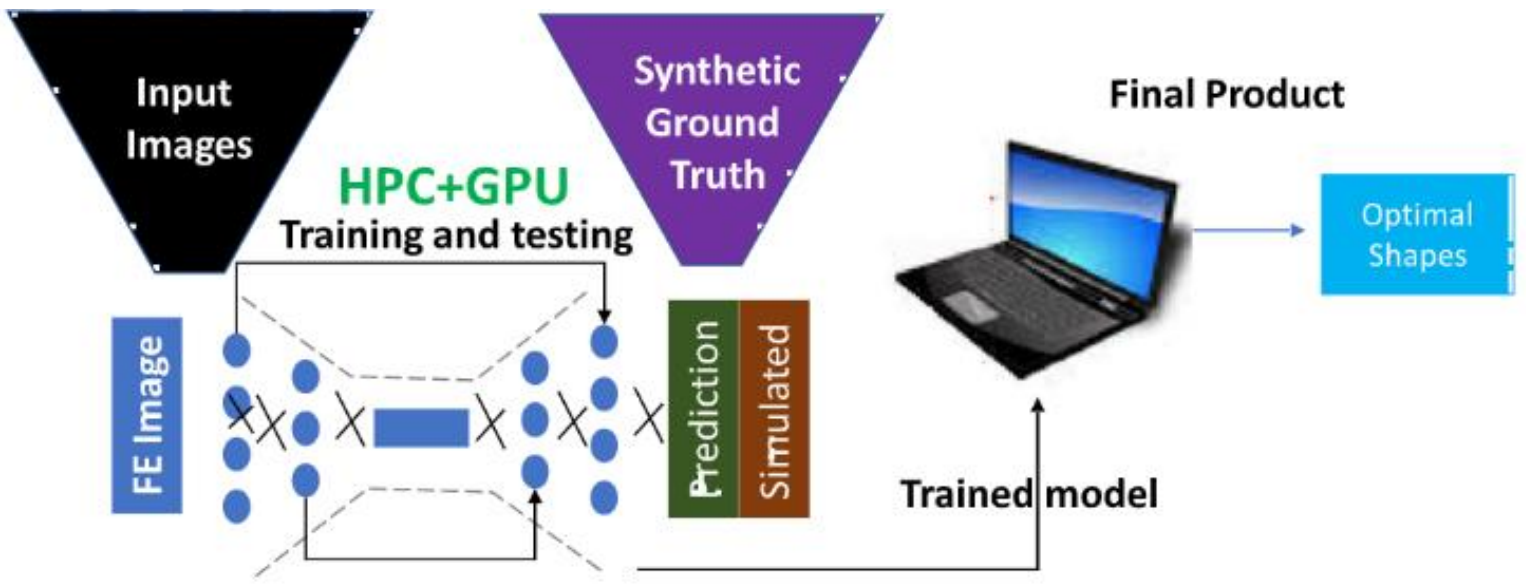

**Figure 5 Single-stage process for non-iterative optimization (Abueidda et al., 2020)**

Studies on non-iterative TO immediately derive an optimal shape that satisfies design conditions, such as loading and boundary conditions, without the iterative process. However, since this method can yield topology that falls short of the objective performance or topology with lower resolution than the optimized shape derived from iterative TO, researchers have proposed frameworks consisting of two or more networks to obtain high-resolution results or optimal designs with satisfying engineering performance. Figure 5 shows an example of the single-stage MLTO, which is the MLTO process through one network, and Figure 6 shows an example of the multistage MLTO process, which is the MLTO process through two or more networks.

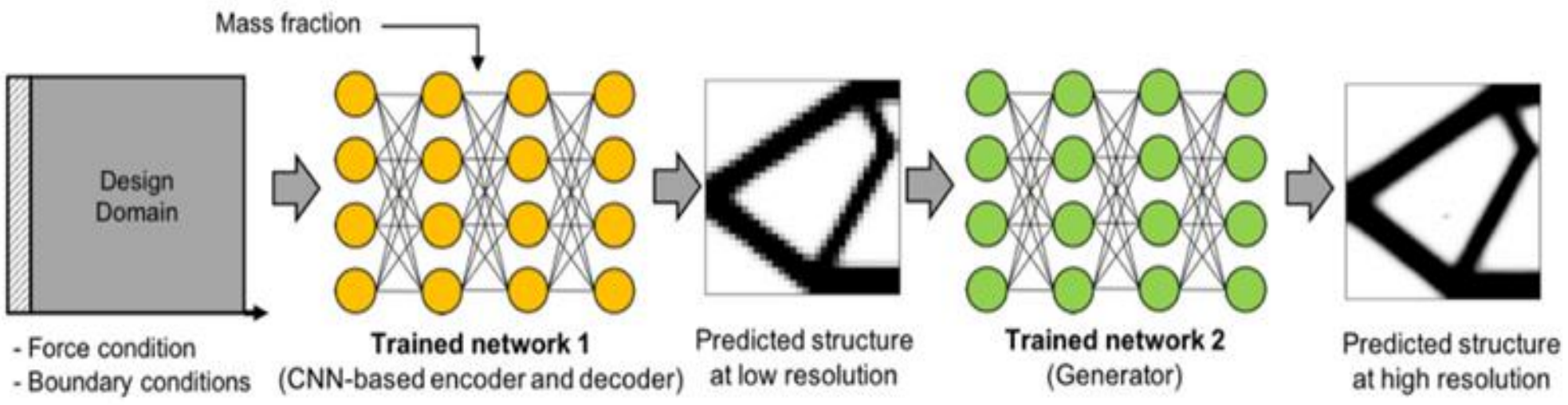

**Figure 6 Multistage (two-stage) process for non-iterative optimization (Yu et al., 2019)**

### 2.2.1. Single-Stage Process

First, as a single-stage method to predict optimization results, Abueidda et al. (2020) presented a CNN (ResUnet) that directly derives an optimized shape with 2D loading and boundary conditions, and volume fraction as the input for three material properties, such as linear elasticity with small deformation (i.e., without nonlinear constraints), nonlinear hyperelasticity with geometric nonlinearity, and linear elasticity with stress constraint (i.e., nonlinear constraint). Sharpe and Seepersad (2019) derived the optimized topology in real-time by inputting conditions, such as volume fraction, loading location, and material type, by using a conditional generative adversarial network (cGAN). Kollmann et al. (2020) predicted the optimal topology of 2D meta-material via encoder and decoder from ResUNet by using volume fraction, filter radius, and design objective (e.g., maximum bulk modulus, maximum shear modulus, or minimum Poisson's ratio) as the input.

Similarly, by using 2D optimization results obtained from the SIMP method under specific loading and boundary conditions as the reference data, Rawat and Shen (2018) used Wasserstein generative adversarial network (WGAN) to predict the optimal shape in real-time and then additionally predicted the optimization condition of the generation result. This approach allows the generated shape to be mapped to the corresponding optimization conditions, such as the volume fraction, penalty, and radius of the smoothing filter through CNN. Extending this scheme to the 3D domain, Rawat and Shen (2019a) predicted the 3D optimal topology through WGAN, and the optimization conditions (e.g., volume fraction, penalty and radius of smoothing filter) are predicted by CNN. Rawat and Shen (2019b) attempted to predict quasi-optimal topology by using volume fraction as a condition to conditional WGAN.

Furthermore, researchers have attempted to predict the optimized shape instantly aided by physics information from FEA for higher accuracy. For example, Zhang et al. (2019) used not only volume fraction and loading and boundary conditions for the input of deep CNN but also the strain field information through FEA to generalize this method to various loading and boundary conditions and various initial designs. Similarly, Nie et al. (2021) proposed cGAN-type TopologyGAN, in which the strain energy density and von Mises stress values obtained from FEA are used as the condition, along with the volume fraction, displacement, and loading and boundary conditions. The baseline of this process is shown in Figure 7.

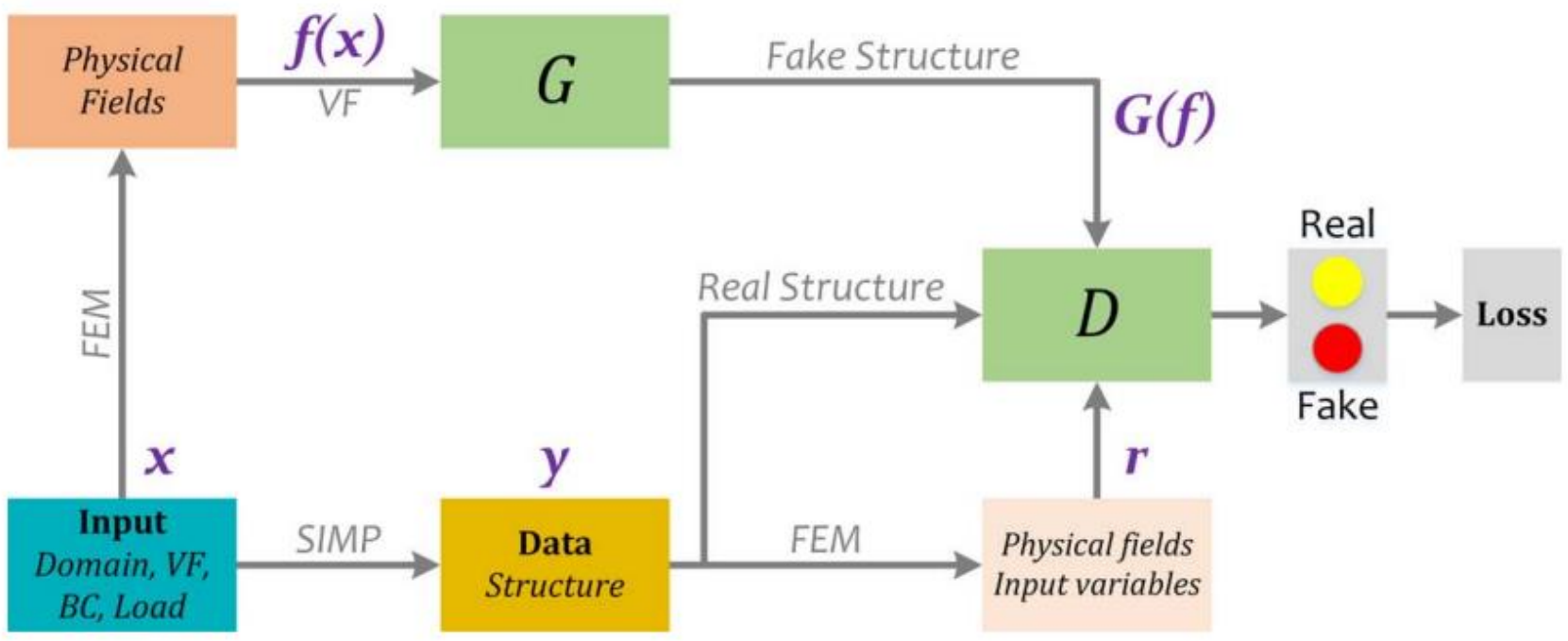

**Figure 7 Framework of physics guided cGAN (Nie et al., 2021)**

Cang et al. (2019) introduced a "theory-driven" mechanism based on adaptive sampling rather than a batch-mode training method to reduce computational costs in the optimization process. The deviation of the topology predicted by the NN from the optimality conditions for TO is calculated and used to select a new data point (optimization problem set) for the next training. This method allows for the better prediction of designs with optimal structural compliance for unseen loadings than using static data collection under the same computational budget.

### 2.2.2. Multistage Process

Scholars have also proposed a multistage non-iterative process to predict the optimization result with better engineering performance by integrating two or more networks.

First, Yu et al. (2019) performed TO in a non-iterative manner and then upscaled the predicted low-resolution image into high resolution. This study was the first to propose a near-optimal optimization process without repeated iterations. The low-resolution optimal images are first predicted by inputting x, and y-directional loading and boundary conditions as the input, and then the final optimal topology in high-resolution can be predicted with the low-resolution image. Li et al. (2019a) also applied a non-iterative method similar to that of Yu et al. (2019) to a conductive heat transfer problem. Low-dimensional topology images are generated through the generative adversarial network (GAN) based on the heat sink position, the heat source position, and the mass fraction information, and upscale the result through super-resolution GAN (SRGAN). Keshavarzzadeh et al. (2021) constructed a prediction model that approximates a high-resolution optimal design, given training data of multiple resolutions. To this end, they proposed a model integrating the disjunctive normal shape model (DNSM) at the end of a fully connected DNN to construct a multiresolution network with both low-resolution and high-resolution designs. After mapping low-dimensional parameters such as loading, boundary conditions, and volume fraction to the DNSM coefficient by using the DNN, the DNSM model is used to predict the optimal topologies of various resolutions.

Rade et al. (2020) proposed two methods to explore high-resolution TO for 2D and 3D features. The first is the density sequence (DS) prediction network that predicts the initial density by using the initial compliance and the volume fraction initialization to predict the final optimal density. The second method is coupled density and compliance sequence (CDCS) prediction network, which uses the intermediate density and the intermediate compliance information together to predict the optimal density though five repetitive processes. By using the first 30 iteration results from ESO, Qiu et al. (2021a) trained the U-net CNN model to predict the intermediate topologies and then used the evolution of the predicted topologies to learn the subsequent structure transformations using long short-term memory (LSTM). This method required less training data but could offer much broader applicability. Lew and Buehler (2021) utilized representations in latent space to explore the latent design space via the variational autoencoder (VAE)-LSTM model. The LSTM was trained to learn the trajectories in a latent space that could compress the optimization routes, allowing the system to learn the valid optimization progression.

In a study that additionally considered the manufacturing conditions, Almasri et al. (2020) used a conditional convolutional dual-discriminator GAN to accelerate TO, considering both a single general manufacturing condition and a mechanical condition. In the prediction of the 2D optimized shape, both geometrical (e.g., the total number of bars) and mechanical (e.g., boundary conditions, loading configuration, and volume fraction) constraints are used as the input to the generator, and two discriminators ensure that both mechanical and geometrical constraints are both satisfied. Furthermore, to evaluate the performance of the generated topology, a CNN-based network is used to predict the compliance of the generated structure.

## 2.3. Meta-Modeling

In the TO process, the calculation of objective function and sensitivity analysis process through FEA occupies a high computational cost. Therefore, some studies that accelerate the TO process through ML are conducted to accelerate the optimization process by replacing the solvers, such as FEA.

### 2.3.1. Objective Function

As a representative study to predict the objective function value, Lee et al. (2020) attempted to accelerate the optimization process by replacing the objective function calculation process through FEA with CNN, as shown in Figure 8. The compliance and volume fraction are predicted by inserting the intermediate material layout as an input, allowing for the optimality criteria (OC) methods to proceed immediately without undertaking FEA. Deng et al. (2020) also used DNN to approximate the objective function.

Kim et al. (2021) used the representative volume element (RVE) method for TO of functionally graded composite structure design, constructing a DNN-based continuous model to predict the effective elasticity tensor (a material property) for given design parameters to replace the FEA process in the iterative TO process. This methodology is applicable to both 2D and 3D problems.

In the work of White et al. (2019), which deals with the problem of multiscale TO, the computational process for effective elastic stiffness of meta-material is replaced by the NN surrogate model. Through this NN, microscale geometric parameters can be mapped to macroscale stiffness. Li et al. (2019b) used feed-forward NNs (FFNN) to map macroscale strain components and macroscale stress components. Using the series of stress values predicted by FFNN, they predicted whether the microstructure is damaged or not through CNN to provide a criterion for sensitivity analysis.

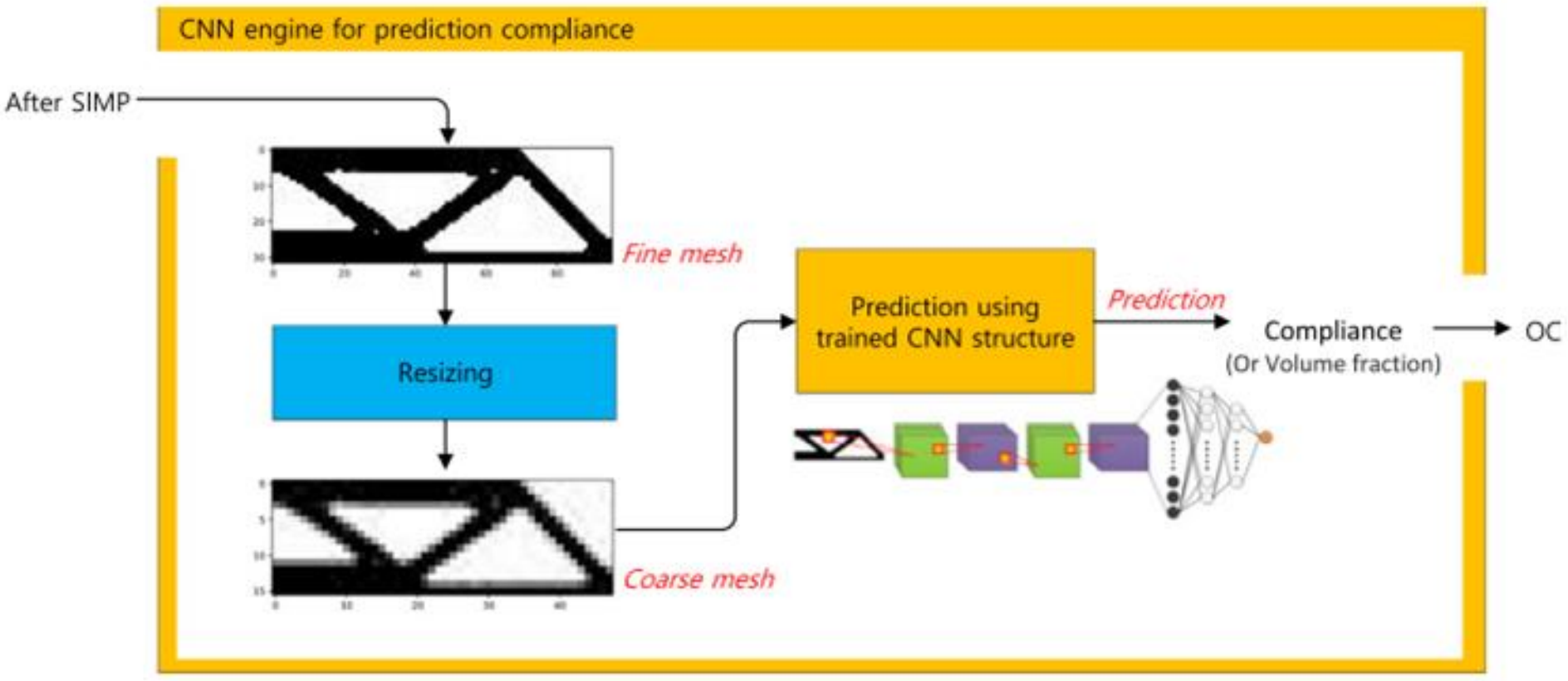

**Figure 8 ML in FEA meta-modeling for objective value prediction in TO (Lee et al., 2020)**

### 2.3.2. Sensitivity

As a study to predict sensitivity, Aulig and Olhofer (2014) replaced sensitivity analysis using MLP, a regression model, as shown in Figure 9. The material density of an element, which is a local state feature, and the displacement values of the element nodes obtained through FEA are used to predict the sensitivity. Xia et al. (2017) proposed a TO method applied with a constrained maximum-weight connected graph (CMWG) to remove the checkerboard pattern from the optimization results and used kriging models based on the support vector machine (SVM) to predict the sensitivity. Through the CMWG algorithm, the result of the previous ten iterations and the data of the surrounding elements are combined and used as training data to calculate the sensitivity value of the element.

Takahashi et al. (2019) attempted to find the global optimum by calculating the sensitivity through CNN in the TO process. Additionally, filtering was also performed at the end to enable the optimization results to be more meaningful from an engineering perspective. Chi et al. (2021) also trained the ML model to directly predict sensitivity without solving the state equations. However, in contrast to the work of Takahashi et al. (2019), they proposed an online training concept in the optimization process and used the history data of the optimization process as a training sample. In providing sufficient information for training while avoiding the most time-consuming step of solving the state equation, TO formulas with two discretization levels, fine-scale mesh and coarse-scale mesh, are used. Design variables are used from the fine-scale mesh, and the strain vectors are used from the coarse-scale mesh.

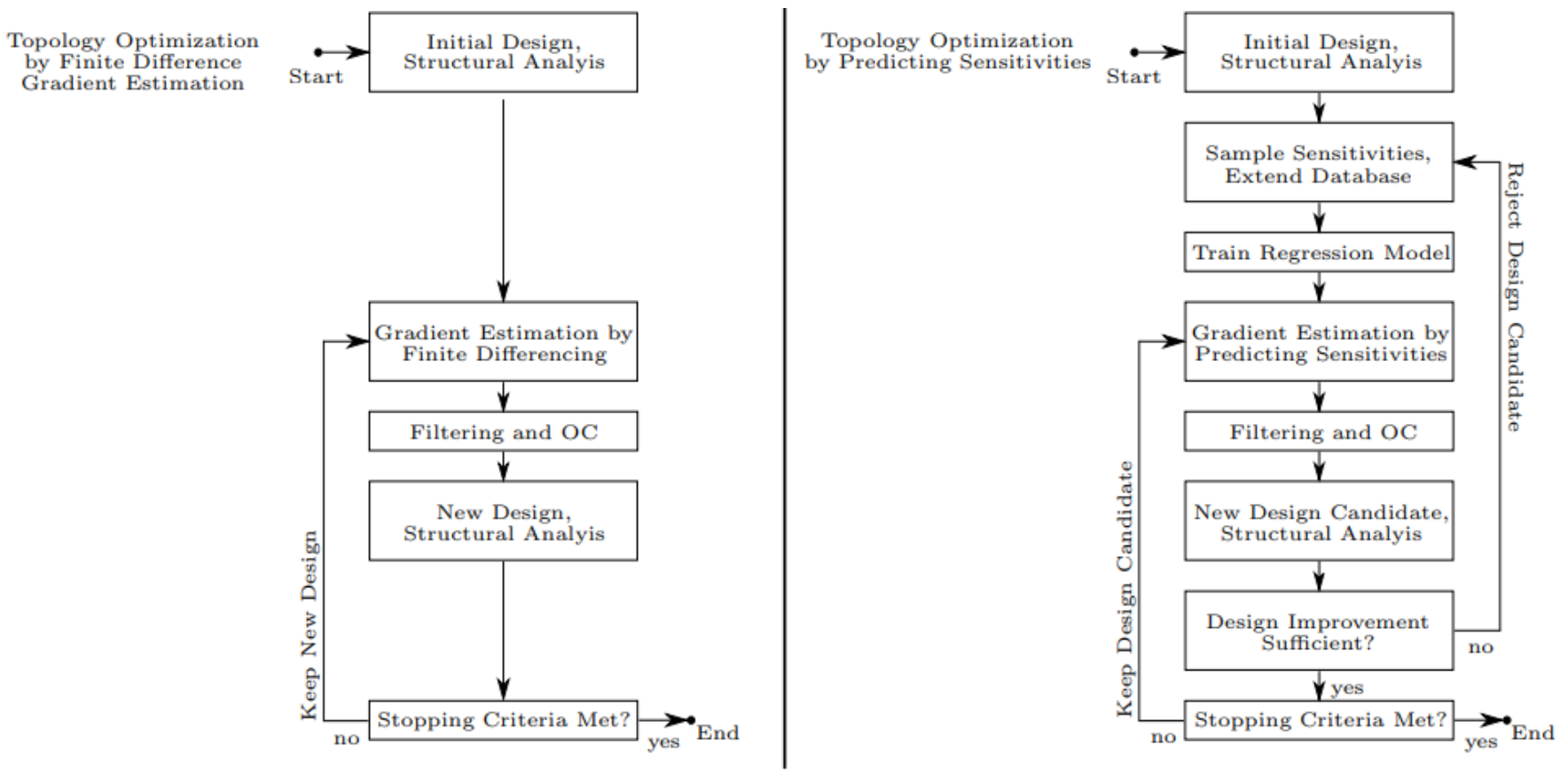

**Figure 9 Comparison between conventional TO and TO with ML as a meta modeling for sensitivity analysis (Aulig & Olhofer, 2014)**

### 2.3.3. Objective Function and Sensitivity

As a study for predicting both objective function calculation and sensitivity analysis, Qian and Ye (2021) used dual model NN for more accurate sensitivity analysis. The initial design obtained through SIMP is used to predict compliance through the forward model of dual model CNN, and the sensitivity of each element is calculated through the adjoint model. These NNs are used to perform forward calculations and sensitivity analysis, which can reduce time because they are much faster than higher-order simulations. Zheng et al. (2021) proposed a multiscale TO framework based on spinodoid TO for cellular structures. By simultaneously optimizing macroscale material distribution, microstructural design, and orientation, FE calculations can be replaced with a data-based surrogate model on a computationally expensive microscale to predict the stiffness tensor quickly and easily obtain the sensitivity of the microscale through backpropagation of gradients. Keshavarzzadeh et al. (2020) adopt the Gaussian process (GP) with a covariance kernel constructed with low-resolution finite element (FE) simulations to predict the maximum von Mises stress value and sensitivity for unseen features. GP was trained using a small amount of simulation data by treating the variability of material properties, geometry, and loadings as parameters in the parameter space.

#### 2.3.4. Constraints

Scholars have attempted to replace constraints used in TO with ML. Reliability-based TO, which is used for highly nonlinear or disjoint failure domain problems, require extremely high computational cost because the finite element method (FEM) is used to evaluate reliability constraints. Therefore, Patel and Choi (2012) used the classification-based probabilistic NNs (PNNs) to predict whether a random variable is a safe or unsafe region with a limit state function in the iterative TO process. The probability of failure is calculated using the predicted value, which is used to check the convergence of the solution in the optimization algorithm.

Zhou and Saitou (2017) replaced the manufacturing constraint with ML. They presented a method for designing a minimum compliance composite structure while reducing resin filling time in a transfer molding process. Random Forest was used to predicting the resin filling time, and the Kriging-interpolated level-set (KLS) approach and multi-objective genetic algorithm (MOGA) were used for optimization methods.

#### 2.3.5. Hybrid Approach

In the case of the studies above, the FEA calculation process was completely replaced. In this section, studies first predict the objective function value by replacing the FEA process with ML and then perform actual FEA to the predicted results with good engineering performance to use accurate objective function values in the optimization process. As a study that greatly reduced the computational cost of TO based on a stochastic algorithm, Sasaki and Igarashi (2019a) predicted the average torque and torque ripple value of the IPM motor through CNN to calculate the objective function. According to the predicted objective function value, probabilities P of FEA is calculated to determine whether to execute FEA or not. The probability P helps to skip unnecessary FE processes because FE calculations are only performed when P is large. In the existing genetic algorithm (GA), many candidates are randomly generated. However, as these initial shapes do not all satisfy the torque performance, CNN predicts the torque performance to reduce the number of samples to be subjected to FEM to save time and computing cost. Sasaki and Igarashi (2019b) also reduced the FE computation through a classification model that predicts the performance of IPM motors. In contrast to the work of Sasaki and Igarashi (2019a), this study used various input values and conducted various experiments based on multi-objective optimization. Doi et al. (2019) also accelerated multi-objective optimization (e.g., maximizing average torque and minimizing torque ripple) for IPM motors. Similar to the previous studies, the probability P to determine whether or not FEM is performed was calculated based on the average torque and torque ripple predicted by CNN. In addition, when the predicted average torque and torque ripple values exceed a critical value, the iron loss is calculated. Similarly, Asanuma et al. (2020) aimed to predict motor performance with a small amount of training data by applying transfer learning (classifier VGG16) to reduce the FEM process during GA.

### 2.4. Dimensionality Reduction of Design Space

To efficiently explore the design space in the iterative TO process, researchers have used the dimensionally reduced latent space through ML.

Guo et al. (2018) proposed an optimization process in the latent space for acceleration and efficiency and performed a multi-objective problem optimization for heat conduction. As shown in Figure 10, using the optimized image through SIMP as training data, latent space is learned through the VAE, and style transfer is also used to allow the generated images to resemble the training data. Several optimization methods, such as GA, gradient-based algorithm, and hybrid algorithm, were attempted by extracting the initial population from the trained latent space, of which MOGA performed the best.

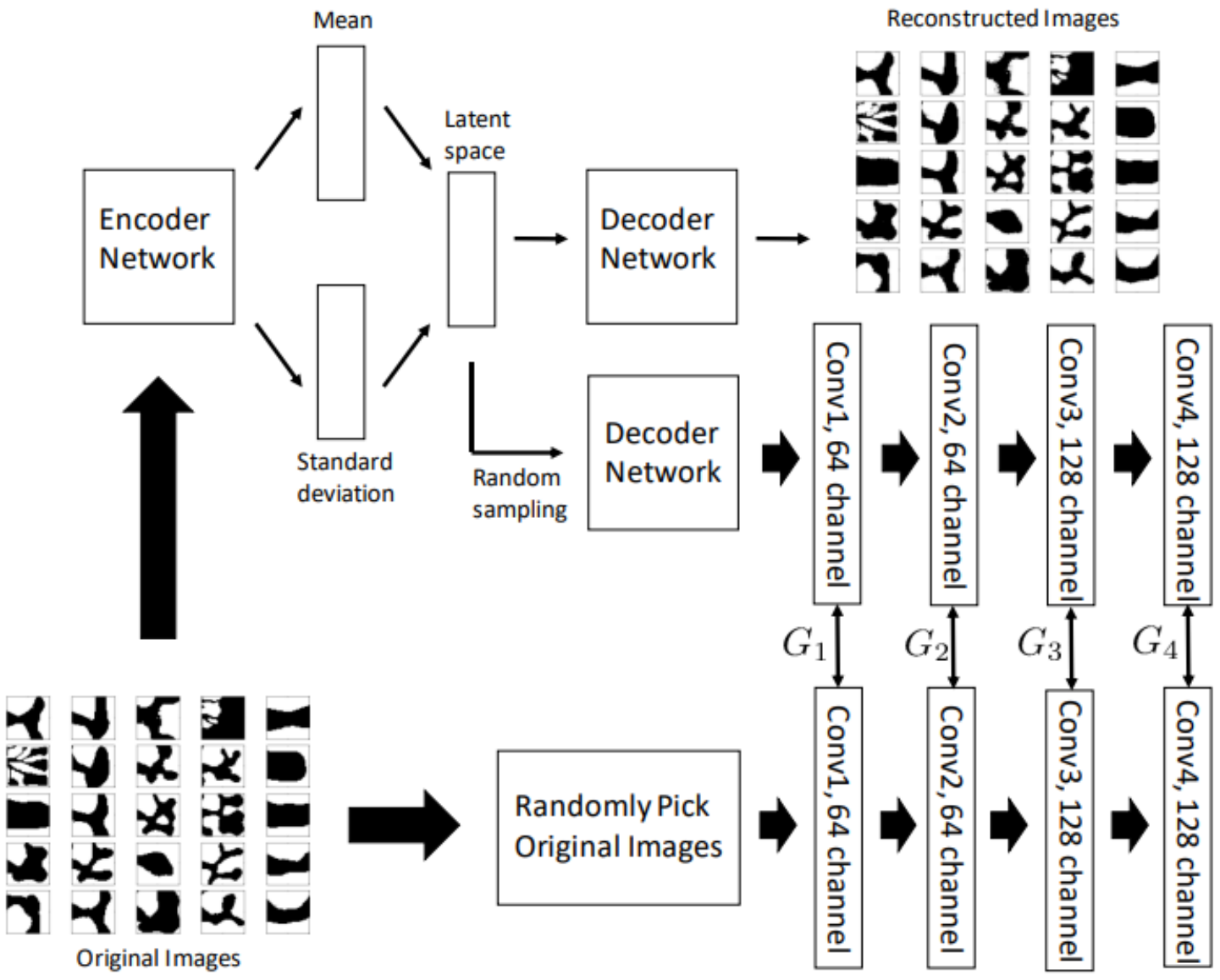

**Figure 10 VAE integrated with style transfer (Guo et al., 2018)**

Ulu et al. (2016) reduced the dimensionality of the existing optimized image in high-dimensional space by using principal components analysis (PCA) to generate new optimized topologies. The optimal topologies (e.g., reduced PCA weight) and the corresponding loading configuration are mapped using FFNN. With the trained model, PCA components (weight) for a new loading configuration can be predicted, and by synthesizing this with an eigen-image, new optimal topologies can be obtained. Li et al. (2019c) proposed efficient dimensionality reduction and surrogate-based TO to reduce the computational cost of TO of periodic structures. This study used logistic PCA to find the lower-dimensional representation of the topologies. To be used as training data, the design variables selected from the latent design space are converted into topologies through inverse transformation of logistic PCA, and the bandgap characteristic corresponding to each topology is obtained through a numerical model. With this, a surrogate model (kriging-based efficient global optimization) for the numerical model is constructed and used to predict the bandgap. By using this method, the optimal topology can be efficiently found within the global optimization.

## 2.5. Improvement of the Optimizer

### 2.5.1. Optimization Parameter Tuning

Scholars have also supplemented conventional TO methods by tuning the optimization parameters based on ML to quickly and efficiently search for parameters used in the optimization process.

The parameters applied to the moving morphable component (MMC) are tuned by experience and can obtain an infeasible solution. To solve this problem, Lei et al. (2019) combined ML to optimize parameters used in the MMC method. After obtaining the eigenvalue and eigenvector of the parameter vector through PCA, nonlinear regression was performed using support vector regression (SVR) and K-nearest neighbor (KNN), respectively. After that, the design variables for characterizing the optimal shape and the corresponding loading may be mapped. This approach is meaningful in that the design parameters of the MMC method, which have fewer parameters than the pixel-level SIMP method, are sampled using ML. Similarly, Jiang et al. (2020) used extra-tree (ET)-based image classification for particle swarm optimization algorithm to tune the parameters of the method of moving asymptote, which is used as an optimizer for MMC. The ET classifier extracts image features from the image dataset, then normalizes and quantifies them to perform k-means clustering. With the grouped clusters, users can judge whether the input image is feasible or infeasible. In this manner, users can quickly determine whether the optimized solution is feasible during the optimization process.

Bujny et al. (2018) proposed the A-EA-LSM, an advanced version of the evolutionary level-set method (EA-LSM), and complements it, to converge faster and find good designs in low-dimensional space. On this basis, they used a new graph-based parameterization instead of the MMC parameterization originally used in EA-LSM to increase the convergence speed by reducing the dimension. Furthermore, after generating various designs, the performance of each design was evaluated and trained for the NN via the ranking of the designs. Through this trained NN, various topology variations for EA-LSM can be predicted.

Many design optimization algorithms have a problem in that a wide range of tuning parameters for controlling the algorithmic function and convergence must be manually set. Accordingly, Lynch et al. (2019) proposed a ML-based framework to recommend tuning parameters to users to reduce the trial and error and computational cost associated with manual tuning. To this end, the framework consisted of two stages: a meta-learning stage that derives parameter recommendations from similar problems and a meta-modeling stage that uses Bayesian optimization to optimize parameters for a specific TO problem efficiently. This scheme allows tuning parameters with similar problems from previous repositories, which is determined using dissimilarity metrics based on the problem's meta-data, to the current problem by using a Bayesian optimization approach.

### 2.5.2. Reparameterization

To achieve efficient optimization, researchers have reparameterized the density fields using NN. In this method, TO is conducted by backpropagating objective function values obtained through solvers such as FEA in the training process. This method works similarly to the process of updating the sensitivity and can obtain the optimized image by optimizing the parameters of the NN.

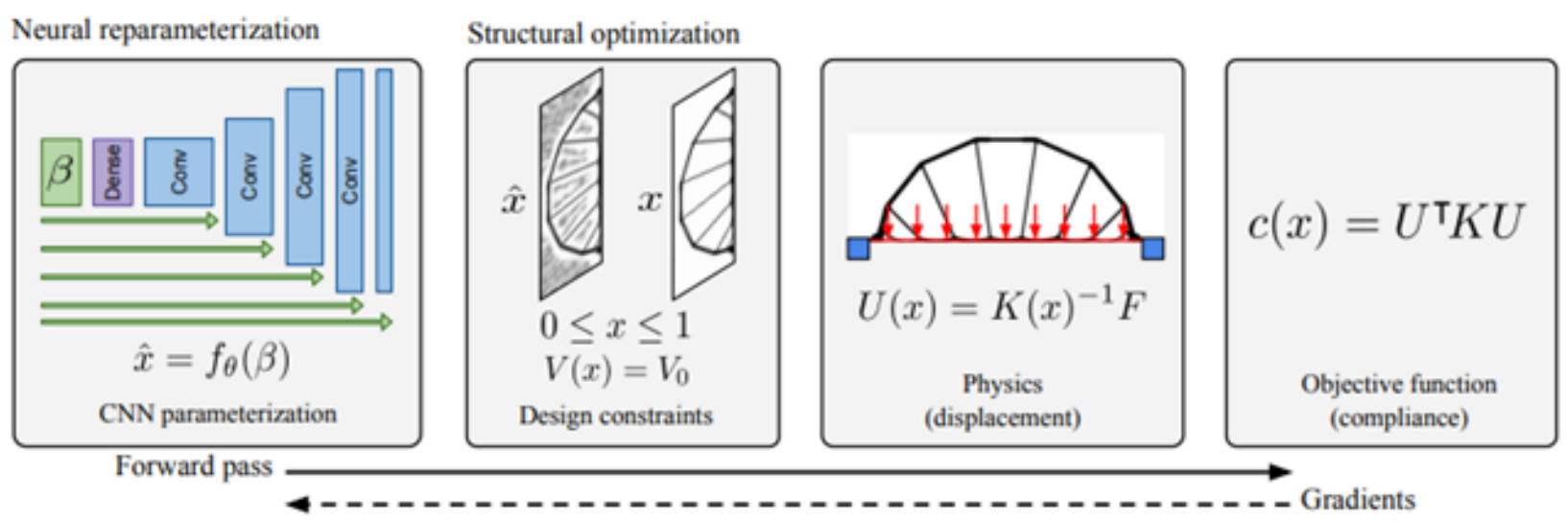

**Figure 11 Implementation of physics information in an NN-based TO (Hoyer et al., 2019)**

Hoyer et al. (2019) obtained the optimal topology for the input noise by adopting the deep image prior of CNN (Figure 11) and then evaluated the displacement of the topology via physics simulation and updated the network parameters based on the physics-based loss function by using the standard gradient-based algorithm. In this manner, the optimal topology can be obtained with the desired compliance. This process highlights the structural optimization method, which can be generalized to various optimization problems. Similarly, Zhang et al. (2021a) proposed TO via a neural reparameterization framework that would not need to construct a dataset in advance and would not suffer from structural disconnection for various problems, including nonlinear TO problems. Physics information was introduced into the loss function and the design constraint processing method, which was based on the volume-preserving Heaviside function with sigmoid transformation.

Halle et al. (2021) proposed a two-stage scheme named the predictor–evaluator network to predict optimized images without the use of a pre-optimized dataset. First, a predictor with an artificial NN (ANN) architecture was used to predict optimized images, with the input of static and kinematic boundary conditions and the target degree of filling both considered. Then, the generated image was processed by an evaluator that was used to calculate the objective function, and the network was trained using the gradient. Among the various evaluators used, the compliance evaluator was determined to be an FEM-based algorithm, and it can be used to predict the compliance of images generated by the predictor.

Similarly, scholars have applied the process mentioned above to the 3D domain by using an implicit neural representation of the density field with spatial coordinates utilized as the input. Deng and To (2020) used a new geometric representation method for deep representation learning (DRL) that approximates density fields in the design domain by using DNN to reduce design variables in a computationally expensive 3D TO problem and updates the network parameters in a direction to minimize the compliance obtained through FEM. The density function is parametrized by parameters of a NN by predicting the corresponding density value for the input coordinates based on the FEM results. As shown in Figure 12, Chandrasekhar and Suresh (2021c) also obtained the density value of elements through a NN, calculated the objective function value through FEA, and backpropagated the loss function. The method has been extended to 3D and 2D, and projection operators can be used to satisfy design and manufacturing constraints. Chandrasekhar and Suresh (2021a) proposed a Fourier-TO using neural network (TOuNN) that enhances the performance of the existing TOuNN to generate manufacturable designs. In this study, a length scale control strategy was proposed to control the minimum and maximum length scale by adding Fourier space projection to TOuNN. No additional constraint is required because the maximum and minimum dimension range is determined using Fourier projection. Chandrasekhar and Suresh (2021b) extended this study and applied it to multimaterial TO (MMTO also to predict the material types of elements. The proposed method calculates the loss by evaluating the volume fraction of each material per element for a given material type through the NN and by obtaining Young's modulus through FEA.

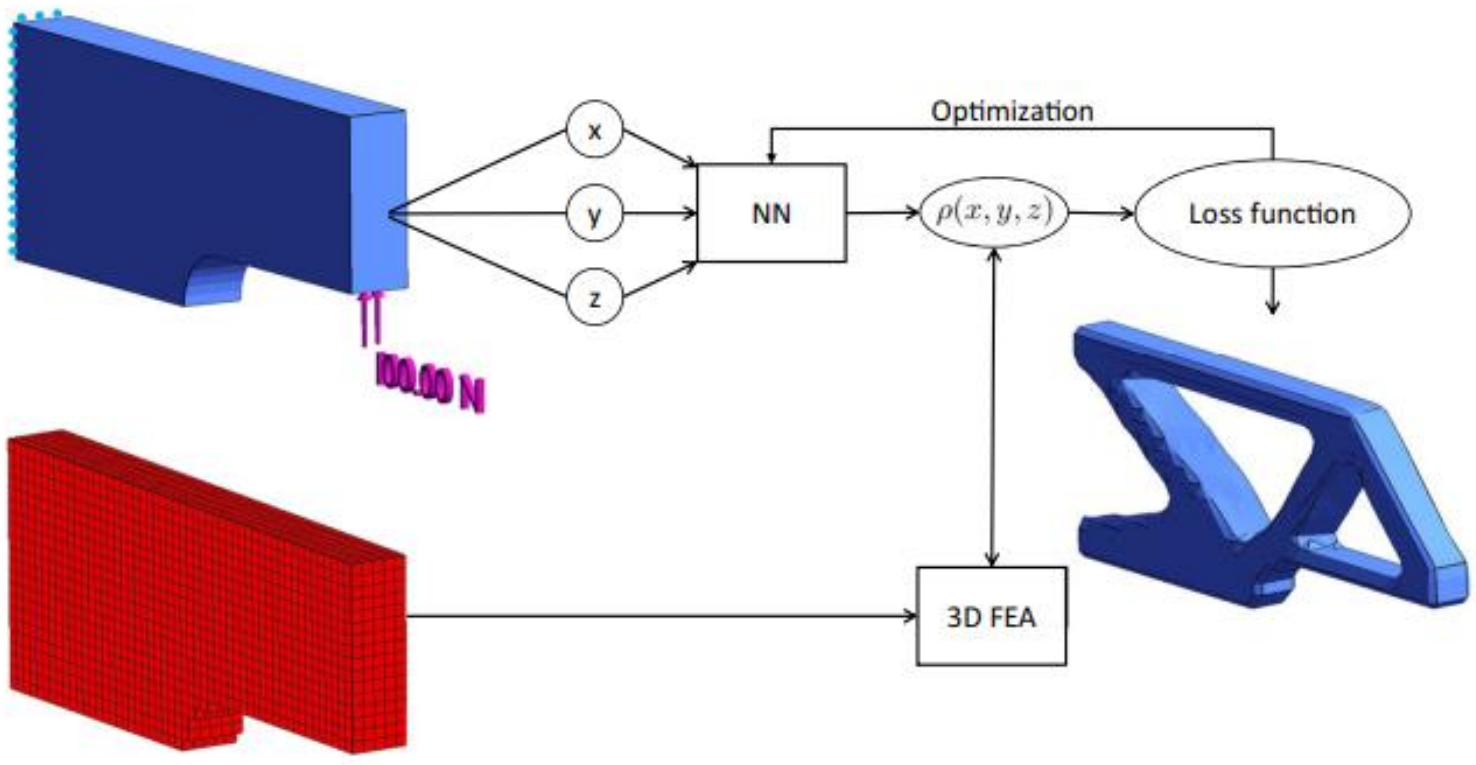

**Figure 12 Process of NN-based TO aided by FEA in 3D domain (Chandrasekhar & Suresh, 2021c)**

Zehnder et al. (2021) parameterized not only density but also displacement fields and used implicit neural representation to perform TO in continuous solution spaces in a mesh-free manner. First, the total gradient of the compliance loss with respect to the neural network parameters was computed to obtain the target density fields. Then, the density network was updated to minimize the MSE between the current and target densities. In this manner, convergence to the bad local minima could be avoided. Although the proposed method requires more computational time than the conventional FEM methods, this mechanism offers a fully mesh-free approach to TO for 2D and 3D problems.

Additionally, researchers have suggested integrating the methods above with generative models to generate various optimal topologies with high engineering performance. This topic is elaborated on in Section 2.6.

## 2.6. Generative Design

There are studies for generative design that quickly and diversely generate new optimal designs.

### 2.6.1. Naïve Approach

The most naïve method is to train a generative model using the existing optimized shape as reference data. Malviya (2020) compared the performance of CNN, UNet, and GAN to explore the effects of different types of deep generative models in a wide range of TO problems, along with various design constraints, and loading and boundary conditions, and consequently evaluated TOPCNN and TOPGAN as the best.

### 2.6.2. Engineering Performance-Aided Approach

Among the studies that aimed at generative design, many studies have suggested a generative design method that guarantees good engineering performance by integrating additional methods. To this end, there are studies that utilize physics information with the help of additional methods, such as solvers, NNs, dimensionality reduction, and clustering, for the designs generated by generative models.

*a) Engineering Performance Aided by Simulation*

First, scholars have reflected the objective function value obtained through the solver to the loss function in the generation process. Figure 13 shows a method of backpropagating the objective function value obtained through simulation, in which a generative design framework reflects the physics information. Most of these methods have the advantage of not requiring training data.

For example, Jiang and Fan (2019a) proposed a new type of global optimization algorithm, generative NN, which can produce high-performance meta-surface designs without training data. For the input noise, the efficiency of the generated structure is evaluated through an EM solver. By backpropagating the efficiency gradient, the generator can generate a high-performance device, allowing the generation of designs that reaches the desired performance. Extending this, Jiang and Fan (2019b) proposed global TO networks to design electromagnetic devices with the target wavelength and deflection angle as the additional condition. Furthermore, Jiang et al. (2019) extended this into an iterative optimization method. Using the same input as the previous research, which is the meta-grating deflection angle, operating wavelength, and random noise, devices generated by cGAN undergo a rigorous coupled-wave analysis (RCWA) solver to calculate the efficiency. Among them, devices with good efficiency are used as additional GAN training data, and devices with good efficiency can be generated during iterations.

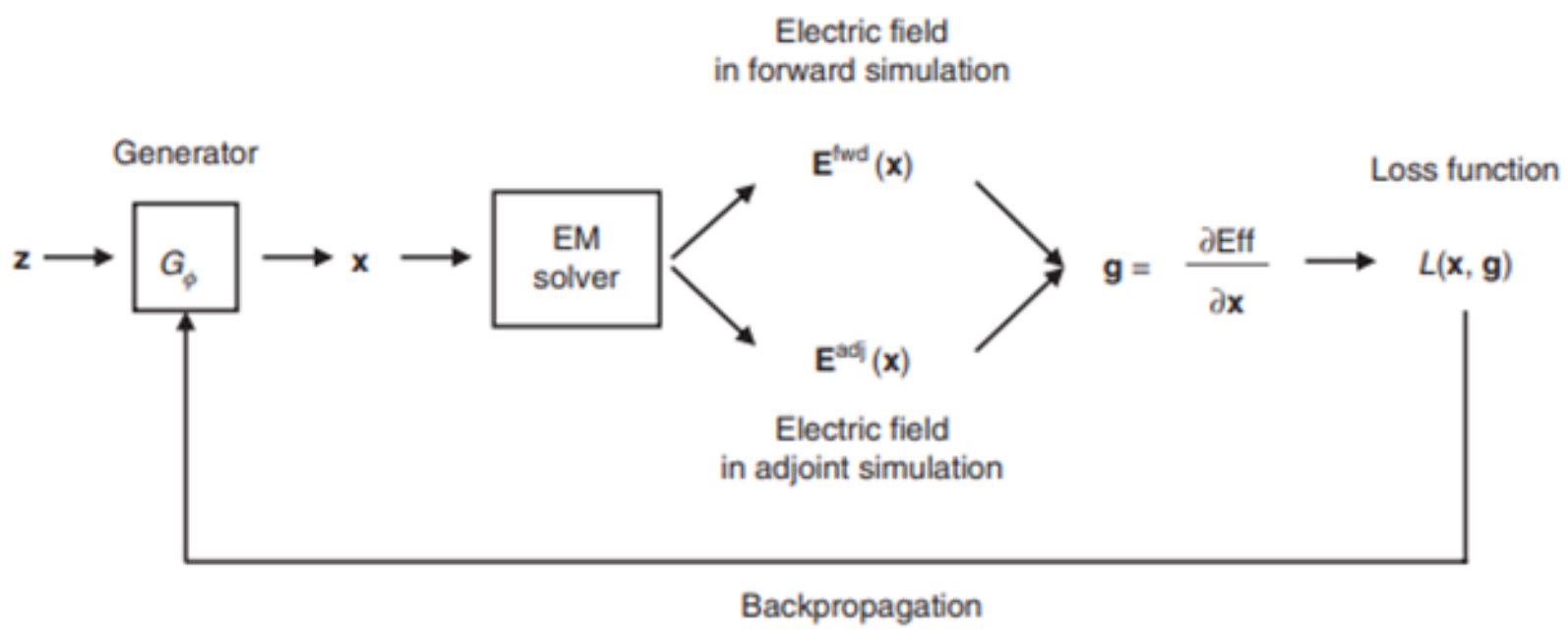

**Figure 13 Generative design framework using physics information-based loss function (Jiang & Fan, 2019a)**

Although with a similar purpose to the previous studies, scholars have used the physics information from FEM to update the training data instead of reflecting it into the loss function. Li et al. (2021) proposed SS (subset simulation)-based TO that integrates SS and GAN to generate samples from the failure distribution of each SS level to generate periodic structures with the desired bandgap properties efficiently. GAN is trained based on failure samples, which are high-performance features (optimal topology), to learn the corresponding failure distribution of the optimal topology. To generate only high-performance shapes, training was conducted by updating the dataset with samples with a good performance through FEM among those generated in the previous iteration. Through a data-driven approach, Yamasaki et al. (2021) produced new material distribution with good performance by constructing VAE, a deep generative model, based on current high-rank data, which is the best design evaluated by FEM among initial material distribution data from the SIMP method. By integrating this into the current high-rank data, the process of screening new high-rank data is iteratively repeated, and when the elite data converges, the final optimal topology can be derived.

*b) Engineering Performance Aided by NN*

To reflect the physics information in the training process, scholars have also suggested a method of evaluating the objective function value of design generated through a DL-based prediction model instead of a solver. Ha et al. (2020) proposed Fit2Form to automate the design of 3D task-specific robot gripper fingers, which consists of 3D CNNs and 3D autoencoders (AEs). A 3D AE that takes the truncated signed distance function volume of the target object as input and outputs the appropriate left and right finger shapes. After that, for the predicted left and right finger shapes, objective values, such as grasp success, force stability, and robustness, were predicted through a fitness network composed of 3D CNN to be reflected in the training process. Blanchard-Dionne and Martin (2021) also generated an optimal cloak with an optimal configuration using DCGAN and forward network based on the existing optimized shapes as the reference. For the designs generated by the generator of DCGAN, the scattering coefficient is also predicted through the forward model. Therefore, the performance of the image generated by the generator is reflected in the loss function to minimize the scattering coefficient to generate designs with high performance and, at the same time, deceive the discriminator. In increasing the accuracy of the DCGAN and forward model, FEA was performed on data predicted to have high performance among the generated shapes, and DCGAN and forward model were re-trained by adding the data as an additional dataset.

A study was also conducted to generate a high-performance design by slightly modifying the discriminator in GAN. Gillhofer et al. (2019) proposed a derivative-free optimization method. Furthermore, in contrast to the existing GAN, the discriminator is trained to classify whether the image generated by the generator corresponds to the region with higher responses or the region with lower responses. In this manner, the generator is trained to generate samples with a lower response, i.e., better performance.

*c) Engineering Performance Aided by Dimensionality Reduction*

Researchers have also used dimensionality reduction to reflect engineering performance in the generative design process can also be observed. Wang et al. (2020) constructed a model that combines VAE and regression, as shown in Figure 14, to solve the problems of high-dimensional topological design space, multiple local optima, and high computational cost. When VAE takes meta-material as input to reconstruct the meta-material, latent space that compresses the features of the meta-material is constructed. Simultaneously, a regressor that predicts a stiffness matrix with the latent variable is attached to the VAE architecture and trained together with the VAE model. In this manner, the latent space can contain the mechanical features of the microstructure.

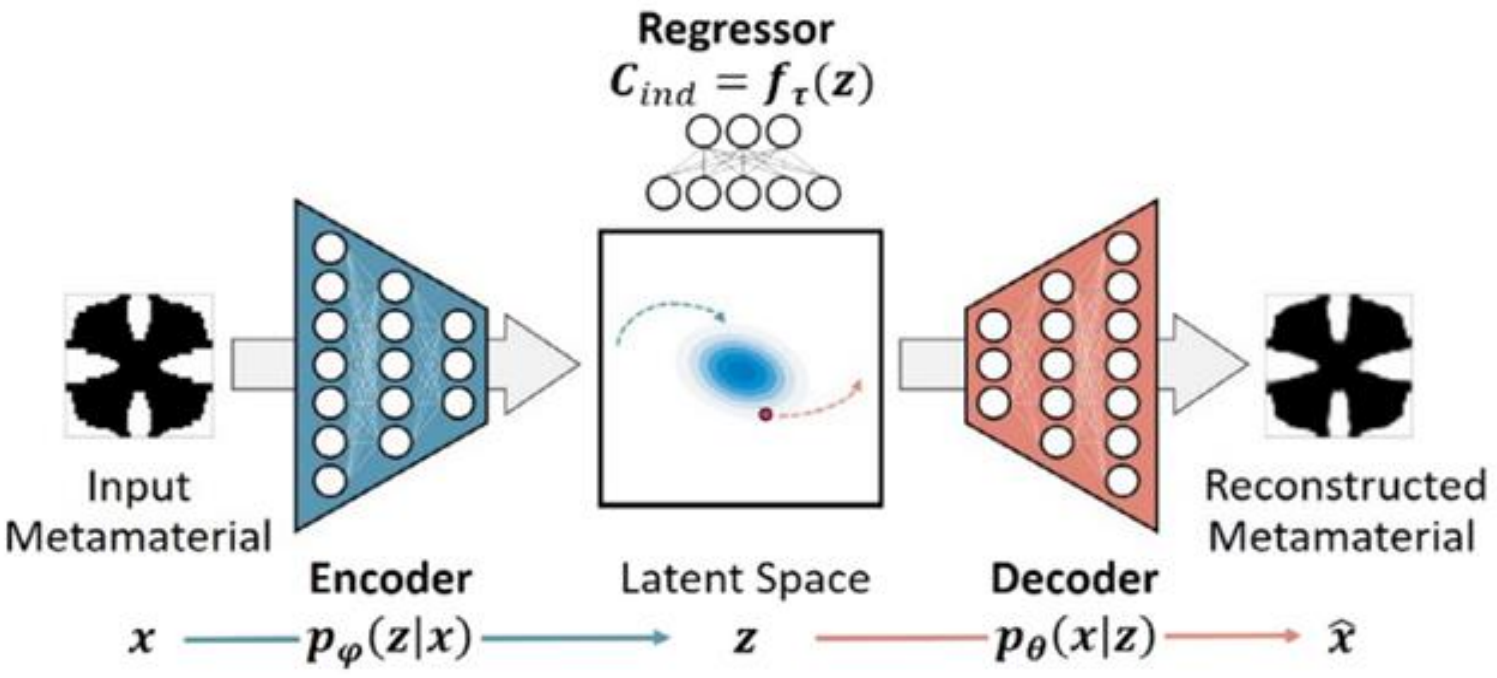

**Figure 14 VAE integrated with a regressor (Wang et al., 2020)**

Kudyshev et al. (2021) applied the adversarial AE (AAE) for photonic meta-device design by presenting three new global optimization frameworks. First, a model consisting of conditional AAE (c-AAE) and regression network is presented, secondly, c-AAE+DE framework, which uses differential evolution (DE) global optimizer for c-AAE, and finally, c-AAE+rDE framework that also uses important physics-driven regularization. Among them, the third model with an architecture similar to Figure 15 allows the physical regularization of the compressed design space to adjust the design space configuration to perform GO exploration better. This method utilizes the physics information inherent in the optical performance of meta-devices with complex material composition.

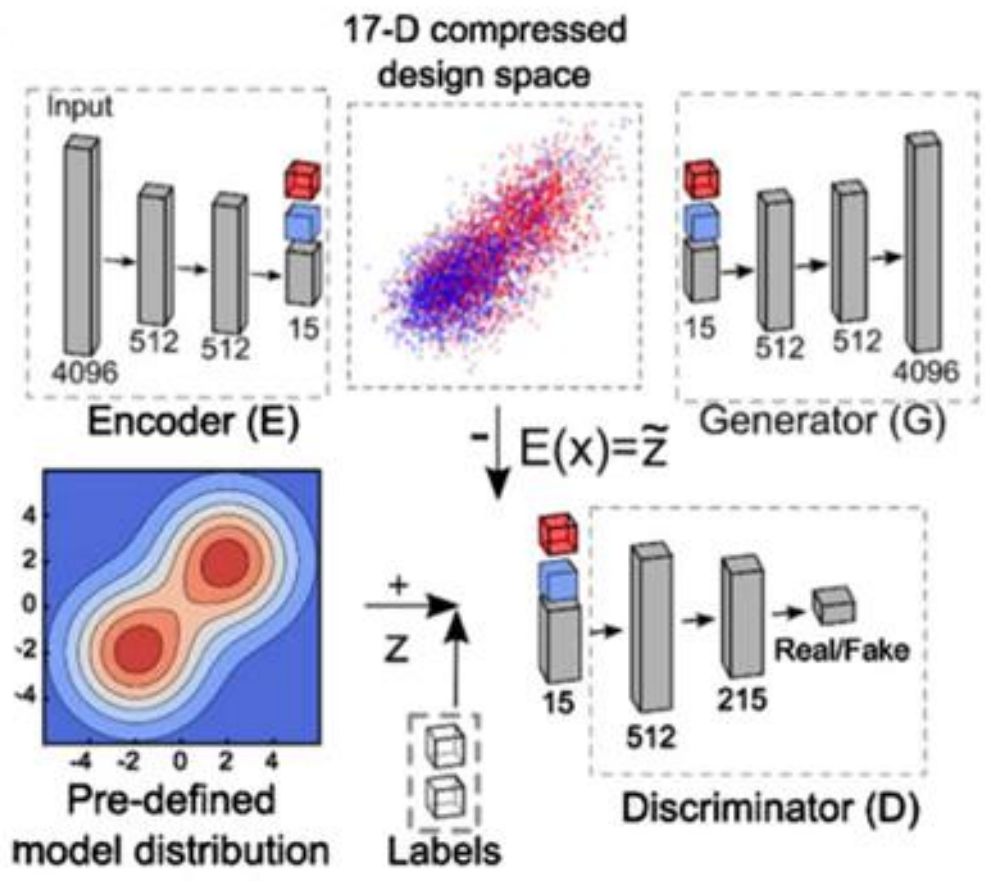

**Figure 15 Physics-driven c-AAE (Kudyshev et al., 2021)**

*d) Engineering Performance Aided by Clustering*

In the previous section, the studies used engineering performance information during the training process. Next, a study considered engineering performance by applying clustering to the generated designs. Sim et al. (2021) attempted to collect optimized shapes with meaningful performance among the designs generated through clustering analysis (CA) with GAN. Data were classified through k-means clustering based on the performance of topologies generated through GAN and DCGAN. They also proposed TO Validation Curve (TOVC) to collect optimized valid data among the generated results.

*e) Engineering Performance Aided by TO*

For more accurate optimization results, scholars have also conducted generative design studies that consider engineering performance by utilizing the conventional iterative TO method with DL. Kallioras and Lagaros (2020) utilized the optimization results from SIMP to diversify into several different results through convolutional layers. After extracting intermediate topology up to a specific iteration through SIMP, 24 different topologies were created through DBN and four different convolutional layers with the intermediate domain with reduced mesh size through re-meshing. The 24 different topologies created using this method are re-entered into SIMP, and the final 24 new optimized structures are created. In a similar manner, Kallioras and Lagaros (2021b) utilized SIMP with LSTM to estimate the final density distribution and performed generative design by using different image filters. This method showed great generalizability to various loading conditions, mesh sizes, and support conditions, as well as 3D problems.

### 2.6.3. Detailed Features

While reflecting on the physics information, scholars have attempted to learn detailed features of complex designs. Wen et al. (2019) used the progressive growth of GAN (PGGAN) to extract detailed features through high-resolution training data (Figure 16). In the initial training stage, starting from a low-resolution image and gradually training with a high-resolution image, large features with low-resolution data may be learned in the beginning, and then detailed features from high-resolution data may be learned as the training progresses. While growing the network architecture progressively, they could also augment the training dataset. For this purpose, high-performance devices are selected through the RCWA EM solver and added as the training data, while low-efficiency data are deleted. To supplement the previous study to learn global geometric trends, Wen et al. (2020) conducted the training by adding a self-attention layer after the deconvolution and convolution layers in the generator and discriminator, respectively.

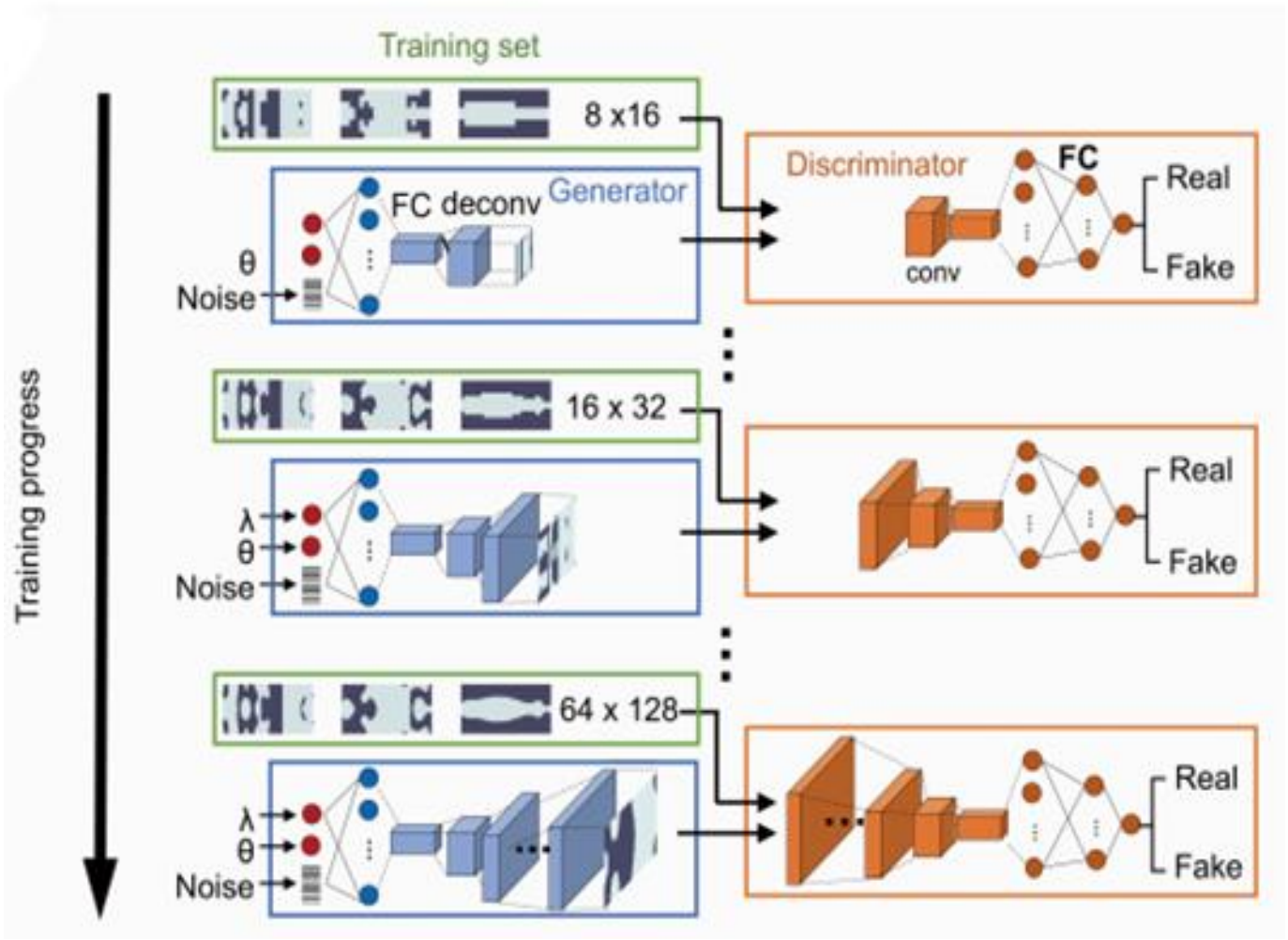

**Figure 16 Schema of PGGAN-based TO (Wen et al., 2019)**

### 2.6.4. Improved GANs

Shen and Chen (2019) introduced a new type of GAN to generate various topologies by proposing a convex relaxing CGAN that converges more efficiently and stably than the existing CGAN. In this method, not only the training sample and condition but also a random condition was additionally inputted to the discriminator, and the minibatch discrimination technique was applied to allow more stable convergence and prevent the mode collapse phenomenon in which the generator continues to generate similar shapes.

### 2.6.5. Manufacturing Constraints

In general, many cases claim that manufacturing is impossible by simply referring to the topological optimization results. To apply manufacturing constraints to TO, Greminger (2020) used synthetic 3D voxel training set to represent the distribution of a set of geometries that can be manufactured by a specific manufacturing technique to generate specific manufacturable 3D designs through multiscale gradient GAN.

### 2.6.6. Aesthetics-Based Generative Design

Moreover, Oh et al. (2018) proposed a generative design method considering aesthetics. Their study presents a method to generate new designs by training BEGAN with authentic wheel images and postprocessing the generated image through the SIMP as an initial design to derive final wheel designs. Although data through GAN does not guarantee stiffness, the proposed method can consider both stiffness and aesthetics by integrating SIMP method. Furthermore, Oh et al. (2019) performed optimization with the same purpose by using SIMP to resemble the reference data, filtered similar designs, and then trained BEGAN with the data to generate various wheel designs. This data is filtered and entered into SIMP again as reference data and repeated until various designs are generated beyond the desired goal, which is called the Deep Generative framework (Figure 17). This work has modified the SIMP method to resemble the reference design and considered both stiffness and aesthetics by generating new designs while filtering similar designs.

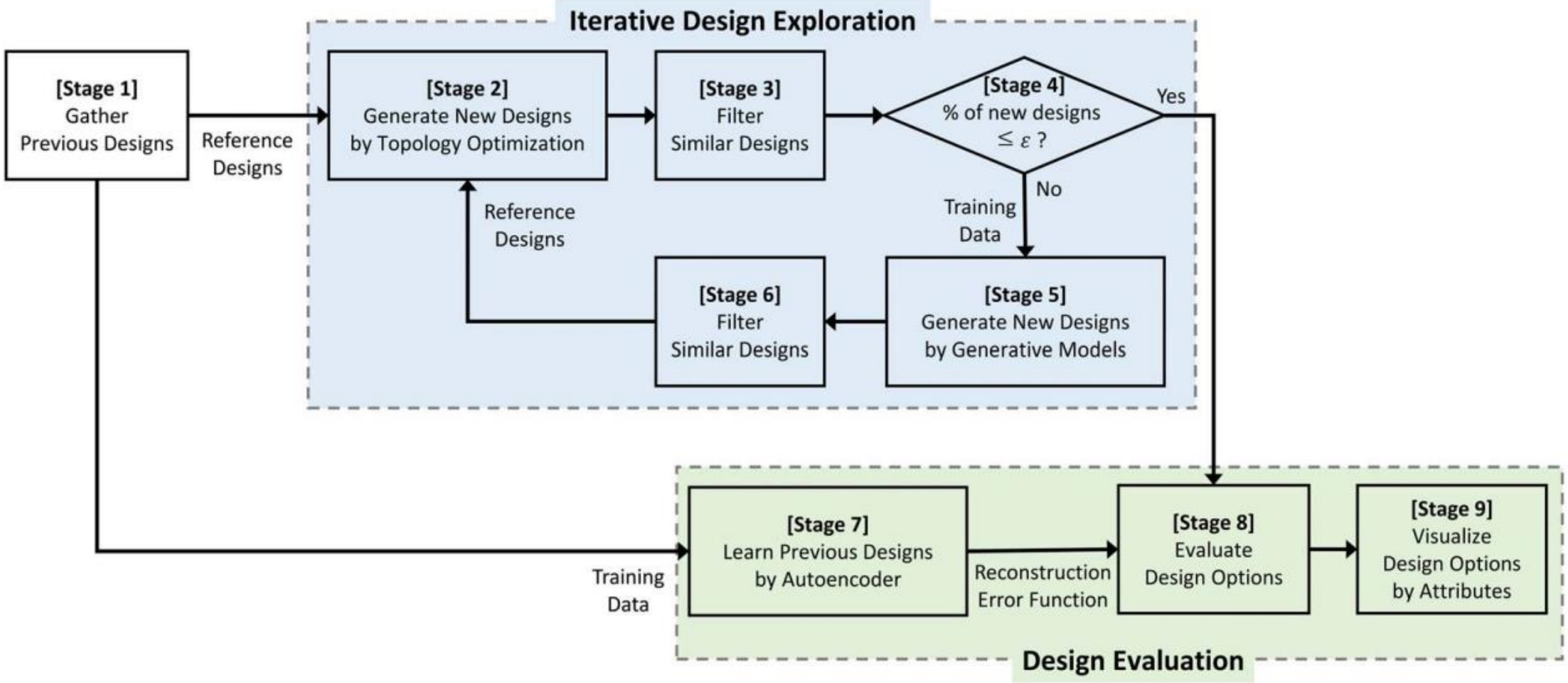

**Figure 17 Deep generative design framework (Oh et al., 2019)**

### 2.6.7. Reinforcement Learning (RL)

Finally, scholars have studied RL algorithms to generate a vast amount of generative design in a new approach. Sun and Ma (2020) extended SIMP and BESO, which is the main density-based structural TO methods, based on RL exploration strategies. Similarly, Jang et al. (2022) proposed a generative design process based on RL to maximize design diversity. The study has the distinction of guaranteeing both maximum diversity and aesthetics and proposed a RL-based generative design process that applies two networks by defining the maximization of design diversity as a reward function. They used TopOpNet to approximate the TO process and found the design parameter combinations with maximum diversity through GDNet. The overall process is shown in Figure 18.

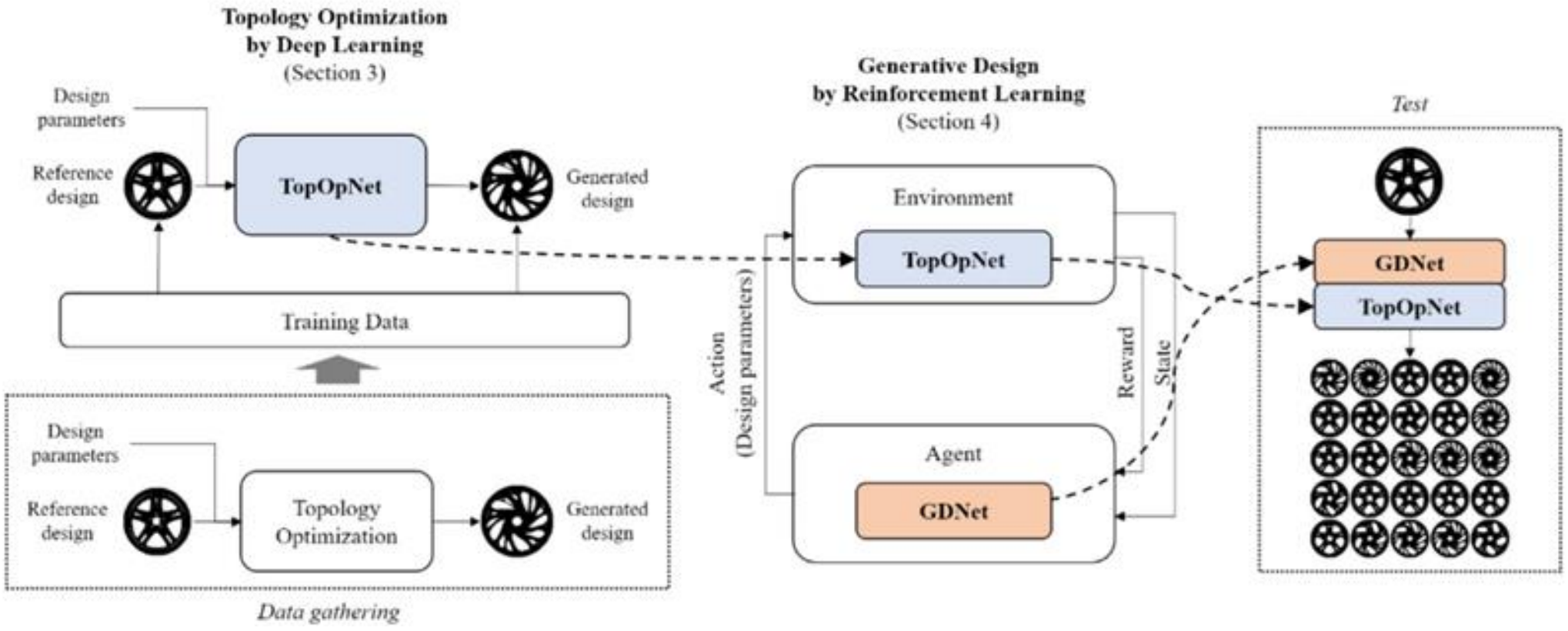

**Figure 18 Generative design of wheels via RL (Jang et al., 2022)**

## 2.7. Postprocessing

The purpose of the studies pertaining to the postprocessing of the optimized image by applying ML is largely divided into two categories. First, as the generated optimized images are often low-resolution images, the studies convert them into high-resolution images (Figure 19). Second, other studies have proposed a structural boundary processing method through ML to postprocess the generated images with gray scale or unclear boundary (Figure 20).

### 2.7.1. High Resolution

A large computational cost is required to obtain a high-resolution optimal image instead of a low-resolution image. If high-resolution images can be predicted from low-resolution structures, then a large amount of computational cost can be reduced. Therefore, by studying ML to obtain high-resolution optimized images, Napier et al. (2020) used ANN to upscale a low-dimensional input to a high-dimensional one. In this study, the FFNN is trained to upscale the low-resolution data to high-resolution data by using the preprocessed data of high-resolution beam optimization data to low resolution through averaging. Similarly, Wang et al. (2021) used CNN to map the relationship between low-resolution and high-resolution structures in TO problems. This approach is helpful for large-scale structural design.

In addition, Xue et al. (2021) proposed a super-resolution CNN (SRCNN) framework with superior applicability to improve the resolution of TO designs and the high-resolution TO (HRTO) method established by implementing SRCNN and pooling strategy. SRCNN is a network that maps low-resolution images to high resolution, and is available for 2D and 3D problems with arbitrary boundary conditions, all design domain shapes, and various loadings, demonstrating superior applicability and high efficiency. In addition to this, this study was able to solve the 3D HRTO problem by extending the 2D SRCNN.

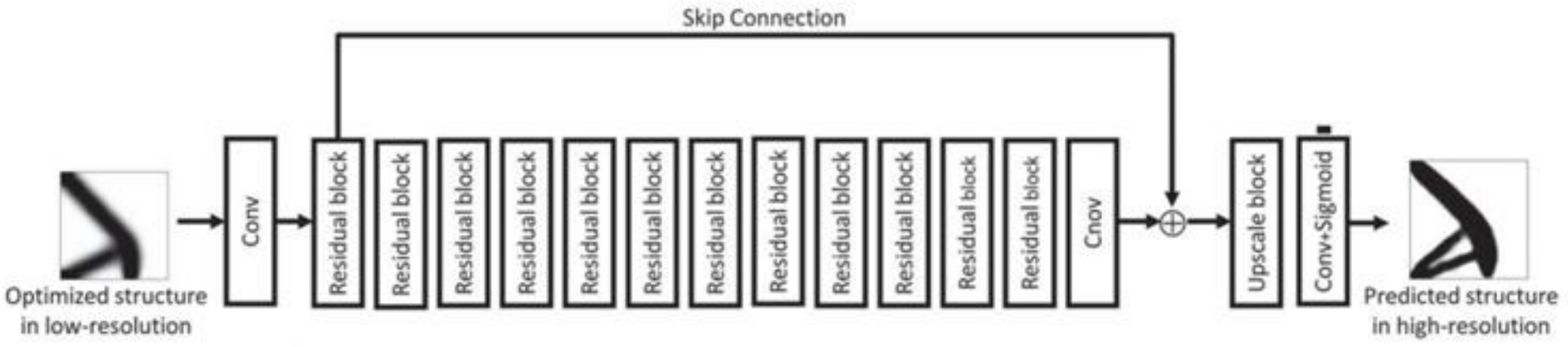

**Figure 19 Postprocessing high-resolution structures (Wang et al., 2021)**

### 2.7.2. Boundary Processing

Chu et al. (2016) used the SVM-based boundary processing method as an effective 2D structural boundary processing to obtain clear and smooth boundaries. Extending this into the 3D domain, Strömberg (2020) also used SVM for postprocessing of the optimized 3D design to smooth the boundary by classifying the outermost part of the optimized structure into parts with or without material. Furthermore, for design optimization for AM, Strömberg (2019) classified the benchmark optimized structure from the SIMP method into parts with or without elements by using the 1-norm SVM to automate the postprocessing and then added the resulting structure lattice structures. Karlsson et al. (2020) used SVM to obtain an implicit surface representation of the TO result. Boundaries were processed like those in the previous works, but this study also conducted an internal design through a Boolean operation.

To automate the combination of TO and shape optimization for boundary processing, Yildiz et al. (2003) used MLP NN. After performing TO and changing the grayscale image output as a mesh image, the shape of each hole is labeled with shapes, such as triangles, squares, trapezoids, and circles. After predicting the shape of the hole obtained from the result of TO, shape optimization is performed based on the predicted shape to select a feature-based model.

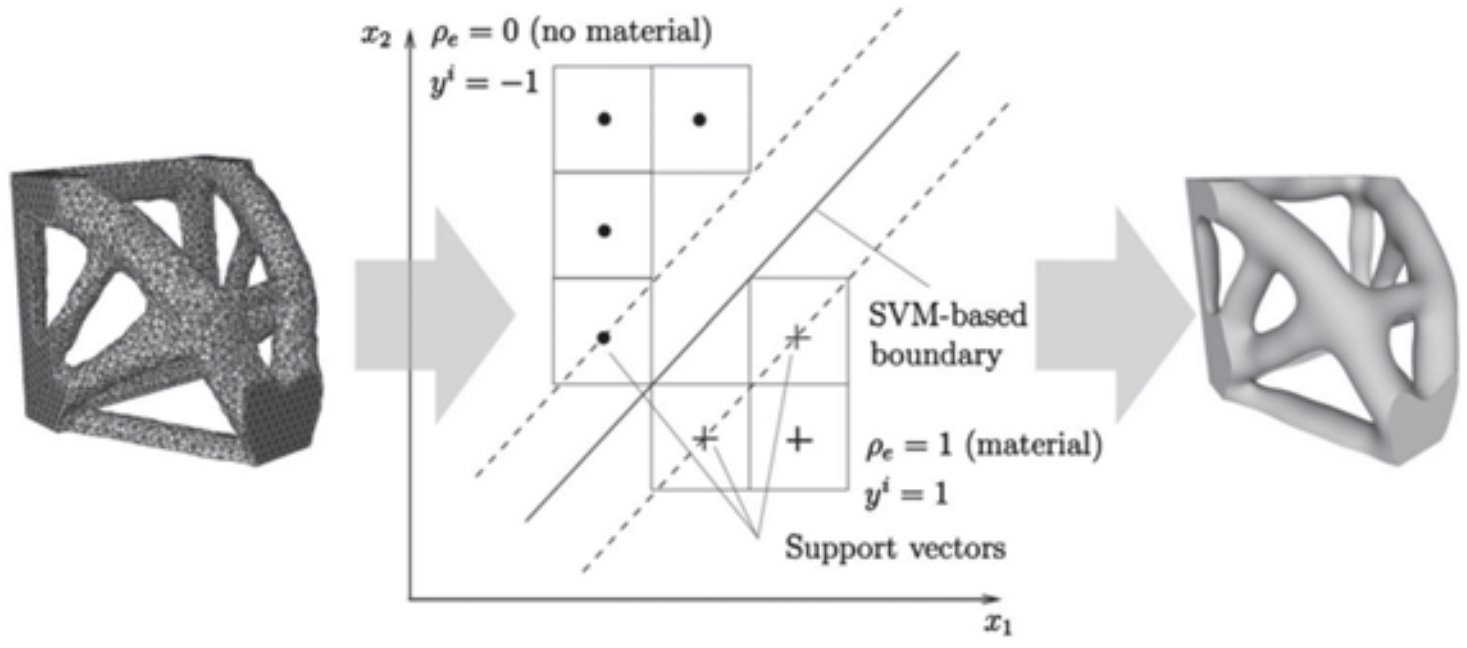

**Figure 20 Postprocessing for boundary processing (Strömberg, 2020)**

## 2.8. Multiscale, Multimaterial, and Other Processes

Other ML-based TO studies have a purpose other than the previously defined ones to solve optimization problems, such as microstructural TO (MTO), multimaterial density-based optimization, and other structural design problems.

### 2.8.1. Multiscale Process

MTO, a simultaneous optimization of macroscale topologies and microscale structures, has a problem of high computational cost and microstructural connectivity loss. To reduce the computational cost of such MTO, scholars have conducted iterative TO using K-clustering. Among them, Kumar and Suresh (2020) reduced computational cost and guaranteed connectivity between microstructures in MTO and performed TO by K-clustering elements based on the basic strain tensor and SIMP density. This methodology consists of three steps: clustering, optimization, and connectivity. This study proposed a density–strain-based method with a fixed clustering pattern according to the static density and stress state, as shown in Figure 21. However, fixed clustering patterns can limit the performance. Consequently, Qiu et al. (2021b) excluded density from CA to optimize cellular structures with uniform porosity and used the direction and ratio of principal stress for clustering. By introducing a stress ratio into clustering, the stiffness performance and connectivity of the microstructure can be improved. The clustering pattern should be updated by the current stress tensor in grouping the microstructure by using the dynamic clustering method. When clustering, the stress direction-based cluster analysis is first performed, and then the elements of each direction cluster are further clustered into several groups according to the stress ratio.

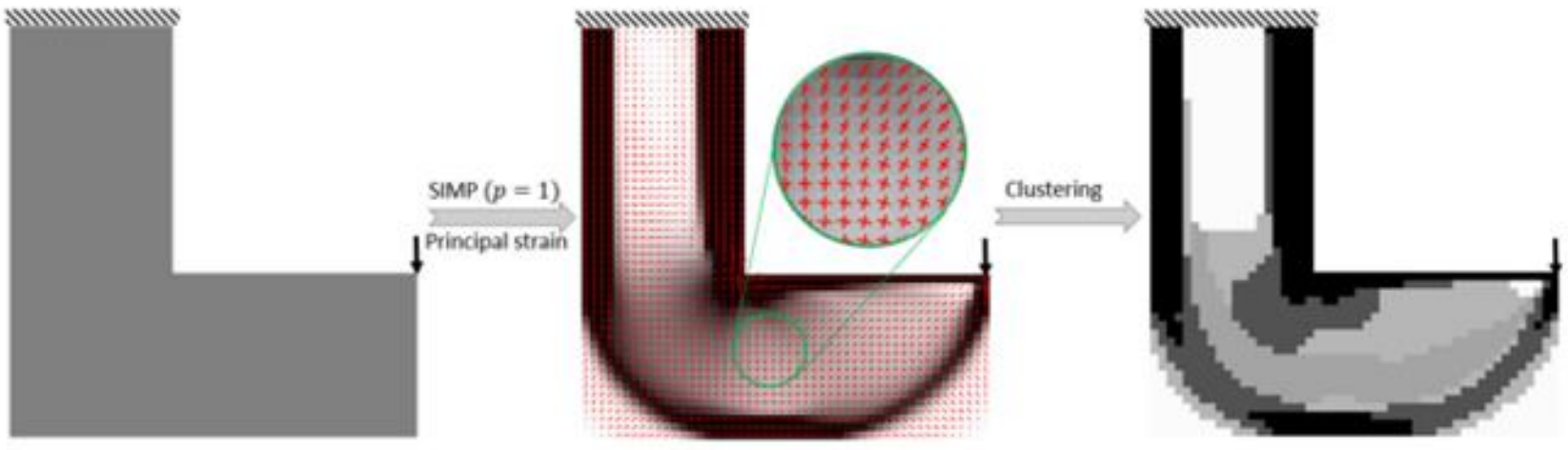

**Figure 21 Combined density–strain-based clustering method for MTO (Kumar & Suresh, 2020)**

### 2.8.2. Multimaterial Process

By conducting a study on clustering techniques to reduce design variables of multimaterial density-based optimization, Liu et al. (2015) demonstrated the optimization process by dividing the steps into continuous density distribution, clustering, and meta-model-based optimization steps. First, with an intermediate density distribution obtained through iterative iteration, the continuous density distribution is classified into a finite number of clusters based on similarity. In this case, the number of clusters means the number of materials used. Finally, the optimal material distribution can be obtained by constructing a meta-model (kriging interpolation) and updating it iteratively according to the global optimization algorithm (i.e., GA).

### 2.8.3. Others

Gaymann and Montomoli (2019) proposed the fluid-structure TO (FSTO) problem and the structural TO as gamification by optimizing the fluid structure that minimizes the pressure difference using DNN and Monte Carlo tree search. As an optimization method that proceeds in a manner in which the winner among the players' shapes can be determined, the DNN receives the player's shape as input to determine who is the winner and whether to end the game and predicts the pressure difference (Figure 22). Based on these results, the process of determining the player's future movements through a Monte Carlo search is repeated, and the final optimization shape is obtained at the end of the game.

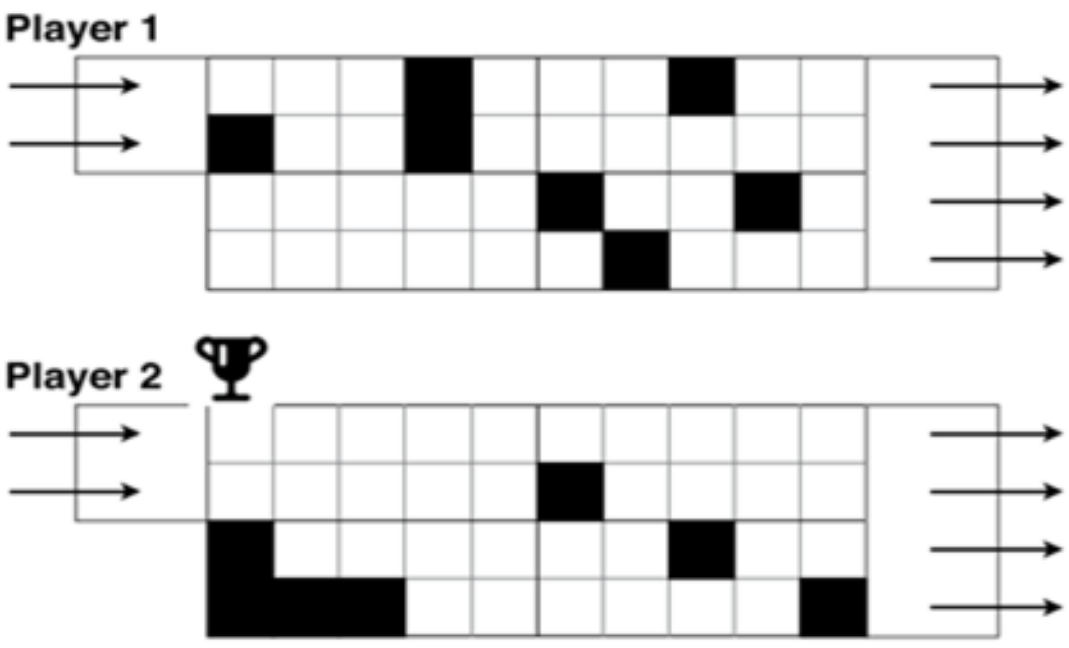

**Figure 22 Gamification for FSTO with two networks competing against each other (Gaymann & Montomoli, 2019)**

Hayashi and Ohsaki (2020) demonstrated the TO of truss structure using graph embedding through RL. To consider the connectivity between the members, the truss structure is regarded as a graph, and an optimization topology can be obtained by performing RL. In addition, Brown et al. (2022) utilized RL to optimize elementally discretized 2D topologies with a multistep progressive refinement approach to improve the computational cost. The proposed method can be tested on unseen load cases during training.

## 2.9. Relations between MLTO Purposes

Previously, ML was applied in TO to describe the process from a TO perspective. The main purpose of applying ML was classified into seven categories. For each purpose, the advantages and disadvantages of accuracy and efficiency inevitably exist due to the tradeoff relationship. Although MLTO has been divided into seven purposes, the boundaries cannot be strictly distinguished. The relations between the varying purposes of MLTO can be described as follows.

Various studies have replaced the inefficient process in iterative TO with ML. For example, the studies presented in Section 2.1 were aimed at accelerating TO by combining ML at the end of the iterative TO, thereby ensuring accuracy and improving efficiency. The studies in Section 2.3 replaced the time-consuming simulation stage in the iterative TO with an ML-based meta-model to reduce the computational cost of the conventional TO. Some studies in Section 2.4 utilized the latent design space to efficiently seek the optimal solution. The studies attempted to accelerate the TO while maintaining the iterative process to ensure accuracy. Meanwhile, the studies in Section 2.2 were aimed at generating the optimal topology in one shot by using ML without the need for implementing a time-consuming iterative process. This method can instantly provide optimal topologies once the datasets are trained. However, the accuracy of this model may be low due to the pixelwise training of the reference data. In addition, the computational cost, including the data collection cost, can be considered a burden as evidenced by the computational complexity of the supervised MLTO. Aimed at addressing these issues, the studies in Section 2.5.2 would reparameterize the density field by using neural networks that utilize physics information, such as structural compliance to the loss function. This method can replace the optimization process by updating the parameters of the neural network. In addition, much fewer datasets are required, and structural connectivity can be ensured. However, as most studies that use this method iteratively evaluate objective function values in FEA, the computational cost problem still exists. The limitation of the MLTO will be further explained in Section 4.

# 3. ML Perspective: How to Use ML

ML techniques used to improve TO can be classified from various perspectives based on the different usages of learning algorithms (Section 3.1), input data form (Section 3.2), ML loss (or objective) function (Section 3.3), usage of physical information (Section 3.4), and various applications and generalizability (Section 3.5).

## 3.1. Learning Algorithm

This section analyzes the types of ML applied to TO. Types of ML are classified and compared by supervised learning which has labeled outputs, unsupervised learning which has unlabeled outputs, semi-supervised learning, an algorithm conceptually situated between supervised learning and unsupervised learning, and reinforcement learning, consisting of action and reward algorithms (Table 2). An explanation of the DL methodology is given in more detail in *Appendix B.*

**Table 2 Classification of studies according to the learning algorithm**

| Category | Methodology | Use | Studies |
|---|---|---|---|
| **Supervised Learning** | Neural Network | Prediction | Yildiz et al., 2003; Patel & Choi, 2012; Aulig & Olhofer, 2014; Bujny et al., 2018; White et al., 2019; Deng et al., 2020; Deng & To, 2020; Chi et al., 2021; Keshavarzzadeh et al., 2021; Zheng et al., 2021 |
| | | Optimization | Bi et al., 2020; Napier et al., 2020; Kim et al., 2021; Chandrasekhar & Suresh, 2021a; Chandrasekhar & Suresh, 2021b; Chandrasekhar & Suresh, 2021c; Halle et al., 2021; Zehnder et al., 2021 |
| | Convolutional Neural Network | Prediction | Doi et al., 2019; Gaymann & Montomoli, 2019; Li et al., 2019b; Sasaki & Igarashi, 2019a; Sasaki & Igarashi, 2019b; Takahashi et al., 2019; Asanuma et al., 2020; Lee et al., 2020 |
| | | Optimization | Banga et al., 2018; Lin et al., 2018; Hoyer et al., 2019; Sosnovik & Oseledets, 2019; Zhang et al., 2019; Abueidda et al., 2020; Kollmann et al., 2020; Rade et al., 2020; Wang et al., 2021; Xue et al., 2021; Zhang et al., 2021a |
| | Long Short-Term Memory | Optimization | Kallioras & Lagaros, 2021b; Lew & Buehler, 2021; Qiu et al., 2021a |
| | Support Vector Machine | Postprocessing | Chu et al., 2016; Strömberg, 2019; Karlsson et al., 2020; Strömberg, 2020 |
| | | Prediction | Xia et al., 2017 |
| | Others | Prediction | Zhou & Saitou, 2017; Lynch et al., 2019; Keshavarzzadeh et al., 2020 |
| **Unsupervised Learning** | Generative Adversarial Network | Exploration | Oh et al., 2018; Rawat & Shen, 2018; Jiang et al., 2019; Li et al., 2019a; Oh et al., 2019; Rawat & Shen, 2019a; Rawat & Shen, 2019b; Sharpe & Seepersad, 2019; Shen & Chen, 2019; Wen et al., 2019; Yu et al., 2019; Almasri et al., 2020; Greminger, 2020; Malviya, 2020; Wen et al., 2020; Li et al., 2021; Nie et al., 2021; Sim et al., 2021 |
| | Auto Encoder | Exploration | Guo et al., 2018; Ha et al., 2020; Yamasaki et al., 2021 |
| | Principal Component Analysis | Exploration | Ulu et al., 2016; Lei et al., 2019 |
| | | Prediction | Li et al., 2019c |
| | Clustering | Optimization | Liu et al., 2015; Jiang et al., 2020; Kumar & Suresh, 2020; Qiu et al., 2021b |
| **Semi-Supervised Learning** | Active Learning | Optimization | Cang et al., 2019 |
| | Modified Networks | Optimization | Gillhofer et al., 2019; Jiang & Fan, 2019a; Jiang & Fan, 2019b; Wang et al., 2020; Blanchard-Dionne & Martin, 2021; Kudyshev et al., 2021 |
| | Deep Belief Network | Optimization | Kallioras et al., 2020; Kallioras & Lagaros, 2020; Kallioras & Lagaros, 2021a |
| **Reinforcement Learning** | Value-based | Optimization | Hayashi & Ohsaki, 2020; Brown et al., 2022 |
| | Policy-based | Exploration | Sun & Ma, 2020; Jang et al., 2020 |

### 3.1.1. Supervised Learning

For supervised learning, NNs and CNNs were used the most, and various other methodologies, such as LSTMs, SVMs, SVRs, KNNs, random forest regression, and GP, were applied.

*a) NN*

The application of ANN or DNN (both referred to as NN) to TO, is distinguished for the purpose of prediction of features or performance and design optimization.

*a1) Prediction*

First, researchers have conducted optimization by predicting design parameters constituting the shape by using NN. The design parameters predicted by Deng and To (2020) correspond to DRL design variables that represent density. The number of design parameters was reduced by converting the existing density field to DRL design variables through NN. The design parameters predicted by Keshavarzzadeh et al. (2021) correspond to the DNSM coefficient and are predicted to map DNSM's coefficient for design parameters (load and boundary conditions and volume fraction) via NN through a DNSM metric model. In addition, in the work of Yildiz et al. (2003), the design parameters correspond to the class for the hole shape. Yildiz et al. (2003) predicted which class the hole of the shape corresponds to for the optimized feature and performed TO based on the results.

Second, in studies predicting the performance of a shape (Guo et al., 2021), NN is used to replace some calculations in the TO process. Deng et al. (2020) used DNN to replace the calculation of the objective function during the optimization process. Chi et al. (2021) and Aulig and Olhofer (2014) replaced the sensitivity calculation in the optimization process, while Zheng et al. (2021) and White et al. (2019) allowed the mapping of objective function to parameters defining microscale meta-materials. Bujny et al. (2018) predicted the performance for topology change by mapping graph (structural information) to performance through NN, and Patel and Choi (2012) used PNN to predict the feasibility in terms of safety or risk in the TO process to determine the convergence of optimization.

*a2) Optimization*

Several studies also train NN for design optimization purposes. Halle et al. (2021) predicted the optimized design by using design parameters as inputs, and Kim et al. (2021) optimized the periodic material structure by replacing the process of obtaining the elasticity tensor for the design parameter with DNN in the FEA process to effectively utilize RVE. Chandrasekhar and Suresh (2021c), Chandrasekhar and Suresh (2021a), Chandrasekhar and Suresh (2021b), and Zehnder et al. (2021) optimized the topology density of the corresponding coordinates. In this case, topology prediction becomes possible by reflecting FEA in the loss function. Napier et al. (2020) approximated the high-dimensional (fine) shape with a low-dimensional (coarse) shape, and Bi et al. (2020) predicted and updated the gradient of the shape through the optimization history.

*b) CNN*

When CNN is applied to TO, it is advanced in terms of the purpose of prediction for utilizing images rather than NN. Similar to NN, CNN is mainly used for the purpose of prediction of the performance of shape and design optimization.

*b1) Prediction*

For the purposes of predicting the performance, Lee et al. (2020), Sasaki and Igarashi (2019a), Sasaki and Igarashi (2019b), Doi et al. (2019), Asanuma et al. (2020), and Li et al. (2019b) predicted the value of the objective of constraint function by CNN. Among them, Li et al. (2019b) provided a criterion for sensitivity analysis by predicting whether the shape is damaged or not. Takahashi et al. (2019) also replaced sensitivity calculations via CNN. Gaymann and Montomoli (2019) determined a topology with better performance among a given topology (player) as a winner.

*b2) Optimization*

For the design optimization purposes, Abueidda et al. (2020), Zhang et al. (2019), and Kollmann et al. (2020) derived the optimal design by inputting the design conditions (load and boundary conditions, etc.). Sosnovik and Oseledets (2019) and Lin et al. (2018) predicted the final optimized design by using the gradient between the n-th iteration shape and the gradient between n-th and (n-1)-th iteration. Banga et al. (2018) derived the final optimized design density by combining design conditions and FEA information. Rade et al. (2020) used the CDCS method to train three encoder–decoder-shaped CNN models. The compliance prediction model (CPM) could predict the next compliance by using the inputs of initial compliance and current density. The density prediction model (DPM) could also predict the next density by using the inputs of current compliance and current density. During the five loops, the two models utilized each other's outputs as inputs. Finally, with the output density determined after five loops, final density prediction model (FDPM) predicts optimal density. In addition, Hoyer et al. (2019) updated the objective function for the predicted design via the deep image prior to the implementation of CNNS, reflected this information using gradient-methods, and allowed the design to be generated to satisfy the objective function. Similarly, Zhang et al. (2021a) proposed the TONR method and introduced physics information into the loss function. With the noise input, the model could predict the quasidensity and conduct FEA to reflect compliance in the loss function. Finally, Xue et al. (2021) and Wang et al. (2021) used CNN to upscale optimal designs.

*c) LSTM*

LSTM is a useful method for predicting a series of processes (e.g., the TO process). It is often used in conjunction with other methodologies for TO optimization.

*c1) Optimization*

Qiu et al. (2021a) utilized the LSTM model to predict the next sequence of TO and then performed DLTO integrated with ESO. An initial 1st–10th sequence image was predicted via CNN(U-net). From the 11th sequence, optimization was performed by predicting the shape of the next sequence, in which the shapes of the previous 1st to 10th sequences were inputted. Similarly, Rade et al. (2020) used DS method with CNN-LSTM and SIMP method to predict the density of subsequent sequences with the input shapes of the initial two iterations. Lew and Buehler (2021) trained a 2D convolutional VAE with the shapes of TO sequences to encode an image into a two-dimensional latent vector. With two sets of initial latent vectors, the LSTM model could predict the time series latent vectors for the subsequent 147 sequences. Kallioras and Lagaros (2021b) used LSTM for the time-series classification for generative design. With multiple mesh domain, after 20 iterations of SIMP, the density values of each element were input into the LSTM model to derive the final density class of every element of the FE discretization. Then, multiple image filters were applied to the output shape to generate various optimized shapes. Finally, 20 SIMP iterations were performed in the final desired mesh size.

*d) SVM*

SVMs are also often used for postprocessing and prediction in TO.

*d1) Postprocessing*

Chu et al. (2016), Strömberg (2019), Strömberg (2020), and Karlsson et al. (2020) discretized the boundary surfaces of designs by using SVMs. The design domains could be classified into two sets. Separating hypersurfaces were used to maximize the distance from the boundary to the closest point of each set. This approach enables the designs to converge with smooth surface boundaries and it can be manufactured. Furthermore, Strömberg (2019), Strömberg (2020), and Karlsson et al. (2020) performed FEA and metamodel-based design optimization (MBDO). For the MBDO, they used an ensemble of metamodels, including quadratic regression, Kriging, radial basis function networks, polynomial chaos expansion, polynomial regression models and support vector regression models). Their proposed schemes were using genetic algorithms (GAs) followed by sequential quadratic programming (SQP). In particular, Strömberg (2019) and Karlsson et al. (2020) considered the use of lattice structures for AM.

*d2) Prediction*

Xia et al. (2017) used SVM to predict sensitivity for the purpose of removing the checkerboard pattern while optimizing the process.

*e) Others*

*e1) Prediction*

Among the other researchers that use supervised learning methodologies, the work of Keshavarzzadeh et al. (2020) applied the GP to uncertainty-based stress-based TO. On the basis of the low-resolution FEA data, they constructed a GP kernel and trained a GP emulator to predict maximum stress values and sensitivity for data points in design domains through high-resolution FEA data. At this time, high-resolution and low-resolution data points are extracted through Monte Carlo search. Lynch et al. (2019) recommended the tuning of parameters by measuring the similarity between the current optimization problem and the optimization problem of the previous database and the meta-learning approach (potential tuning parameters for the cantilevered beam SIMP TO problem: weight of objective functions, mesh size, constraint scaling, density initial values, SIMP exponent penalty, objective tolerance, maximum number of forward problem evaluations, minimum allowable stiffness, meshing method/element type). By using parameters defined through meta-learning, the parameters are tuned through Bayesian optimization in the optimization progression (meta-modeling). Zhou and Saitou (2017) approximated the objective function for the optimal design through an ensembled random forest regression model.

### 3.1.2. Unsupervised Learning

The use of unsupervised learning for TO usually heavily utilizes the generative model and generally has a design exploration purpose similar to the original purpose of the generative model methodology.

*a) GAN*

*a1) Exploration*

Greminger (2020) generated feasible data by using a 3D GAN. Li et al. (2021) evaluated the performance of designs generated by GAN through FEM, and Sim et al. (2021) clustered the designs generated by GAN model into k-means clustering. Besides simply generating shapes, Oh et al. (2018) and Oh et al. (2019) refined the design through the process of re-optimizing the GAN-generated structures as an initial design. Rawat and Shen (2018) and Rawat and Shen (2019a) used GAN to generate structures and then replace FEA with meta-models to predict design characteristics over CNN.

As a manner of reflecting the design features or performance during training, Rawat and Shen (2019b), Nie et al. (2021), Shen and Chen (2019), Malviya (2020), Sharpe and Seepersad (2019), and Almasri et al. (2020) added features, such as design constraints and load and boundary conditions, as the conditions to GAN to generate designs under the influence of features (conditions) during training.

Jiang et al. (2019), Wen et al. (2019), and Wen et al. (2020) utilized design constraints as the input conditions in the GAN, while simultaneously evaluating performance on generated designs and reflecting performance in training to generate performance-good designs. In this case, Wen et al. (2019) and Wen et al. (2020) derived high-resolution designs by using PGGAN. Yu et al. (2019) and Li et al. (2019a) also used GAN for upscaling purposes for low-resolution designs predicted by CNN.

*b) AE*

*b1) Exploration*

There are also quite a few methodologies using AEs, another representative methodology of unsupervised learning. In the work of Ha et al. (2020), AE generates a left, right robot gripper that fits the optimized designs and 3D CNN model predicts its performance. Yamasaki et al. (2020) implemented the re-training process using high-performance designs among the data generated using VAE, and Guo et al. (2018) combined style transfer with the VAE so that the designs generated in latent space resemble the original design and are feasible.

*c) PCA*

As another unsupervised learning methodology, the method using PCA is distinguished for the purpose of design exploration and prediction.

*c1) Exploration*

Lei et al. (2019) used PCA for parameter optimization used in MMC TO. Through PCA dimension reduction, they get principal components of MMC variables and conduct SVR or KNN regression to obtain a new design. Ulu et al. (2016) trained a meta-model that maps the principal component of the PCA dimensionally reduced space for the optimal shape and the load configuration corresponding to the design parameters of the shape. Accordingly, it is possible to predict the principal components of PCA for the new design for the new loading configuration and derive a new design.

*c2) Prediction*

Li et al. (2019c) sampled data in a low-dimensional space gained through PCA, obtained the performance of the sample data numerically, and built a meta-model for the performance of the sample data to proceed with optimization.

*d) Clustering*

*d1) Optimization*

As a methodology of using clustering algorithm, scholars have added clustering for shapes, design characteristics, or FEM results to an iterative TO step to reduce the calculation costs (Kumar & Suresh, 2020; Qiu et al., 2021b; Liu et al., 2015; Jiang et al., 2020). Among them, Jiang et al. (2020) trained an ensembled Extra-Trees surrogate model that determines whether it is feasible through the results of clustering.

### 3.1.3. Semi-Supervised Learning

Traditional semi-supervised learning is the branch of ML concerned with the use of labeled data as well as unlabeled data to perform certain learning tasks; it is conceptually situated between supervised and unsupervised learning (Van Engelen & Hoos, 2020). This method includes the field of active learning, in which the learning algorithm can ask queries for the labels of previously unlabeled data points to be labeled by an oracle (e.g., a human annotator). Learning algorithms can achieve much greater accuracy with fewer labeled training data (Settles, 2012). However, in some cases, some algorithms are difficult to define as supervised or unsupervised learning. These configurations are conceptually situated between supervised and unsupervised learning because they combine the concepts of both algorithms, using them in parallel (e.g., combining GANs or AEs with NNs (McDermott et al., 2018)) or step by step (e.g., modifying GANs, DBNs). Accordingly, we consider both traditional semi-supervised learning and models that combine the concepts of supervised and unsupervised learning as semi-supervised models. From this point of view, semi-supervised learning is a useful methodology for design generation while ensuring good performance. All of the studies introduced in this section were aimed at design optimization. Some methods of training were achieved by active learning, modifying traditional networks (e.g., combining GANs or AEs with NNs or modifying existing GANs) and DBNs (combination of supervised and unsupervised learning steps).

Cang et al. (2019) conducted active learning with adaptive sampling. By providing design parameters to NN, they could train the model, obtain near-optimal solutions and find new data points to compute corresponding optimal designs. In this manner, the model could compute the design with a less expensive computational budget.

Wang et al. (2020) used a combined model of VAE and regression NN to allow latent space to contain physical properties. Similarly, Gillhofer et al. (2019) and Blanchard-Dionne and Martin (2021) proposed a methodology for training to generate performance-satisfactory designs by changing the label of the discriminator to reflect it on loss or by attaching a forward network. Kudyshev et al. (2021) used an AAE, which combines conditional-AAE with NN to create a design containing physical information. Jiang and Fan (2019a) and Jiang and Fan (2019b) trained the model to approximate the optimal design by reflecting the performance of design generated using the generative NN in loss function to reflect the physical information. Further that, Kallioras et al. (2020), Kallioras and Lagaros (2020), and Kallioras and Lagaros (2021a) proposed methods that combined DBNs and the SIMP method to refine on similar predicted features.

#### 3.1.4. Reinforcement Learning

Reinforcement Learning (RL) has a high potential for generating optimal designs, but the research on the application of RL to TO remains to be insufficient. For design optimization, value-based RL is used. Hayashi and Ohsaki (2020) derived designs corresponding to design parameters on a graph basis by using Q-learning (Watkins and Dayan, 1992). Brown et al. (2022) reformulated TO problems to a sequential RL task, a Markov decision process (MDP) (Bellman, 1957), and optimized elementally discretized 2D topologies by using the double-deep Q-learning algorithm (Van Hasselt et al., 2016).

For design exploration, Sun and Ma (2020) used the e-greedy policy to disturb the search direction, UCB to add a sensitivity feature, TS and IDS to direct the search. The optimal shape changes were compared on the basis of the parameter changes. Jang et al. (2020) conducted a generative design method on designs to explore and eventually maximize design diversity by using the proximal policy optimization (PPO) algorithm (Schulman et al., 2017).

## 3.2. Data

In TO, ML is used for various purposes, as mentioned in Section 2. The input type of data also varies depending on the purpose of the ML model. The form of the input may be characterized according to the form of use of the ML model.

Design parameters input to derive features that meet the design constraints, intermediate optimization design input to accelerate TO or FEA meta-modeling, final optimization or structural analysis input to predict the performance, and FEA meta-modeling, design generation, and coordinate information input to derive the optimal density value as a means of coordinating the various combined information input. This section reviews the DLTO frameworks focusing on DL input.

#### 3.2.1. Design Parameter (Condition)

The design parameter corresponds to condition information for designing. For example, the value of the load and boundary conditions, the value set as constraints in the TO process, such as volume fraction and mass, or the coordinate information, are included, and the design parameter is variously defined according to the applications. When design parameters are used as input, the model is trained to the optimal design shape to satisfy the design parameter settings, and it can be divided into three methods. First, a prediction method was used to predict the performance corresponding to the design parameter. Second, a design optimization method was implemented to map the information in the design parameters to the optimal geometry. Finally, a design exploration method was designed to create new geometry under the influence of design parameters.

*a) Prediction*

As a study to predict the performance corresponding to the design parameters, White et al. (2019) predicted the performance of parameters that present the meta-material to the microscale as an input. On this basis, sensitivity analysis of the optimization of the macroscale is performed. Kim et al. (2021) predicted effective elasticity tensor using the characteristic parameter of the material by using RVE methodology as an input. Thereafter, the FEA process may be replaced, and the structure that considers the periodic structure may be optimized.

*b) Optimization*

In the studies for design optimization, Cang et al. (2019), Abueidda et al. (2020), Kollmann et al. (2020), Halle et al. (2021), and Malviya (2020) predicted the design with design parameters as inputs for non-iterative optimization. Keshavarzzadeh et al. (2021) generates design through the DNSM parameters after training a model that maps design parameters to DNSM parameters for non-iterative optimization. To carry out sequential optimization process, Lew and Buehler (2021) used latent vectors as design parameters. Lynch et al. (2019) recommended the tuning of parameters for the corresponding design problem during optimization and uses Bayesian optimization to tune the parameters and proceed with optimization. Lei et al. (2019) used PCA dimension reduction for design parameters to obtain eigenvector and eigenvalue. Using this information, the design is derived through nonlinear regression. Ulu et al. (2016) predicted the PCA dimensionally reduced value of the topology by using the design parameter (i.e., loading configuration) as an input. When a new design parameter is introduced, the shape corresponding to the predicted PCA value is derived.

*c) Exploration*

As for the studies on design exploration, Jiang and Fan (2019b), Jiang et al. (2019), Wen et al. (2019), Rawat and Shen (2019b), Shen and Chen (2019), Almasri et al. (2020), and Sharpe and Seepersad (2019) trained generative models that generate a new design by entering design parameters. Yu et al. (2019) and Li et al. (2019a) generated a low-resolution image by inputting a design parameter and upscale it into a GAN. Hayashi and Ohsaki (2020) selected design parameters with initial environment and derived the graph features through Q-Learning.

### 3.2.2. Intermediate Optimization Result

When the intermediate optimization result is used as the input, it is divided into a method for prediction and a method for design optimization from that point.

*a) Prediction*

The purpose of Prediction is primarily meta-modeling. DL models replace calculation of TO process (calculation of objective function or sensitivity, etc.) to accelerate TO. Lee et al. (2020), Deng et al. (2020), Sasaki and Igarashi (2019a), Sasaki and Igarashi (2019b), Doi et al. (2019), Asanuma et al. (2020), and Zheng et al. (2020) proposed surrogate models that replace objective functions, such as performance with DL models, by inputting intermediate optimization results for each application. Takahashi et al. (2019) utilized intermediate optimization results as the input to replace sensitivity calculations with CNNs, while Xia et al. (2017) removed checkerboard patterns by replacing the process of calculating sensitivity with inputting information from specific data points at n-th iteration to SVM. Gaymann and Montomoli (2019) focused on a game format in which the design means player, and is used as input. They predicted the winner and the end of the game, repeated the process of determining the next movement through Monte Carlo search, and optimized the fluid structure that could minimize the pressure difference. Keshavarzzadeh et al. (2020) predicted the maximum stress value and sensitivity for data points of unknown designs through GPemulator. At this time, a covariance kernel affecting the GPemulator is constructed through low-resolution FEA data.

Patel and Choi (2012) performed a process of determining whether the region exceeds the safety criterion by inputting each data point when the iteration is repeated. Jiang et al. (2020) preprocessed the shape with representing features(called visual word vectors in the study) during the optimization process and quickly determined whether the shape is feasible through the corresponding vectors.

*b) Optimization*

In a study for design optimization, Sosnovik and Oseledets (2019) and Lin et al. (2018) predicted the final design by inputting the intermediate density distribution of the design in the middle n-th iteration and the gradients between the n-th and n-1th iteration to a model for acceleration of TO. Kallioras et al. (2020), Kallioras and Lagaros (2020), Kallioras and Lagaros (2021a), and Kallioras and Lagaros (2021b) predicted their respective designs by using the density fluctuation pattern of the finite element discretization provided by the initial steps of the SIMP method. Although Hoyer et al. (2019) and Zhang et al. (2021a) used noise as an input, not the intermediate optimization results, it determines the performance of the shape generated from the noise and optimizes the design by allowing the performance to affect network parameter updates with the backwards pass. Liu et al. (2015) classified the shape of intermediate iteration into k number of materials through K-means clustering. On this basis, optimization is re-progressed and obtained optimal material distribution.

### 3.2.3. Final Optimization Result

Studies with the final shape as an input can be largely divided into three purposes: the purpose of Prediction, the purpose of design exploration, and the purpose of postprocessing.

*a) Prediction*

In studies for performance prediction purposes, Rawat and Shen (2018) and Blanchard-Dionne and Martin (2021) predict the parameters or performance of the shape based on that generated with the generative model trained with the optimal designs. Bujny et al. (2018) predicted the performance ranking of the node–edge pair of the graph representing the shape as an input graph. This method accelerates optimization by creating a predictor for shape changes.

Li et al. (2019c) dimensionally reduced the dimension of the optimal shape by using logistic PCA and sampled data in the latent space. Then, the shape is obtained through inverse transformation, and then a surrogate model is trained to predict the performance of the shape. Zhou and Saitou (2017) predicted the performance (resin filling time) corresponding to the input of the resin filling data generated using the KLS.

*b) Exploration*

In studies aimed at design exploration, Guo et al. (2018) trained VAE with optimization features as inputs and to find new designs in latent space. To create a better latent space, Wang et al. (2020) determined the optimization features by adding a regressor to the VAE, and predicted the performance of the design when latent variables are input through the regression model. Through this approach, latent space can be formed to generate designs with good performance. Yamasaki et al. (2021) generated designs by using VAE, to generate high-performance shape, re-training is conducted using high-performance generated data. In addition, studies that generate new designs with GANs are trained using shapes generated from noise and final optimization results. Li et al. (2021), Jiang and Fan (2019a), Sim et al. (2021), Greminger (2020), Oh et al. (2018), Oh et al. (2019), and Gillhofer et al. (2019) performed studies in which a discriminator is trained to distinguish optimal shapes and generated shapes by the generator without any condition. Oh et al. (2019) further trained AE to evaluate design originality through reconstruction error, and Gillhofer et al. (2019) trained the discriminator to classify performance criteria rather than simply focusing on real/fake classification, allowing the generator to generate better performance features.

Jang et al. (2020) conducted PPO by using AE as an environment pretrained by final optimization results. At this time, the agent network generates a wheel and gives its parameters as action, and the environment outputs reward and state. Furthermore, Ha et al. (2020) generated left and right robot grippers suitable for the optimized 3D shape by using the optimized 3D shape as input and predicted their performance.

*c) Postprocessing*

In the postprocessing fields, Strömberg (2019), Strömberg (2020), Chu et al. (2016), and Karlsson et al. (2020) processed the boundaries through SVM when optimization designs are input. Moreover, Xue et al. (2021), Wang et al. (2021), and Napier et al. (2020) predicted high-resolution shapes when optimal low-resolution shapes are input. In addition, Yildiz et al. (2003) predicted the shape of the hole (triangle, trapezoid, rectangle, etc.) of the design by using topology optimized shapes as input, thereby enabling appropriate shape optimization to proceed.

### 3.2.4. Structural Information (FEA Information)

The two main purposes in which FEA information is used as input are prediction and design optimization.

*a) Prediction*

As a study aimed at FEA meta-modeling, Aulig and Olhofer (2014) replaced sensitivity calculations with predictive models by entering density FEA information (displacement) of shapes, and Bi et al. (2020) used high-fidelity FEM simulation data to predict the gradient of shape in each iteration, replacing gradient calculation, accelerating TO. Chi et al. (2021) conducted an online training of TO and predicted the sensitivity information by using the FEA value of the coarse data and the filtered design variables of fine data of the same shape as inputs for TO. Furthermore, Li et al. (2019b) used the FEA value as an input to determine whether the shape is damaged and to determine whether the sensitivity analysis proceeds.

*b) Optimization*

As design optimization studies, Zhang et al. (2019) and Nie et al. (2021) predicted and generated optimal shapes accordingly by inputting both design parameter information and FEA information. Qiu et al. (2021b) utilized FEA information as the input, and this FEA information was clustered to determine the internal shape structure, and Kumar and Suresh (2020) performed clustering with the density of the shape and the FEA information input to improve the performance.

### 3.2.5. Coordinates in Design Space

The two purposes related to the inputting of the coordinate value of the design domain are the design optimization purpose and the meta-modeling purpose. Chandrasekhar and Suresh (2021c), Chandrasekhar and Suresh (2021a), Chandrasekhar and Suresh (2021b) predicted the density value of the coordinates, and the optimal shape is derived through minimization of loss function. At this time, the loss function contains FEA to satisfy the TO settings. Zehnder et al. (2021) also used data points as input and subsequently trained the displacement network by minimizing the total potential energy of the system and updating the displacement. Then, sensitivity analysis was performed to compute the density–space gradients, which, after applying our sensitivity filtering, resulted in target density fields. Finally, the density was updated by minimizing the loss function between the current and target densities. Deng and To (2020) predicted the DRL design variables that represent voxel shapes. Through this approach, the design variable was reduced to accelerate optimization.

#### 3.2.6. Multimodal Input

The study of multimodal input data refers to studies that use various input data for a model. The works of Zhang et al. (2019), Nie et al. (2021), Aulig and Olhofer (2014), and Chi et al. (2021) mentioned in Section 3.2.4 are exactly included in this section, but they are not elaborated in the previous section to highlight examples with FEA information. Brown et al. (2022) provided the design parameters (load and boundary conditions) and the FEA information (stress) to the RL agent to take a new action. Banga et al. (2018) predicted the optimal designs by inputting design parameters and intermediate optimization results for the purpose of deriving optimal designs. Qiu et al. (2021a) also used the design parameter (load condition) and the shapes of previous iterations as inputs. Rade et al. (2020) implemented two methodologies, DS and CDCS prediction, as mentioned in Section 2.2, in which the DS methodology was used to predict optimal designs by using initial design parameters and FEA information. The CDCS methodology can predict optimal designs through intermediate optimization results and intermediate FEA information. Qian and Ye (2020) replaced compliance and sensitivity calculations by inputting design parameters and intermediate optimization results for meta-modeling purposes.

### 3.3. Loss Function

The form of loss function of DL model in TO is defined as follows depending on the objectives to achieve through DL models.

- Difference between generated and ground truth shape
- Difference between predicted and ground truth performance
- Add performance value/gradient obtained from FEA
- Add additional constraints
- Add two types of loss function for multiple models

#### 3.3.1. Difference between generated and ground truth shapes

To minimize the difference between generated and ground truth shapes, Abueidda et al. (2020) and Kollmann et al. (2020) calculate the difference between the shape predicted using the NN and the ground truth shape in quadratic norm form. Xue et al. (2021) uses mean squared error (MSE) loss to calculate the difference between high-resolution generated shape and real shape to map low-resolution shape to high-resolution shape. Furthermore, Keshavarzzadeh et al. (2021) predicted multiresolution TO results, including low-resolution predictive shape and combining MSE loss function for low-resolution shapes and high-resolution shapes. Lastly, because Qiu et al. (2021a) used binary data (i.e. ESO method), they chose binary cross entropy (BCE) loss.

#### 3.3.2. Difference between Predicted and Ground Truth Performance

To minimize the difference between predicted and ground truth performance, Lee et al. (2020) used MSE and MAPE as loss function to calculate the difference between the predictive and real values of objective function and constraints in the optimization. White et al. (2019) defined the loss function in combination with the quadratic loss form of the difference of the objective function and the derivative component of the objective function. Gaymann and Montomoli (2019) also used the quadratic loss form to approximate the objective function of the optimization process. Patel and Choi (2012) used a quadratic loss function to determine the feasibility of the shape in the TO process.

Gillhofer et al. (2019) used BCE to determine the low or high response region. The weight was adjusted for shapes with a high response higher than the threshold value. In this manner, the generator can generate a shape of the low response region.

#### 3.3.3. Adding Performance Value/Gradient Obtained from FEA

The studies aimed at adding performance value/gradient objective from FEA defer the use of performance data as the label, but they add a physics term to the loss function itself to satisfy the objective function, as shown in Method 2 in Figure 23. Chandrasekhar and Suresh (2021c), Chandrasekhar and Suresh (2021a), and Chandrasekhar and Suresh (2021b) defined the loss function to reflect the objective function and constraint values calculated through FEA of shape in each iteration to optimize the structure. Jiang and Fan (2019a), Jiang and Fan (2019b), Qian and Ye (2021), Hoyer et al. (2019), and Zhang et al. (2021a) also incorporated the performance values obtained through FEA into loss function and repeat the process of updating the variables by calculating gradients through backpropagation to optimize the shape.

#### 3.3.4. Integrating Additional Constraints

The loss function to add additional constraints is defined in the form of adding a specific constraint loss term to the existing loss. Zhang et al. (2019) used Kullback–Leible (KL) divergence (Kingma & Welling, 2013) to obtain the difference between the prediction and the real shapes corresponding to Section 3.3.1, and further defines additional loss in the form of L2 regulation term to prevent overfitting. Guo et al. (2018) added a reconstruction error corresponding to standard VAE loss, a style loss (Gatys et al., 2015), which evaluates how well the generated shape match the style of the real shape, and a mode-collapse loss (Zhao et al., 2016) to prevent only similar samples from being generated. Nie et al. (2021) defined BCE of standard GAN loss, L2 loss term between real and fake shapes corresponding to Section 3.3.1 and further added an absolute error term between the target volume and the volume of predicted shape. Sosnovik and Osledets (2019), Lin et al. (2018), and Banga et al. (2018) defined the BCE loss to determine the difference between predicted and real shapes (Section 3.3.1) and add MSE loss for the volume constraint term to the loss.

#### 3.3.5. Adding Two Types of Loss Functions for Multiple Models

In studies involving two or more networks, the use of more than two loss functions for each network is reasonable. Rawat and Shen (2018) and Rawat and Shen (2019a) generated shapes by WGAN by using Wasserstein distance (Section 3.3.1) and predicted the parameters of the generated shapes by using the MSE loss (Section 3.3.2). Similarly, Ha et al. (2020) used MSE loss for a generative network to generate shapes and used BCE to determine the feasibility of generated shapes through 3D CNN. Li et al. (2019a) used standard GAN loss to generate low-resolution shapes over GAN and added the MSE loss between real, predicted shapes in SRGAN for upscaling (Section 3.3.1). Yu et al. (2019) selected mean absolute error (MAE) loss of real and predicted low-resolution shapes on CNN network (Section 3.3.1) and added MAE loss term on the standard GAN loss to resemble the real shape on the cGAN network (Section 3.3.1). Li et al. (2019b) used MSE loss to map macroscale strain and macroscale stress (FFNN) and predict the feasibility of microstructure with a stress value (Section 3.3.2). Rade et al. (2020) utilized a different type of loss for two methodologies. The study proposed a DS method with MSE loss to calculate the differences between real and predicted shapes (Section 3.3.1) and a CDCS prediction network with loss that combines MAE loss that calculates the difference between real and predicted shapes in the early stages (Section 3.3.1) while the BCE loss used to predict compliance in terms of optimal design convergence (Section 3.3.2).

### 3.4. Physics

FEA is used in DLTO to ensure a much more efficient convergence of optimal designs and produce meaningful results. This concept is similar to a physics-guided NN (PGNN)and a physics-informed NN (PINN) methodologies, which have recently emerged in engineering fields, and represent a method for allowing DL models to learn the physical meaning of the data (Zhuang et al., 2021a). For DLTO problems, FEA information is reflected largely in three ways (Figure 23):

- Method 1. FEA as a method of evaluating performance during the training process
- Method 2. FEA as a term of loss function during the training process
- Method 3. FEA as an information to use as input when training

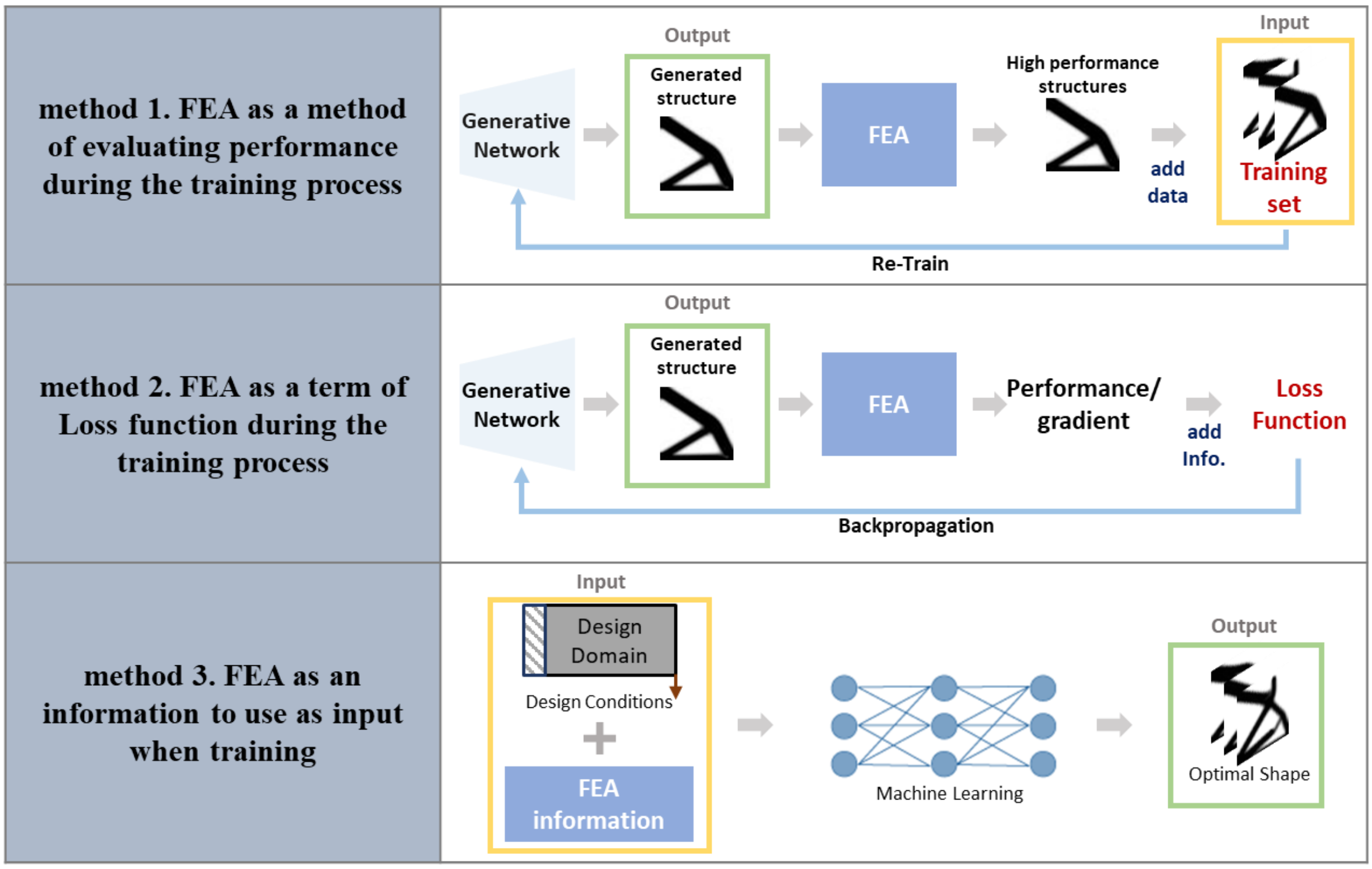

**Figure 23 Methods of using FEA information for training models**

### 3.4.1. FEA as a method of evaluating performance in the training process

Method 1 uses FEA as a performance evaluation tool for the result shapes derived through the trained model. Then, the dataset is reconstructed with the generated data that is determined to be good through FEA and re-training is conducted on the reconstructed dataset, allowing the model to predict the shape with good performance (Wen et al., 2019; Wen et al., 2020; Jiang et al., 2019; Halle et al., 2021; Yamasaki et al., 2021). Thus, FEA is used as a performance evaluation tool to reinforce the performance of generated data.

### 3.4.2. FEA as a loss function term in the training process

Method 2 reflects FEA information into a loss function during training. The loss function reflecting FEA information converges in the direction of minimization. In other words, if the FEA information is used as a loss function, then the model is trained in the direction of increasing performance through backpropagation for the corresponding FEA loss term (Hoyer et al., 2019; Chandrasekhar & Suresh, 2021c; Chandrasekhar & Suresh, 2021a; Chandrasekhar & Suresh, 2021b; Zhang et al., 2021a; Zehnder et al., 2021; Jiang & Fan, 2019a; Jiang & Fan, 2019b).

### 3.4.3. FEA as an information for use as input during training

Method 3 allows the model to be trained with the FEA information itself by inputting FEA information as an input of the model. Nie et al. (2021) and Zhang et al. (2019) predicted the features by putting basic design conditions and FEA information together as inputs. Bi et al. (2020) used only the data obtained through FEA as an input to predict the gradient of the shapes during TO, and Chi et al. (2021) predicted the sensitivity by using FEA information of low-resolution shapes to optimize high-resolution shapes. Rade et al. (2020) presented two methodologies by using [design parameter + FEA information], [intermediate optimization result + FEA information] to predict optimal shapes and the methodology presented by Kudyshev et al. (2020) also used design parameters and physical information through FEA to generate data. As a study considering microscale, Qiu et al. (2021b) performed clustering with only FEA information, and Kumar and Suresh (2020) used density and FEA information of shape to determine microstructure similar to that in the work of Li et al. (2019b) that considered micro/macroscales, predicts damage by using FEA information on macro/microstructures.

## 3.5. Applications and Generalization

The application fields of DLTO are diverse and are gradually expanding. This section reviews application examples and generalization cases of this methodology.

### 3.5.1. Application

Studies on DLTO is a representative application such as beam, bridge, truss, nanophotonics, meta-surface, wheel, motor, heat conduction system, and fluid structure, etc. Until now, studies on 2D shapes have been dominated. It is gradually being used in various fields, and 3D shape optimization research is also being conducted (Table 3).

**Table 3 DLTO applications**

| Scale | Applications | Studies | Examples |
|---|---|---|---|
| ***2D*** | 2D Design Domain (Beam, Bridge, Truss, Bracket, etc.) | Patel & Choi, 2012; Aulig & Olhofer, 2014; Liu et al., 2015; Chu et al., 2016; Ulu et al., 2016; Zhou & Saitou, 2017; Bujny et al., 2018; Rawat & Shen, 2018; Cang et al., 2019; Gillhofer et al., 2019; Hoyer et al., 2019; Lei et al., 2019; Li et al., 2019a; Lynch et al., 2019; Rawat & Shen, 2019b; Sharpe & Seepersad, 2019; Shen & Chen, 2019; Sosnovik & Oseledets, 2019; Takahashi et al., 2019; Yu et al., 2019; Zhang et al., 2019; Abueidda et al., 2020; Almasri et al., 2020; Bi et al., 2020; Deng et al., 2020; Halle et al., 2021; Hayashi & Ohsaki, 2020; Jiang et al., 2020; Kallioras et al., 2020; Kallioras & Lagaros, 2020; Keshavarzzadeh et al., 2020; Lee et al., 2020; Malviya, 2020; Napier et al., 2020; Qian & Ye, 2020; Kallioras & Lagaros, 2021a; Kallioras & Lagaros, 2021b; Lew & Buehler, 2021; Nie et al., 2021; Sim et al., 2021; Wang et al., 2021; Yamasaki et al., 2021; Qiu et al., 2021a; Zhang et al., 2021a; Zehnder et al., 2021; Brown et al., 2022 | 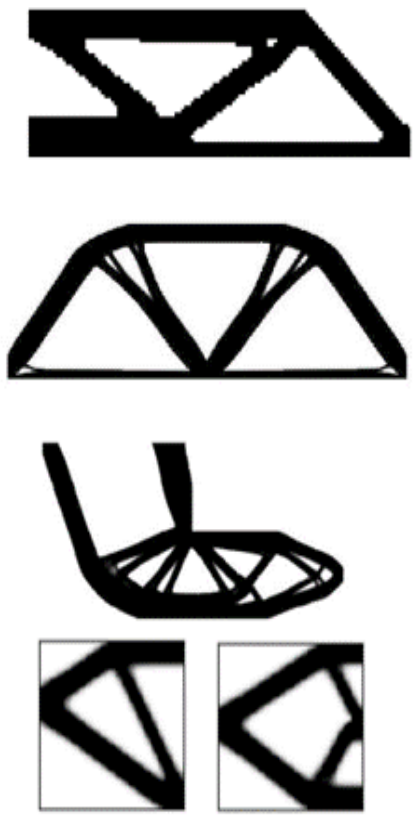 |
| | Nanophotonics, Spinodoid, Metasurface, Periodic structure | Jiang & Fan, 2019a; Jiang & Fan, 2019b; Jiang et al., 2019; Li et al., 2019b; Li et al., 2019c; Wen et al., 2019; Kollmann et al., 2020; Kumar & Suresh, 2020; Wang et al., 2020; Wen et al., 2020; Blanchard-Dionne & Martin, 2021; Kim et al., 2021; Kudyshev et al., 2021; Li et al., 2021; Qian & Ye, 2021; Qiu et al., 2021b | 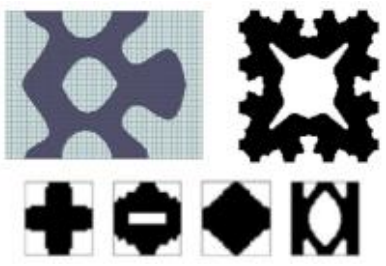 |
| | Wheel | Oh et al., 2018; Oh et al., 2019; Jang et al., 2020 | 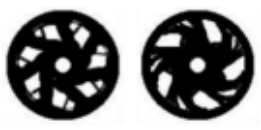 |
| | Motor | Doi et al., 2019; Sasaki & Igarashi, 2019a; Sasaki & Igarashi, 2019b; Asanuma et al., 2020 | 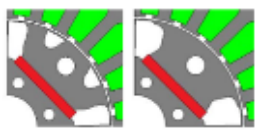 |
| | Heat Conduction | Guo et al., 2018; Lin et al., 2018; Li et al., 2019a; Deng et al., 2020; Keshavarzzadeh et al., 2021; Zhang et al., 2021a | 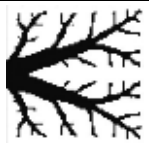 |
| | Fluid Structure | Gaymann & Montomoli, 2019; Deng et al., 2020 | 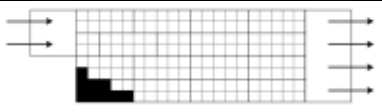 |
| ***3D*** | 3D Design Domain (Beam, Bracket, etc.) | Yildiz et al., 2003; Banga et al., 2018; Rawat & Shen, 2019a; Strömberg, 2019; Deng et al., 2020; Deng & To, 2020b; Greminger, 2020; Rade et al., 2020; Strömberg, 2020; Sun & Ma, 2020; Chandrasekhar & Suresh, 2021a; Chandrasekhar & Suresh, 2021b; Chandrasekhar & Suresh, 2021c; Chi et al., 2021; Keshavarzzadeh et al., 2021; Xue et al., 2021; Zehnder et al., 2021 | 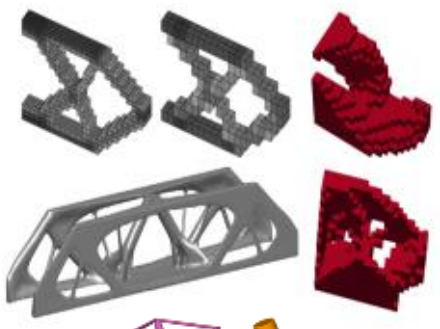 |
| | Metamaterial | White et al., 2019; Karlsson et al., 2020; Zheng et al., 2021 | 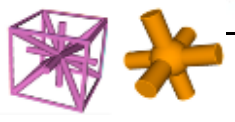 |
| | Robot Gripper | Ha et al., 2020 | 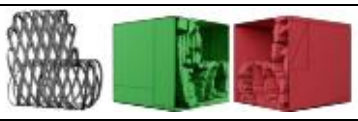 |

### 3.5.2. Generalization

The generalizability of the methodology can be classified into three possibilities.

- Can be applied to various load and boundary conditions
- Can be applied to 3D design problem
- Can be applied to any design domain

This section explored each topic in more detail.

*a) Can be applied to various load and boundary conditions*

The generalization for various load and boundary conditions has the advantage of being able to cover various structures. Studies applicable to various load and boundary conditions within the fixed design domain may be considered in this section.

*a1) Diversification of Load Conditions*

Qiu et al. (2021a), Abueidda et al. (2020) and Ulu et al. (2016) changed the position and direction of the load while the boundary conditions were fixed for 2D brackets, so allowed load condition generalization. Similarly, Sharpe and Seppersad (2019) generated optimal shapes by diversifying loading conditions for the 2D truss structure. Cang et al. (2019) also fed diversified loading conditions to a model to adaptively learns from true solutions from a distribution of problems. Thereafter, solutions to other unobservable problems from the same distribution could be predicted. Bujny et al. (2018) selected one of the three cases of boundary conditions and the load conditions at random locations within the design domain. Oh et al. (2018), Oh et al. (2019), and Jang et al. (2020) changed the direction and ratio of load conditions with fixed boundary conditions in the automobile wheel design. Li et al. (2019a) proposed a methodology to allow the heat sink position, the heat source position to be changed for the conductive heat transfer system. Qian and Ye (2020) diversified the load conditions for 2D bracket designs and used Poisson's ratio for meta-material design. Furthermore, Li et al. (2019c) and Li et al. (2021) reconstructed or generated periodic structures from latent space trained with optimal shapes at various load conditions, respectively, allowing for the use of the general load conditions.

*a2) Diversification of Load and Boundary Conditions*

Zhang et al. (2019), Nie et al. (2021), Yu et al. (2019), and Malviya (2020) utilized load and boundary conditions for design as inputs to model, enabling optimal shape prediction for various load and boundary conditions. Halle et al. (2021) also predicted the optimal shapes by inputting load and boundary conditions for the 2D bracket and repeated re-meshing and predicting until the objective function is converged. Hoyer et al. (2019) and Zhang et al. (2021a) updated the loss function containing load and boundary condition information for the 2D shape predicted by NN through a gradient based algorithm to enable generalization. Wang et al. (2021) trained a model that upscale low-resolution 2D bracket shapes to high resolution with diversified load and boundary conditions.

<u>*b) Application to 3D Design Problems*</u>

DLTO still lacks an expansion mechanism for the 3D design domain. However, as expansion to 3D is essential for future usability, studies with 3D applications were reviewed.

Rawat and Shen (2019a) optimized for 3D beams but did not consider various load and boundary conditions. Meanwhile, Banga et al. (2018), Rade et al. (2020), and Keshavarzzadeh et al. (2021) generalized for various load and boundary conditions and proceeded with extensions to 3D objects. Chi et al. (2021) updated the strain information about coarse mesh online and optimizes fine mesh. The study calculated strain for various load and boundary conditions and applied it to 3D. Kim et al. (2021) performed the optimization of structures considering materials in the problem of 2D and 3D structure optimization through FEA meta-modeling of periodic structure. Extensions to 3D are also applied in various areas. Greminger (2020) conducted a study of generating synthetic 3D voxel dataset. White et al. (2019) replaced the process of calculating effective elasticity of 3D metal with NN, and Ha et al. (2020) optimized 3D robot grippers for 3D optimized shapes.

<u>*c) Application to Any Design Domain*</u>

The most important point of expanding the application area is whether the design domain can be set flexibly beyond a) and b) in Section 3.5.2, which is the most important issue in the applicability of DLTO. Therefore, we explore the studies that enable the change in design domains. Hayashi and Ohsaki (2020) expressed the 2D truss structure with nodes and edges, allowing the size of the design domain to be adjusted. The RL agent learned generalizations for the design domain and could discover optimal truss topologies for various load cases. Brown et al. (2022) trained the RL agent to design optimal topologies (i.e., $6\times 6$ in size) with diverse load and boundary conditions. Then, they tested a multistep progressive refinement approach to prove that the RL agent could adopt a generalized design strategy.

Bi et al. (2020) applied the approach to general problems by updating the corresponding gradient online for a given 2D shape rather than using the trained model. Kumar and Suresh (2020) and Qiu et al. (2021b) optimized the 2D design domain by clustering microstructure with strain and stress information, respectively. The clustering process may be applied to various design domains.

While some studies have expanded their focus to various 2D design domains, other studies have expanded to 3D. Chandrasekhar and Suresh (2021c), Chandrasekhar and Suresh (2021a), Chandrasekhar and Suresh (2021b), Zehnder et al. (2021), and Deng and To (2020) used the network itself as an optimization calculation tool, thereby allowing the network to be used for any design domain and any load and boundary condition, and it can even be expanded to 3D. The DLTOP methodology (Kallioras et al., 2020) and other applied versions (Kallioras & Lagaros, 2020; Kallioras & Lagaros, 2021a; Kallioras & Lagaros, 2021b) are also applicable to any design domain and various load and boundary conditions when optimization progression is combined with SIMP and DL. Zheng et al. (2021) matched the stiffness tensors of 3D macroscale structures with the predicted stiffness tensors of microscale spinodoid meta-materials to optimize the spinodoid topologies of macrostructures with various load and boundary conditions. Xue et al. (2021) generalized the scheme to all different load and boundary conditions, and even 3D design domains, for the upscaling at high resolution.

# 4. Limitations and Research Opportunities

The previous sections have verified that studies are being conducted with various approaches to accelerate TO and improve the accuracy. However, limitations exist in relation to the usability and necessity of MLTO. In this section, the limitations commonly found in existing MLTO studies are summarized. On the basis, future research opportunities are examined.

## 4.1. Computational Complexity

As shown in the previous sections, the aim of MLTO is to reduce the computational cost of TO problems. For example, as shown in Table 4, MLTO can instantly infer the optimal topology once trained compared with the conventional TO.

### 4.1.1. Data Collection Burden

An important consideration in the MLTO field is the advantageous feature of MLTO, which is to focus on the short inference time after completing the training. As shown in Table 4, the data collection cost to complete the MLTO and the efficiency of front loading, such as the training time of the model, should also be considered. Many studies on supervised MLTO require a large amount of TO to collect optimized training data (Kallioras & Lagaros, 2020; Kallioras & Lagaros, 2021b; Rade et al., 2020; Sharpe & Seepersad, 2019; Wang et al., 2021; Wen et al., 2019; Wen et al., 2020; Yu et al., 2019). In particular, more data are required in 3D MLTO, but sufficient amounts of data are not easy to collect, and the initial computational cost for data generation is much larger. For instance, Zhang et al., 2019 generated a dataset of 80,000 with 88 line codes(Andreassen et al., 2011) to train a non-iterative MLTO neural network. The input of the 41× 81×6 tensor, which included the boundary conditions, loadings, initial displacement field, strain field, and volume fraction, was used. The output of a 40×80 matrix to represent the density of the elements was used for the neural network, replacing the classical compliance minimization problem solved by SIMP. Similarly, Yu et al. 2019 used 100,000 optimized structures to train non-iterative MLTO networks, with a solution domain of a 32×32 grid at low resolution and a 128×128 grid at high resolution. To solve a 3D TO problem in a supervised manner, Rade et al. 2020 used a 3D mesh cube as an initial design domain, which entailed 31,093 nodes and 154,677 elements. A total of 13,500 samples were generated by the iterative TO, which took an average of 13 iterations each time.

### 4.1.2. Computational Cost of FEA

Computational cost and accuracy generally have a tradeoff relationship. Although MLTO aims for acceleration, the accuracy is still inferior to that of the conventional iterative TO. However, studies attempted to use solvers, such as FEA in the intermediate process or the iterative TO in the middle part of the MLTO process, to increase the accuracy would also raise the problems of high computational cost and time consumption (Blanchard-Dionne & Martin, 2021; Doi et al., 2019; Gaymann & Montomoli, 2019; Sasaki & Igarashi, 2019b; Zhang et al., 2021a). For example, Chandrasekhar and Suresh (2021c) demonstrated that their FEA consumed approximately 50% of the computational cost in the case of a 60×30 mesh size, and the percentage would likely increase with increasing mesh size. The design generated in this method guaranteed engineering performance and increase the accuracy of the network, but it imposed a burden on the computational cost; this scenario implies a more significant disadvantage when used in large-scale problems, such as 3D design problems. Consequently, some studies have attempted to introduce methods such as parallel FEM computation (Deng et al., 2020), although more research is needed to utilize FEA while attempting to effectively minimize the computational cost.

### 4.1.3. Physics vs. Data

The computational complexity of MLTO shows the need for PINN that can learn the underlying physics instead of the use of a large pre-optimized dataset as the label. Similarly, recent studies have shown the possibility of solving the PDE through DL by capturing the underlying physics patterns (Anitescu et al., 2019; Samaniego et al., 2020). PINN is a method of reflecting additional information obtained by applying the laws of physics into network training and utilizing partial differential equations (PDEs) that describe physics problems into the loss functions (Raissi et al., 2019).

According to Zhang et al. (2019), by reflecting the physics information in the loss function, a structural disconnection of the generated optimization result may be prevented by generating the design based on engineering knowledge rather than simply predicting features based on the similarity between pixels. In addition, this method can train even with a small amount of training data, and it can reduce the burden and cost of data collection. In other words, an advantageous method may be provided for 3D TO. Consequently, scholars have planned to apply PINN to MLTO as a future work (Chandrasekhar & Suresh, 2021b; Ha et al., 2020; Keshavarzzadeh et al., 2021; Rade et al., 2020; Zhang et al., 2019). In addition, as it is not a method of simply predicting a shape based on the reference data, extrapolation may help to generate a new result that has not been seen before (Sim et al., 2021). However, PINN research showing the product-level output that can be used in the actual industry in MLTO is not yet shown, and it remains only a benchmark problem.

**Table 4 Computational cost of TO and MLTO**

| Research | Element Size | TO Time | Training | Data Generation | MLTO | Improvement (times faster) | Computational Resources |
|---|---|---|---|---|---|---|---|
| Almasri et al., 2020 | 101×101 | 0.3244 s | - | - | 0.0046 ms | 70,521× | GPU |
| Bi et al., 2020 | 480×160 | 600 s | - | - | ~ 150 s | 8.60× | ORNL OLCF Summit supercomputer |
| | 960×320 | 1400 s | - | - | ~ 200 s | - | |
| Cang et al., 2019 | - | 100~1000 s | - | - | 0.01 s | 10,000~ 100,000× | Intel Xeon CPU E5-1620 @ 3.50 GHz |
| Chi et al., 2021 | 85,750 | 1,346 s | - | - | 924 s | 1.46× | two Titan Xp GPUs |
| | 250,000 | 3,665 s | - | - | 1,515 s | 2.42× | |
| | 1,458,000 | 25,455 s | - | - | 6,069 s | 4.19× | |
| Halle et al., 2021 | 64×64 | 1.9 s | - | - | 7.3 ms | 259× | Nvidia Titan RTX |
| Hayashi & Ohsaki, 2020 | 3×2 | 1.6 s | - | - | 0.7 s | 2.29× | Intel® Core(TM) i9-7900X @ 3.30GHz |
| | 10×10 | 137.7 s | - | - | 7 s | 19.67× | |
| Jang et al., 2022 | 128×128 | 600 s | - | - | 0.01 s | 60,000× | NVidia GeForce 1080 Ti GPU |
| Kallioras et al., 2020 | 83,200 | 54,287 s | - | - | 20,910 s | 2.60× | Intel Xeon E5-1620 @ 3.70 GHz quad-core (with eight threads) with 16 GB RAM & NVIDIA GeForce 640 with 384 cores and 2 GB RAM |
| Keshavarzzadeh et al., 2021 | 128×128 | 2 s | 1,639.3 s | 20,203.8 s | 0.002 s | 1,000× | Intel 12 Core i7−5930K @ 3.5 GHz CPU with 32 GB RAM & 2x Intel 8 Core E5−2660 @ 2.20 GHz processor with 64 GB of RAM & 2x Nvidia K20 GPUs |
| | 40×20×10 | 33 s | 8,836.4 s | 331,213.1 s | 0.002 s | 16,500× | |
| Li et al., 2019a | 160×160 | 23 s | 14,400 s | 18 s | 0.006 s | 3,833.33× | Intel i7-8750H CPU with 8 GB RAM & NVidia GeForce GTX 1060 GPU |
| Lin et al., 2018 | 80×80 | 64.635 s | - | - | 18.297 s | 3.5× | - |
| Napier et al., 2020 | 160×100 | 591.7 s | - | 259,200 s | 22.6 s | 26.18× | Intel i7-5930k CPU at 3.50 GHz and 16GB of ram |
| | 320×120 | 3,912.62 s | - | - | 554.9 s | 7.3× | |
| Nie et al., 2021 | 64×128 | 119.83 s | - | - | 0.036 s | 3,328.61× | TO : Intel i7-6500U central processing unit (software ToPy)<br>MLTO : NVIDIA GeForce GTX 2080Ti GPU |
| Qian & Ye, 2021 | 36×36 | 0.42 s | 917.4 s | 973.8 s | 0.0127 s | 33.1× | Intel(R) Xeon(R) CPU E5-2687W v2 (3.40 GHz) |
| | 64×64 | 0.97 s | 4,545.6 s | 1,897.80 | 0.0101 s | 96.04× | - |
| Qiu et al., 2021a | 20×60 | 486 s | 1,710 s | 7,560 s | 0.6 s | 810× | Intel(R) Core(TM) i9-9900 3.10 GHz processors |
| | 20×60×10 | 2,250 s | 2,946 s | 33,840 s | 2.9 s | 775.86× | - |
| Rawat & Shen, 2019b | - | ~114.635 s | - | - | ~0.353 s | 324.75× | - |
| Sosnovik & Oseledets, 2019 | 40×40 | - | 4,800 ~ 5,400 s | - | - | 20× | NVIDIA Tesla K80 |
| Ulu et al., 2016 | 20×20 | - | ~ 625.19 s | - | ~ 0.0004 s | - | 2.4 GHz Core CPU and 8 GB RAM |
| Wang et al., 2021 | 128×128 | 24.13 s | - | - | 0.19 s | 127× | Intel CoreTM i5-7500 processor |
| Yu et al., 2019 | 128×128 | 22.7 s | - | 2377900 s | 0.014 s | 1,621.4× | Intel Xeon (8 cores and 2.5 GHz CPU clock) & NVIDIA Tesla P100 (3584 CUDA cores of 1.19 GHz GPU clock and 16 GB memory) |
| Zhang et al., 2019 | 40×80 | 3.913 s | - | - | 0.001 s | 3,913× | Intel(R)Core(TM) i7-8700k CPU<br>NVIDIA GeForce GTX1080 Ti GPU |
| Zheng et al., 2021 | - | 300 s | 600 s | - | 0.001 s | 300,000× | two 18-core 2.7 GHz Intel Xeon Gold 6150 processors and 192 GB of DDR4 memory at 2666 MHz |

## 4.2. Scalability

One of the most important issues of MLTO studies is the scalability of applicability to actual product development and various domains. With a single methodology, we should consider whether it can be extended to various TO or real-world problems and not simply to one design problem or condition. However, as reviewed in Section 3.5.2, most studies show limitations that these extensions and applications are impossible. The problem of the computational complexity of MLTO is linked to the scalability of MLTO. If MLTO does not have enough scalability, then it raises the question of whether MLTO is effective enough to risk the computational cost of data collection willingly. Therefore, the problem definition and usage scenario should be studied to emphasize and prove the need for MLTO to overcome the aforementioned shortcomings.

### 4.2.1. 3D Design

Most of the current studies have a limitation in that they cannot be applied to 3D problems (Abueidda et al., 2020; Almasri et al., 2020; Chandrasekhar & Suresh, 2021a; Chandrasekhar & Suresh, 2021b; Halle et al., 2021; Jiang & Fan, 2019a; Kumar & Suresh, 2020; Liu et al., 2015; Napier et al., 2020; Nie et al., 2021; Oh et al., 2019; Qian & Ye, 2021; Qiu et al., 2021b; Sharpe & Seepersad, 2019; Ulu et al., 2016; Wang et al., 2020; Wang et al., 2021; Yildiz et al., 2003; Yu et al., 2019; Zhou & Saitou, 2017). In other words, the works mentioned above are not yet applicable in the actual product development stage (Bujny et al., 2018; Oh et al., 2019; Wang et al., 2021). In reality, the 3D design domain must be applied, especially in large-scale problem. Furthermore, the TO process is more time-consuming, and the computational cost is expensive, so the application of MLTO is an urgent research topic. However, as the ML-based 3D TO problem entails numerous design parameters, more training data are required, which causes the problem of computational cost attributable to data collection and simulation analysis. Therefore, MLTO can be used while supplementing this problem.

### 4.2.2. Boundary Conditions

Many studies are applicable to only one design problem (Asanuma et al., 2020; Ha et al., 2020; Hayashi & Ohsaki, 2020; Jang et al., 2022; Jiang & Fan, 2019a; Jiang & Fan, 2019b; Jiang et al., 2019; Wang et al., 2021; Wang et al., 2020; Wen et al., 2019; Zhang et al., 2019; Zheng et al., 2021; Zhou & Saitou, 2017), and in most cases, the initial design domain has to be fixed (Nie et al., 2021; Sharpe & Seepersad, 2019) and cannot be applied to new boundary conditions or loading conditions (Bujny et al., 2018; Halle et al., 2021; Hayashi & Ohsaki, 2020; Lew & Buehler, 2021; Malviya, 2020; Rawat & Shen, 2018; Zheng et al., 2021). Furthermore, additional studies on complex problems (Kallioras & Lagaros, 2021a; Li et al., 2019b; Li et al., 2021; Lynch et al., 2019), such as multi-objective problems or various objective functions (Bujny et al., 2018; Chandrasekhar & Suresh, 2021c; Chandrasekhar & Suresh, 2021b; Jiang & Fan, 2019a; Kumar & Suresh, 2020; Qiu et al., 2021b; Sun & Ma, 2020), robust TO (Chi et al., 2021), nonlinear TO (Halle et al., 2021; Qian & Ye, 2021), and MMTO (Li et al., 2019c; Napier et al., 2020; Kim et al., 2021), are needed.

### 4.2.3. Industrial Applicability

Scalability is a crucial factor directly related to the usability of MLTO in industrial applications. From an industrial perspective, an adaptive tool with excellent generalizability to solve various problems is urgently needed. Most MLTO methods require a large amount of pre-optimized data, such as those in the studies presented in Section 2.2. However, collecting sufficient data, especially those used in the industry, takes time. Furthermore, the computational complexity of supervised MLTO limits its use in the industry, as explained in Section 4.1. Considering that these studies were conducted on the basis of benchmark data, the computational complexity of supervised MLTO for industrial applications will be more complicated. Therefore, recent studies have suggested the use of adaptive MLTO that can reparameterize the density field by using a neural network as shown in Section 2.5.2. This method can be generalized to various design domains and boundary conditions and extended to 3D problems without the need to prepare data in advance. Even though the problem of computational cost still needs to be supplemented due to the FEA process, this method can be advantageous for the industry.

## 4.3. Connectivity, Feasibility, and Robustness

### 4.3.1. Topology Resolution and Connectivity

Compared with the optimization result through the conventional iterative TO, the result of MLTO has a lower resolution, and a separate refinement process is often required. In particular, because physics information is not properly reflected and the optimization results are predicted with the similarity between pixels, connectivity may not be guaranteed, resulting in TO results that the user cannot trust and cannot be manufactured (Banga et al., 2018; Sharpe & Seepersad, 2019; Zhang et al., 2019). In addition, the checkerboard pattern phenomenon of the TO result is a significant problem to be solved (Chandrasekhar & Suresh, 2021c), and some studies have shown that MLTO's results do not produce detailed features or have blurry results because detailed features cannot be learned during the training process (Gaymann & Montomoli, 2019; Guo et al., 2018; Halle et al., 2021; Keshavarzzadeh et al., 2021; Li et al., 2019a).

#### 4.3.2. Manufacturing Feasibility

For the optimized result to be used in actual product development, it is necessary to examine the manufacturing possibility. Given that AM processing is often impossible for many TO results, it is important to consider the manufacturing constraints in the optimization process. Recent TO studies that consider manufacturability are being conducted. For instance, Cheng et al. (2019) proposed a functional gradient lattice structure TO for AM. Liu et al. (2021) proposed a stress-constrained TO method for fused deposition modeling in AM. Liu et al. (2022) constructed a simultaneous product design workflow while considering the materials and the manufacturing process for AM. Kim et al. (2022) implemented computational homogenization to increase the computational efficiency and reliability of FEA results in microscale and multiscale TOs. Wu & Xiao (2022) incorporated self-supporting factors to generate manufacturable designs. However, manufacturing constraints are rarely considered in the MLTO field. Therefore, more studies are needed to predict manufacturable optimization results considering complex manufacturing constraints (Karlsson et al., 2020; Wang et al., 2020; Zhang et al., 2019), such as AM conditions (Almasri et al., 2020) and injection molding (Greminger, 2020).

#### 4.3.3. Robust Learning Performance

Whether the network used for MLTO has robust quality should be verified and improved. This issue is also linked to the cherry-picking problem, showing only one-time good results frequently occurring in ML/DL research. As some of the networks are unstable, additional research on the convergence of the network (Gillhofer et al., 2019; Li et al., 2021; Takahashi et al., 2019) is needed. Furthermore, scholars have attempted to additionally apply various DL models, such as state-of-the-art GAN models (Li et al., 2019a) and cGAN that inputs the desired target volume or manufacturing process as a condition (Greminger, 2020), to improve further the accuracy of the network (Kallioras & Lagaros, 2021a; Malviya, 2020; Jiang & Fan, 2019b; Wen et al., 2020). Other scholars have focused on generative design to solve the mode collapse problem, which lacks diversity (Rawat & Shen, 2019b). Furthermore, studies that use GAN for generative design lack an appropriate evaluation method, and they need to consider the evaluation metric for the generated designs (Rawat & Shen, 2019a; Rawat & Shen, 2019b).

# 5. Conclusion

Various studies on data-based TO have been recently conducted, and the aim is to effectively derive optimal topologies by applying various ML techniques to supplement the problems of conventional methodologies. Each study was conducted using various ML methods, such as supervised learning, unsupervised learning, and reinforcement learning, for the purpose of acceleration of iteration, non-iterative optimization, meta-modeling, dimensionality reduction of design space, improvement of the optimizer, generative design, and postprocessing to TO field. These approaches allowed the studies to speed up the optimization process and reduce the computational cost.

Although various studies in this field have been conducted, many studies still have problems, such as the scalability of applicability to various design domains (e.g., 3D domain), the cost burden of data collection, low resolution of the resulting topology, the infeasibility of manufacturing, weak performance of the ML method, and expensive FEA and computational cost. We expect effective ML-based TO studies applicable to 3D TO to be conducted in the future to obtain optimal results in a manner that can explain engineering phenomena based on physics information.

# Acknowledgements

This work was supported by the National Research Foundation of Korea grant (2018R1A5A7025409) and the Ministry of Science and ICT of Korea grant (No.2022-0-00969, No.2022-0-00986)

# Reference

Aage, N., Andreassen, E., & Lazarov, B. S. (2015). Topology optimization using PETSc: An easy-to-use, fully parallel, open source topology optimization framework. Structural and Multidisciplinary Optimization, 51(3), 565-572.

Abueidda, D. W., Koric, S., & Sobh, N. A. (2020). Topology optimization of 2D structures with nonlinearities using deep learning. Computers & Structures, 237, 106283.

Albawi, S., Mohammed, T. A., & Al-Zawi, S. (2017, August). Understanding of a convolutional neural network. In 2017 International Conference on Engineering and Technology (ICET) (pp. 1-6). Ieee.

Allaire, G., Jouve, F., & Toader, A. M. (2002). A level-set method for shape optimization. Comptes Rendus Mathematique, 334(12), 1125-1130.

Almasri, W., Bettebghor, D., Ababsa, F., & Danglade, F. (2020). Shape related constraints aware generation of Mechanical Designs through Deep Convolutional GAN. arXiv preprint arXiv:2010.11833.

Amir, O., Bendsøe, M. P., & Sigmund, O. (2009). Approximate reanalysis in topology optimization. International Journal for Numerical Methods in Engineering, 78(12), 1474-1491.

Amir, O., Stolpe, M., & Sigmund, O. (2010). Efficient use of iterative solvers in nested topology optimization. Structural and Multidisciplinary Optimization, 42, 55-72.

Amir, O., & Sigmund, O. (2011). On reducing computational effort in topology optimization: how far can we go?. Structural and Multidisciplinary Optimization, 44, 25-29.

Andreassen, E., Clausen, A., Schevenels, M., Lazarov, B. S., & Sigmund, O. (2011). Efficient topology optimization in MATLAB using 88 lines of code. Structural and Multidisciplinary Optimization, 43, 1-16.

Anitescu, C., Atroshchenko, E., Alajlan, N., & Rabczuk, T. (2019). Artificial neural network methods for the solution of second order boundary value problems. Computers, Materials and Continua, 59(1), 345-359.

Asanuma, J., Doi, S., & Igarashi, H. (2020). Transfer Learning Through Deep Learning: Application to Topology Optimization of Electric Motor. IEEE Transactions on Magnetics, 56(3), 1-4.

Aulig, N., & Olhofer, M. (2014, July). Topology optimization by predicting sensitivities based on local state features. In Proc. 5th Eur. Conf. Comput. Mech.(ECCM V) (pp. 3578-3589).

Banga, S., Gehani, H., Bhilare, S., Patel, S., & Kara, L. (2018). 3d topology optimization using convolutional neural networks. arXiv preprint arXiv:1808.07440.

Bellman, R. (1957). A Markovian decision process. Journal of mathematics and mechanics, 679-684.

Bendsøe, M. P., & Kikuchi, N. (1988). Generating optimal topologies in structural design using a homogenization method. Computer methods in applied mechanics and engineering, 71(2), 197-224.

Bendsøe, M. P. (1989). Optimal shape design as a material distribution problem. Structural optimization, 1(4), 193-202.

Bendsøe, M. P., & Sigmund, O. (1999). Material interpolation schemes in topology optimization. Archive of applied mechanics, 69(9), 635-654.

Bendsoe, M. P., & Sigmund, O. (2013). Topology optimization: theory, methods, and applications: Springer Science & Business Media.

Bi, S., Zhang, J., & Zhang, G. (2020). Scalable deep-learning-accelerated topology optimization for additively manufactured materials. arXiv preprint arXiv:2011.14177.

Blanchard-Dionne, A. P., & Martin, O. J. (2021). Successive training of a generative adversarial network for the design of an optical cloak. OSA Continuum, 4(1), 87-95.

Borrvall, T., & Petersson, J. (2001). Large-scale topology optimization in 3D using parallel computing. Computer methods in applied mechanics and engineering, 190(46-47), 6201-6229.

Bourdin, B. (2001). Filters in topology optimization. International journal for numerical methods in engineering, 50(9), 2143-2158.

Breiman, L. (2001). Random forests. Machine learning, 45(1), 5-32.

Brown, N. K., Garland, A. P., Fadel, G. M., & Li, G. (2022). Deep reinforcement learning for engineering design through topology optimization of elementally discretized design domains. Materials & Design, 218, 110672.

Bruns, T. E., & Tortorelli, D. A. (2001). Topology optimization of non-linear elastic structures and compliant mechanisms. Computer methods in applied mechanics and engineering, 190(26-27), 3443-3459.

Bujny, M., Aulig, N., Olhofer, M., & Duddeck, F. (2018, July). Learning-based topology variation in evolutionary level set topology optimization. In Proceedings of the genetic and evolutionary computation conference (pp. 825-832).

Burges, C. J. (1998). A tutorial on support vector machines for pattern recognition. Data mining and knowledge discovery, 2(2), 121-167.

Cang, R., Yao, H., & Ren, Y. (2019). One-shot generation of near-optimal topology through theory-driven machine learning. Computer-Aided Design, 109, 12-21.

Cheng, L., Bai, J., & To, A. C. (2019). Functionally graded lattice structure topology optimization for the design of additive manufactured components with stress constraints. Computer Methods in Applied Mechanics and Engineering, 344, 334-359.

Chandrasekhar, A., & Suresh, K. (2021a). Multi-material topology optimization using neural networks. Computer-Aided Design, 136, 103017.

Chandrasekhar, A., & Suresh, K. (2021b). Length Scale Control in Topology Optimization using Fourier Enhanced Neural Networks. arXiv preprint arXiv:2109.01861.

Chandrasekhar, A., & Suresh, K. (2021c). TOuNN: topology optimization using neural networks. Structural and Multidisciplinary Optimization, 63(3), 1135-1149.

Chi, H., Zhang, Y., Tang, T. L. E., Mirabella, L., Dalloro, L., Song, L., & Paulino, G. H. (2021). Universal machine learning for topology optimization. Computer Methods in Applied Mechanics and Engineering, 375, 112739.

Chu, S., Xiao, M., Gao, L., Gui, L., & Li, H. (2016, May). An effective structural boundary processing method based on support vector machine for discrete topology optimization. In 2016 IEEE 20th International Conference on Computer Supported Cooperative Work in Design (CSCWD) (pp. 10-15). IEEE.

Cover, T., & Hart, P. (1967). Nearest neighbor pattern classification. IEEE transactions on information theory, 13(1), 21-27.

Deng, C., Qin, C., & Lu, W. (2020). Deep-Learning-Enabled Simulated Annealing for Topology Optimization. arXiv preprint arXiv:2002.01927.

Deng, H., & To, A. C. (2020). Topology optimization based on deep representation learning (DRL) for compliance and stress-constrained design. Computational Mechanics, 66, 449-469.

Doi, S., Sasaki, H., & Igarashi, H. (2019). Multi-objective topology optimization of rotating machines using deep learning. IEEE transactions on magnetics, 55(6), 1-5.

Ferrari, F., & Sigmund, O. (2020). Towards solving large-scale topology optimization problems with buckling constraints at the cost of linear analyses. Computer Methods in Applied Mechanics and Engineering, 363, 112911.

Gatys, L. A., Ecker, A. S., & Bethge, M. (2015). A neural algorithm of artistic style. arXiv preprint arXiv:1508.06576.

Gaymann, A., & Montomoli, F. (2019). Deep neural network and Monte Carlo tree search applied to fluid-structure topology optimization. Scientific reports, 9(1), 1-16.

Gillhofer, M., Ramsauer, H., Brandstetter, J., Schäfl, B., & Hochreiter, S. (2019). A GAN based solver of black-box inverse problems.

Goodfellow, I., Pouget-Abadie, J., Mirza, M., Xu, B., Warde-Farley, D., Ozair, S., ... & Bengio, Y. (2014). Generative adversarial nets. Advances in neural information processing systems, 27.

Goodfellow, I., Bengio, Y., & Courville, A. (2016). Deep learning. MIT press.

Greminger, M. (2020, August). Generative adversarial networks with synthetic training data for enforcing manufacturing constraints on topology optimization. In International Design Engineering Technical Conferences and Computers and Information in Engineering Conference (Vol. 84003, p. V11AT11A005). American Society of Mechanical Engineers.

Guest, J. K., Prévost, J. H., & Belytschko, T. (2004). Achieving minimum length scale in topology optimization using nodal design variables and projection functions. International journal for numerical methods in engineering, 61(2), 238-254.

Guo, X., Zhang, W., & Zhong, W. (2014). Doing topology optimization explicitly and geometrically—a new moving morphable components based framework. Journal of Applied Mechanics, 81(8).

Guo, T., Lohan, D. J., Cang, R., Ren, M. Y., & Allison, J. T. (2018). An indirect design representation for topology optimization using variational autoencoder and style transfer. In 2018 AIAA/ASCE/AHS/ASC Structures, Structural Dynamics, and Materials Conference (p. 0804).

Guo, H., Zhuang, X., & Rabczuk, T. (2021). A deep collocation method for the bending analysis of Kirchhoff plate. arXiv preprint arXiv:2102.02617.

Ha, H., Agrawal, S., & Song, S. (2020). Fit2form: 3d generative model for robot gripper form design. arXiv preprint arXiv:2011.06498.

Halle, A., Campanile, L. F., & Hasse, A. (2021). An Artificial Intelligence–Assisted Design Method for Topology Optimization without Pre-Optimized Training Data. Applied Sciences, 11(19), 9041.

Hayashi, K., & Ohsaki, M. (2020). Reinforcement learning and graph embedding for binary truss topology optimization under stress and displacement constraints. Frontiers in Built Environment, 6, 59.

Hoyer, S., Sohl-Dickstein, J., & Greydanus, S. (2019). Neural reparameterization improves structural optimization. arXiv preprint arXiv:1909.04240.

Huang, X., & Xie, Y. M. (2007). Convergent and mesh-independent solutions for the bi-directional evolutionary structural optimization method. Finite elements in analysis and design, 43(14), 1039-1049.

Jang, S., Yoo, S., & Kang, N. (2022). Generative design by reinforcement learning: enhancing the diversity of topology optimization designs. Computer-Aided Design, 146, 103225.

Jensen, P. D. L., Wang, F., Dimino, I., & Sigmund, O. (2021, August). Topology Optimization of Large-Scale 3D Morphing Wing Structures. In Actuators (Vol. 10, No. 9, p. 217). MDPI.

Jiang, J., & Fan, J. A. (2019a). Dataless training of generative models for the inverse design of metasurfaces. arXiv preprint arXiv:1906.07843, 401, 402.

Jiang, J., & Fan, J. A. (2019b). Global optimization of dielectric metasurfaces using a physics-driven neural network. Nano letters, 19(8), 5366-5372.

Jiang, J., Sell, D., Hoyer, S., Hickey, J., Yang, J., & Fan, J. A. (2019). Free-form diffractive metagrating design based on generative adversarial networks. ACS nano, 13(8), 8872-8878.

Jiang, X., Wang, H., Li, Y., & Mo, K. (2020). Machine Learning based parameter tuning strategy for MMC based topology optimization. Advances in Engineering Software, 149, 102841.

Jolliffe, I. T. (2002). Principal component analysis for special types of data (pp. 338-372). Springer New York.

Kallioras, N. A., & Lagaros, N. D. (2020). DzAIℕ: Deep learning based generative design. Procedia Manufacturing, 44, 591-598.

Kallioras, N. A., Kazakis, G., & Lagaros, N. D. (2020). Accelerated topology optimization by means of deep learning. Structural and Multidisciplinary Optimization, 62(3), 1185-1212.

Kallioras, N. A., & Lagaros, N. D. (2021a). DL-SCALE: A novel deep learning-based model order upscaling scheme for solving topology optimization problems. Neural Computing and Applications, 33(12), 7125-7144.

Kallioras, N. A., & Lagaros, N. D. (2021b). MLGen: Generative Design Framework Based on Machine Learning and Topology Optimization. Applied Sciences, 11(24), 12044.

Kaluarachchi, T., Reis, A., & Nanayakkara, S. (2021). A review of recent deep learning approaches in human-centered machine learning. Sensors, 21(7), 2514.

Karlsson, P., Pejryd, L., & Strömberg, N. (2020, September). Generative Design Optimization and Characterization of Triple Periodic Lattice Structures in AlSi10Mg. In International Conference on Additive Manufacturing in Products and Applications (pp. 3-16). Springer, Cham.

Keshavarzzadeh, V., Kirby, R. M., & Narayan, A. (2020). Stress-based topology optimization under uncertainty via simulation-based Gaussian process. Computer Methods in Applied Mechanics and Engineering, 365, 112992.

Keshavarzzadeh, V., Alirezaei, M., Tasdizen, T., & Kirby, R. M. (2021). Image-Based Multiresolution Topology Optimization Using Deep Disjunctive Normal Shape Model. Computer-Aided Design, 130, 102947.

Kim, C., Lee, J., & Yoo, J. (2021). Machine learning-combined topology optimization for functionary graded composite structure design. Computer Methods in Applied Mechanics and Engineering, 387, 114158.

Kim, J. E., Cho, N. K., & Park, K. (2022). Computational homogenization of additively manufactured lightweight structures with multiscale topology optimization. Journal of Computational Design and Engineering, 9(5), 1602-1615.

Kingma, D. P., & Welling, M. (2013). Auto-encoding variational bayes. arXiv preprint arXiv:1312.6114.

Kingma, D. P., & Welling, M. (2014, April). Stochastic gradient VB and the variational auto-encoder. In Second International Conference on Learning Representations, ICLR (Vol. 19, p. 121).

Kollmann, H. T., Abueidda, D. W., Koric, S., Guleryuz, E., & Sobh, N. A. (2020). Deep learning for topology optimization of 2D metamaterials. Materials & Design, 196, 109098.

Kudyshev, Z. A., Kildishev, A. V., Shalaev, V. M., & Boltasseva, A. (2021). Machine learning–assisted global optimization of photonic devices. Nanophotonics, 10(1), 371-383.

Kumar, T., & Suresh, K. (2020). A density-and-strain-based K-clustering approach to microstructural topology optimization. Structural and Multidisciplinary Optimization, 61(4), 1399-1415.

LeCun, Y., Bengio, Y., & Hinton, G. (2015). Deep learning. nature, 521(7553), 436-444.

Lee, S., Kim, H., Lieu, Q. X., & Lee, J. (2020). CNN-based image recognition for topology optimization. Knowledge-Based Systems, 198, 105887.

Lei, X., Liu, C., Du, Z., Zhang, W., & Guo, X. (2019). Machine learning-driven real-time topology optimization under moving morphable component-based framework. Journal of Applied Mechanics, 86(1), 011004.

Lew, A. J., & Buehler, M. J. (2021). Encoding and exploring latent design space of optimal material structures via a VAE-LSTM model. Forces in Mechanics, 5, 100054.

Li, B., Huang, C., Li, X., Zheng, S., & Hong, J. (2019a). Non-iterative structural topology optimization using deep learning. Computer-Aided Design, 115, 172-180.

Li, H., Kafka, O. L., Gao, J., Yu, C., Nie, Y., Zhang, L., ... & Liu, W. K. (2019b). Clustering discretization methods for generation of material performance databases in machine learning and design optimization. Computational Mechanics, 64(2), 281-305.

Li, M., Cheng, Z., Jia, G., & Shi, Z. (2019c). Dimension reduction and surrogate based topology optimization of periodic structures. Composite Structures, 229, 111385

Li, M., Jia, G., Cheng, Z., & Shi, Z. (2021). Generative adversarial network guided topology optimization of periodic structures via Subset Simulation. Composite Structures, 260, 113254.

Li, J., Gao, L., Ye, M., Li, H., & Li, L. (2022). Topology optimization of irregular flow domain by parametric level set method in unstructured mesh. Journal of Computational Design and Engineering, 9(1), 100-113.

Liang, Q. Q., Xie, Y. M., & Steven, G. P. (2000). Optimal topology selection of continuum structures with displacement constraints. Computers & Structures, 77(6), 635-644.

Limkilde, A., Evgrafov, A., & Gravesen, J. (2018). On reducing computational effort in topology optimization: we can go at least this far!. Structural and Multidisciplinary Optimization, 58, 2481-2492.

Lin, Q., Hong, J., Liu, Z., Li, B., & Wang, J. (2018). Investigation into the topology optimization for conductive heat transfer based on deep learning approach. International Communications in Heat and Mass Transfer, 97, 103-109.

Liu, K., & Tovar, A. (2014). An efficient 3D topology optimization code written in Matlab. Structural and Multidisciplinary Optimization, 50, 1175-1196.

Liu, K., Tovar, A., Nutwell, E., & Detwiler, D. (2015, August). Towards nonlinear multimaterial topology optimization using unsupervised machine learning and metamodel-based optimization. In International Design Engineering Technical Conferences and Computers and Information in Engineering Conference (Vol. 57083, p. V02BT03A004). American Society of Mechanical Engineers.

Liu, W., Wang, Z., Liu, X., Zeng, N., Liu, Y., & Alsaadi, F. E. (2017). A survey of deep neural network architectures and their applications. Neurocomputing, 234, 11-26.

Liu, J., Yan, J., & Yu, H. (2021). Stress-constrained topology optimization for material extrusion polymer additive manufacturing. Journal of Computational Design and Engineering, 8(3), 979-993.

Liu, G., Xiong, Y., & Rosen, D. W. (2022). Multidisciplinary design optimization in design for additive manufacturing. Journal of Computational Design and Engineering, 9(1), 128-143.

Lynch, M. E., Sarkar, S., & Maute, K. (2019). Machine learning to aid tuning of numerical parameters in topology optimization. Journal of Mechanical Design, 141(11)

MacQueen, J. (1967, June). Some methods for classification and analysis of multivariate observations. In Proceedings of the fifth Berkeley symposium on mathematical statistics and probability (Vol. 1, No. 14, pp. 281-297).

Malviya, M. (2020). A Systematic Study of Deep Generative Models for Rapid Topology Optimization.

Martínez-Frutos, J., Martínez-Castejón, P. J., & Herrero-Pérez, D. (2017). Efficient topology optimization using GPU computing with multilevel granularity. Advances in Engineering Software, 106, 47-62.

McDermott, M., Yan, T., Naumann, T., Hunt, N., Suresh, H., Szolovits, P., & Ghassemi, M. (2018, April). Semi-supervised biomedical translation with cycle wasserstein regression GANs. In Proceedings of the AAAI Conference on Artificial Intelligence (Vol. 32, No. 1).

Mirza, M., & Osindero, S. (2014). Conditional generative adversarial nets. arXiv preprint arXiv:1411.1784.

Mlejnek, H. P. (1992). Some aspects of the genesis of structures. Structural optimization, 5(1), 64-69.

Mnih, V., Kavukcuoglu, K., Silver, D., Graves, A., Antonoglou, I., Wierstra, D., & Riedmiller, M. (2013). Playing atari with deep reinforcement learning. arXiv preprint arXiv:1312.5602.

Mnih, V., Kavukcuoglu, K., Silver, D., Rusu, A. A., Veness, J., Bellemare, M. G., ... & Petersen, S. (2015). Human- level control through deep reinforcement learning. nature, 518(7540), 529-533.

Mnih, V., Badia, A. P., Mirza, M., Graves, A., Lillicrap, T., Harley, T., ... & Kavukcuoglu, K. (2016, June). Asynchronous methods for deep reinforcement learning. In International conference on machine learning (pp. 1928-1937). PMLR.

Murthy, S. K. (1998). Automatic construction of decision trees from data: A multi-disciplinary survey. Data mining and knowledge discovery, 2(4), 345-389.

Napier, N., Sriraman, S. A., Tran, H. T., & James, K. A. (2020). An artificial neural network approach for generating high-resolution designs from low-resolution input in topology optimization. Journal of mechanical design, 142(1).

Nie, Z., Lin, T., Jiang, H., & Kara, L. B. (2021). Topologygan: Topology optimization using generative adversarial networks based on physical fields over the initial domain. Journal of Mechanical Design, 143(3).

Oh, S., Jung, Y., Lee, I., & Kang, N. (2018, August). Design automation by integrating generative adversarial networks and topology optimization. In International Design Engineering Technical Conferences and Computers and Information in Engineering Conference (Vol. 51753, p. V02AT03A008). American Society of Mechanical Engineers.

Oh, S., Jung, Y., Kim, S., Lee, I., & Kang, N. (2019). Deep generative design: Integration of topology optimization and generative models. Journal of Mechanical Design, 141(11).

Park, S. M., Park, S., Park, J., Choi, M., Kim, L., & Noh, G. (2021). Design process of patient-specific osteosynthesis plates using topology optimization. Journal of Computational Design and Engineering, 8(5), 1257-1266.

Patel, J., & Choi, S. K. (2012). Classification approach for reliability-based topology optimization using probabilistic neural networks. Structural and Multidisciplinary Optimization, 45(4), 529-543.

Qian, C., & Ye, W. (2021). Accelerating gradient-based topology optimization design with dual-model artificial neural networks. Structural and Multidisciplinary Optimization, 63(4), 1687-1707.

Qiu, C., Du, S., & Yang, J. (2021a). A deep learning approach for efficient topology optimization based on the element removal strategy. Materials & Design, 212, 110179.

Qiu, Z., Li, Q., Liu, S., & Xu, R. (2021b). Clustering-based concurrent topology optimization with macrostructure, components, and materials. Structural and Multidisciplinary Optimization, 63(3), 1243-1263.

Rade, J., Balu, A., Herron, E., Pathak, J., Ranade, R., Sarkar, S., & Krishnamurthy, A. (2020). Physics-consistent deep learning for structural topology optimization. arXiv preprint arXiv:2012.05359.

Raissi, M., Perdikaris, P., & Karniadakis, G. E. (2019). Physics-informed neural networks: A deep learning framework for solving forward and inverse problems involving nonlinear partial differential equations. Journal of Computational physics, 378, 686-707.

Rawat, S., & Shen, M. H. (2018). A novel topology design approach using an integrated deep learning network architecture. arXiv preprint arXiv:1808.02334.

Rawat, S., & Shen, M. H. (2019a). Application of adversarial networks for 3d structural topology optimization (No. 2019-01-0829). SAE Technical Paper.

Rawat, S., & Shen, M. H. H. (2019b). A novel topology optimization approach using conditional deep learning. arXiv preprint arXiv:1901.04859.

Ringnér, M. (2008). What is principal component analysis?. Nature biotechnology, 26(3), 303-304.

Rozvany, G. I., Zhou, M., & Birker, T. (1992). Generalized shape optimization without homogenization. Structural optimization, 4(3-4), 250-252.

Samaniego, E., Anitescu, C., Goswami, S., Nguyen-Thanh, V. M., Guo, H., Hamdia, K., ... & Rabczuk, T. (2020). An energy approach to the solution of partial differential equations in computational mechanics via machine learning: Concepts, implementation and applications. Computer Methods in Applied Mechanics and Engineering, 362, 112790.

Sasaki, H., & Igarashi, H. (2019a). Topology optimization accelerated by deep learning. IEEE Transactions on Magnetics, 55(6), 1-5.

Sasaki, H., & Igarashi, H. (2019b). Topology optimization of IPM motor with aid of deep learning. International journal of applied electromagnetics and mechanics, 59(1), 87-96.

Schulman, J., Levine, S., Abbeel, P., Jordan, M., & Moritz, P. (2015, June). Trust region policy optimization. In International conference on machine learning (pp. 1889-1897). PMLR.

Schulman, J., Wolski, F., Dhariwal, P., Radford, A., & Klimov, O. (2017). Proximal policy optimization algorithms. arXiv preprint arXiv:1707.06347.

Sethian, J. A., & Wiegmann, A. (2000). Structural boundary design via level set and immersed interface methods. Journal of computational physics, 163(2), 489-528.

Settles, B. (2012). Active learning. Synthesis Lectures on Artificial Intelligence and Machine Learning, 6(1), 1–114.

Sharpe, C., & Seepersad, C. C. (2019, August). Topology design with conditional generative adversarial networks. In International Design Engineering Technical Conferences and Computers and Information in Engineering Conference (Vol. 59186, p. V02AT03A062). American Society of Mechanical Engineers.

Shen, M. H. H., & Chen, L. (2019). A new cgan technique for constrained topology design optimization. arXiv preprint arXiv:1901.07675.

Shin, S., Shin, D., Kim, M., Ryu, H., and Kang, N.* (2021) "Machine Learning-based Topology Optimization: A Review", The 2021 World Congress on Advances in Structural Engineering and Mechanics (ASEM21).

Sigmund, O. (1997). On the design of compliant mechanisms using topology optimization. Journal of Structural Mechanics, 25(4), 493-524.

Sigmund, O., & Torquato, S. (1997). Design of materials with extreme thermal expansion using a three-phase topology optimization method. Journal of the Mechanics and Physics of Solids, 45(6), 1037-1067.

Sigmund, O. (2001a). A 99 line topology optimization code written in Matlab. Structural and multidisciplinary optimization, 21(2), 120-127.

Sigmund, O. (2001b). Design of multiphysics actuators using topology optimization–Part II: Two-material structures. Computer methods in applied mechanics and engineering, 190(49-50), 6605-6627.

Sigmund, O. (2007). Morphology-based black and white filters for topology optimization. Structural and Multidisciplinary Optimization, 33(4-5), 401-424.

Sim, E. A., Lee, S., Oh, J., & Lee, J. (2021). GANs and DCGANs for generation of topology optimization validation curve through clustering analysis. Advances in Engineering Software, 152, 102957.

Sosnovik, I., & Oseledets, I. (2019). Neural networks for topology optimization. Russian Journal of Numerical Analysis and Mathematical Modelling, 34(4), 215-223.

Strömberg, N. (2019). A generative design optimization approach for additive manufacturing. In Sim-AM 2019: II International Conference on Simulation for Additive Manufacturing (pp. 130-141). CIMNE.

Strömberg, N. (2020). Efficient detailed design optimization of topology optimization concepts by using support vector machines and metamodels. Engineering Optimization, 52(7), 1136-1148.

Sutton, R. S., & Barto, A. G. (2018). Reinforcement learning: An introduction. MIT press.

Sun, H., & Ma, L. (2020). Generative design by using exploration approaches of reinforcement learning in density-based structural topology optimization. Designs, 4(2), 10.

Takahashi, Y., Suzuki, Y., & Todoroki, A. (2019). Convolutional neural network-based topology optimization (cnn-to) by estimating sensitivity of compliance from material distribution. arXiv preprint arXiv:2001.00635.

Ulu, E., Zhang, R., & Kara, L. B. (2016). A data-driven investigation and estimation of optimal topologies under variable loading configurations. Computer Methods in Biomechanics and Biomedical Engineering: Imaging & Visualization, 4(2), 61-72.

Van Engelen, J. E., & Hoos, H. H. (2020). A survey on semi-supervised learning. Machine learning, 109(2), 373-440.

Van Hasselt, H., Guez, A., & Silver, D. (2016, March). Deep reinforcement learning with double q-learning. In Proceedings of the AAAI conference on artificial intelligence (Vol. 30, No. 1).

Wang, C., Yao, S., Wang, Z., & Hu, J. (2021). Deep super-resolution neural network for structural topology optimization. Engineering Optimization, 53(12), 2108-2121.

Wang, L., Chan, Y. C., Ahmed, F., Liu, Z., Zhu, P., & Chen, W. (2020). Deep generative modeling for mechanistic-based learning and design of metamaterial systems. Computer Methods in Applied Mechanics and Engineering, 372, 113377.

Wang, M. Y., Wang, X., & Guo, D. (2003). A level set method for structural topology optimization. Computer methods in applied mechanics and engineering, 192(1-2), 227-246.

Watkins, C. J., & Dayan, P. (1992). Q-learning. Machine learning, 8, 279-292.

Wen, F., Jiang, J., & Fan, J. A. (2019). Progressive-growing of generative adversarial networks for metasurface optimization. arXiv preprint arXiv:1911.13029.

Wen, F., Jiang, J., & Fan, J. A. (2020). Robust freeform metasurface design based on progressively growing generative networks. ACS Photonics, 7(8), 2098-2104.

White, D. A., Arrighi, W. J., Kudo, J., & Watts, S. E. (2019). Multiscale topology optimization using neural network surrogate models. Computer Methods in Applied Mechanics and Engineering, 346, 1118-1135.

Wu, Z., Wang, S., Xiao, R., & Yu, L. (2020). A local solution approach for level-set based structural topology optimization in isogeometric analysis. Journal of Computational Design and Engineering, 7(4), 514-526.

Wu, Z., & Xiao, R. (2022). A topology optimization approach to structure design with self-supporting constraints in additive manufacturing. Journal of Computational Design and Engineering, 9(2), 364-379.

Xia, M., Yang, S., & Ho, S. L. (2017). A new topology optimization methodology based on constraint maximum-weight connected graph theorem. IEEE Transactions on Magnetics, 54(3), 1-4.

Xie, Y. M., & Steven, G. P. (1993). A simple evolutionary procedure for structural optimization. Computers & structures, 49(5), 885-896.

Xie, Y. M., & Steven, G. P. (1997). Basic evolutionary structural optimization. In Evolutionary structural optimization (pp. 12-29). Springer, London.

Xue, L., Liu, J., Wen, G., & Wang, H. (2021). Efficient, high-resolution topology optimization method based on convolutional neural networks. Frontiers of Mechanical Engineering, 16(1), 80-96.

Yago, D., Cante, J., Lloberas-Valls, O., & Oliver, J. (2022). Topology optimization methods for 3D structural problems: a comparative study. Archives of Computational Methods in Engineering, 29(3), 1525-1567.

Yamasaki, S., Yaji, K., & Fujita, K. (2021). Data-driven topology design using a deep generative model. Structural and Multidisciplinary Optimization, 64(3), 1401-1420.

Yang, X. Y., Xie, Y. M., Steven, G. P., & Querin, O. M. (1999). Bidirectional evolutionary method for stiffness optimization. AIAA journal, 37(11), 1483-1488.

Yildiz, A. R., Öztürk, N., Kaya, N., & Öztürk, F. (2003). Integrated optimal topology design and shape optimization using neural networks. Structural and Multidisciplinary Optimization, 25(4), 251-260.

Yu, Y., Hur, T., Jung, J., & Jang, I. G. (2019). Deep learning for determining a near-optimal topological design without any iteration. Structural and Multidisciplinary Optimization, 59(3), 787-799.

Zehnder, J., Li, Y., Coros, S., & Thomaszewski, B. (2021). Ntopo: Mesh-free topology optimization using implicit neural representations. Advances in Neural Information Processing Systems, 34, 10368-10381.

Zhang, Y., Peng, B., Zhou, X., Xiang, C., & Wang, D. (2019). A deep convolutional neural network for topology optimization with strong generalization ability. arXiv preprint arXiv:1901.07761.

Zhang, Z., Li, Y., Zhou, W., Chen, X., Yao, W., & Zhao, Y. (2021a). TONR: An exploration for a novel way combining neural network with topology optimization. Computer Methods in Applied Mechanics and Engineering, 386, 114083.

Zhang, C., Liu, J., Yuan, Z., Xu, S., Zou, B., Li, L., & Ma, Y. (2021b). A novel lattice structure topology optimization method with extreme anisotropic lattice properties. Journal of Computational Design and Engineering, 8(5), 1367-1390.

Zhao, J., Mathieu, M., & LeCun, Y. (2016). Energy-based generative adversarial network. arXiv preprint arXiv:1609.03126.

Zheng, L., Kumar, S., & Kochmann, D. M. (2021). Data-driven topology optimization of spinodoid metamaterials with seamlessly tunable anisotropy. Computer Methods in Applied Mechanics and Engineering, 383, 113894.

Zhou, Y., & Saitou, K. (2017). Topology optimization of composite structures with data-driven resin filling time manufacturing constraint. Structural and Multidisciplinary Optimization, 55(6), 2073-2086

Zhuang, X., Guo, H., Alajlan, N., Zhu, H., & Rabczuk, T. (2021a). Deep autoencoder based energy method for the bending, vibration, and buckling analysis of Kirchhoff plates with transfer learning. European Journal of Mechanics-A/Solids, 87, 104225.

Zhuang, C., Xiong, Z., & Ding, H. (2021b). Temperature-constrained topology optimization of nonlinear heat conduction problems. Journal of Computational Design and Engineering, 8(4), 1059-1081.

# Appendix A. Background of TO

Appendix A introduces the representative TO methodologies that are widely used in ML for TO studies as a foundation.

A typical TO problem is defined as follows. Find the material distribution that minimizes the objective function F, subject to a volume constraint $G_0 \leq 0$ and possibly M other constraints $G_i \leq 0$, i = 1 ...M. The material distribution is described by the density variable ρ(x) that can take either 0 (void) or 1 (solid material) at any point in the design domain $\Omega$ (Sigmund & Torquato, 1997; Bendsøe & Sigmund, 1999; Sigmund, 2001a).

TO can be expressed as follows:

$$\begin{aligned} \min \quad & F = F(u(\rho),\rho) = \int_{\Omega} f(u(\rho),\rho)dV \\ \text{s.t.} \quad & G_0(\rho) = \int_{\Omega} \rho(x)dV - V_0 \leq 0 \\ & G_i(u(\rho),\rho) \leq 0, \qquad j = 1,\dots,M \\ & \rho(x) = 0 \ \ or \ \ 1, \forall x \in \Omega \end{aligned} \tag{1}$$

where the state field u satisfies a linear or nonlinear state equation.

This section introduces the density-based method, evolutionary structural optimization (ESO) method, level set method, and moving morphable component (MMC) method among various TO methods. Figure 24 shows the results of optimizing MBB half beam, a representative benchmark problem of TO, by using the four methods.

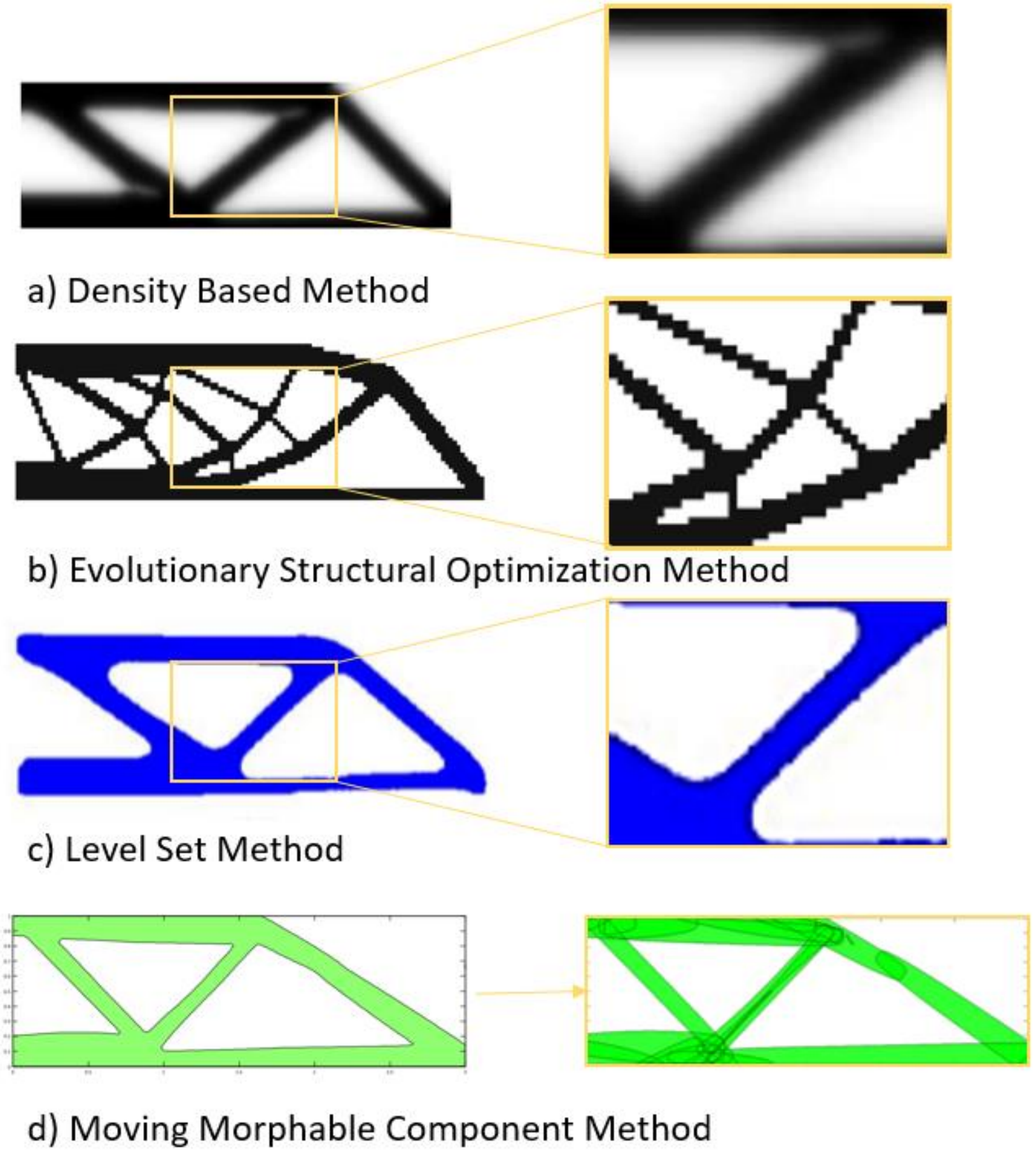

**Figure 24 Topology Optimization of MBB half beam using four different method**

### A.1. Density-Based Method

The SIMP method is the most representative density-based method, which obtains the optimal material distribution for specific load and boundary conditions in a given design space (Bendsøe & Kikuchi, 1988; Rozvany et al., 1992). Bendsøe (1989) mentioned that "shape optimization in its most general setting should consist of a determination for every point in space whether there is material in that point or not." In the SIMP method, the relative density of each element has a density value that varies between the minimum value $x_{min}$ and 1 rather than a value of 0,1 binary ($0 \leq x_{min} \leq x_e \leq 1$). $x_{min}$ is greater than 0, but it is the minimum allowable relative density of the empty element, which is to prevent the singularity of the FE matrices. It can also solve the problem in which an element with zero density cannot be re-densed.

According to Bendsoe and Sigmund (2013) and Sigmund (2001b), the basic foliation of SIMP for minimizing compliance can be expressed as follows:

$$\begin{aligned} \min \quad & c(x) = U^T K U = \sum_{e=1}^{N_e} u_e^T (E_e(x_e) k_0) u_e \\ \text{s.t.} \quad & \frac{V(x)}{V_0} = f \qquad KU = F \qquad 0 \leq x_e \leq 1, e = 1, \dots, N_e, \end{aligned} \tag{2}$$

where U is a displacement vector; K is a global stiffness matrix; c (x) is the compliance; $u_e$ is an element displacement vector; $k_0$ is an element stiffness matrix; f is the volume fraction; $N_e$ is the number of elements; $x_e$ is the design variable (i.e., density) of element e; and V (x) and $V_0$ are the material volume and the volume of design domain, respectively.

As the relative density of the material is continuous, and accordingly, the Young's modulus of the material may change continuously.

The modulus of elasticity $E(x_e)$ for each element is calculated by Power Law, reflecting the concept of penalty factor P for Young's modulus of solid material $E_0$ and the relative density of the material $x_e$. $E_{min}$ is the stiffness of soft (void) material (non-zero to avoid singularity of the stiffness matrix). E0 is Young's modulus of solid material.

The relationship between the density of elements in Modified SIMP and Young's modulus is expressed as follows (Sigmund, 2007):

$$E(x_e) = E_{min} + x_e^P (E_0 - E_{min}) \tag{3}$$

Penalty factor P reduces the effect of gray elements (elements with medium density) on total stiffness (i.e., numerical experiments indicate that a penalty factor value of p = 3 is suitable.).

In order for the optimizer to find the correct search direction in gradient-based optimization, sensitivity analysis of the objective function and constraints for the design variable is required in each iteration. This optimization algorithm for maximizing stiffness performs sensitivity analysis to determine the effect of element density changes on the objective function and constraints. Therefore, sensitivity analysis can be expressed as a differential of the objective function and the constraint for the element density under the setting that all elements have unit volume.

$$\frac{\partial C}{d\partial} = -p(x_e)^{p-1}[u_e]^T [K_e][u_e] \tag{4}$$

$$\frac{\partial V}{\partial x_e} = \frac{\partial}{\partial x_e}\left(\sum_{e=1}^{N_e} x_e v_e\right) = 1 \tag{5}$$

The OC method is a method commonly used to update design variables in TO problems, and is most commonly applied to the density-based method. The method of updating design variables through OC is given by

$$x_e^{new} = \begin{cases} \max(0, x_e - m) & \text{if } x_e B_e^{\eta} \leq \max(0, x_e - m) \\ \min(1, x_e + m) & \text{if } x_e B_e^{\eta} \geq \min(1, x_e + m) \\ x_e B_e^{\eta} & \text{otherwise} \end{cases} \tag{6}$$

where m denotes a positive move-limit, η denotes a numerical damping coefficient, $B_e$ follows the formula

$$B_e = \frac{-(\partial c / \partial x_e)}{\lambda(\partial V / \partial x_e)} \tag{7}$$

In addressing the checkerboard patterns and grayscale issues, filtering methods, such as density filter and sensitivity filter, are presented, which allow the optimal design to be derived by considering the connection between elements instead of performing sensitivity analysis independently for each element.

The basic filters applied to TO are the sensitivity filter and the density filter, which consider the connection between elements through the method of weighted average with neighboring elements of sensitivity or element density, respectively. The neighborhood element is determined based on the distance from the center of the reference element, and the maximum distance considered as a neighbor is a parameter set by the user as mesh-independent radius. On this basis, the convolutional operator is defined as follows:

$$H_f = r_{min} - \mathrm{dist}(\mathrm{e}, f) \tag{8}$$

where $r_{min}$ means mesh-independent radius $\mathrm{dist}(\mathrm{e}, f)$ is the distance between elements and $f$ is the element whose distance from the center of e is less than $r_{min}$. The sensitivity filter defined using this convolution operator is expressed as follows (Sigmund & Torquato, 1997):

$$\widehat{\frac{\partial c}{\partial x_e}} = \frac{1}{x_e \sum_{f=1}^{N} H_f} \sum_{f=1}^{N} H_f x_f \frac{\partial c}{\partial x_f} \tag{9}$$

In the case of a density filter, the weighted average concept is the same as that of a sensitivity filter, but there is a difference that physical density is used instead of sensitivity, and it is defined as follows (Bruns & Tortorelli, 2001; Bourdin, 2001):

$$\tilde{x}_e = \frac{\sum_{f=1}^{N} H_f x_f}{\sum_{f=1}^{N} H_f} \tag{10}$$

Accordingly, the original density corresponds to the design variable, the filtered density corresponds to the physical density, and the sensitivity analysis to the design variable is modified for the physical density (Guest et al., 2004).

### A.2. Evolutionary Structural Optimization Method

ESO methods were first introduced in Xie and Steven (1993) and Xie and Steven (1997). The methodology proceeds by removing unnecessary materials from the structure to find the optimal structure. BESO, an advanced method of the ESO, was introduced in Yang et al. (1999), which removes unnecessary materials while adding necessary materials. The optimization problem that minimizes compliance with the most basic ESO is defined as follows:

$$\begin{aligned} \min \quad & c(x) = U^T K U \\ \text{s.t.} \quad & \frac{V(x)}{V_0} = f \qquad KU = F \qquad x_e = \\ & [\ 0,1]\ , e = 1, \dots, N_e, \end{aligned} \tag{11}$$

At this time, the criteria for removing elements are based on the sensitivity number, and the sensitivity number can be obtained through the following equation (Liang et al., 2000).

$$\alpha_i^e = \frac{1}{2} u_i^T K_i u_i \tag{12}$$

Through this, the sensitivity number is calculated to remove elements with elements lower than the criteria.

BESO is a method of removing elements as the process of ESO while adding elements to the required area. Huang and Xie (2007) introduced a complementary BESO method that solves the mesh dependency problem without converging. The nodal sensitivity number is used to solve the checkerboard pattern which is the average of the element sensitivity connected to the j-th node and can be obtained as follows:

$$\alpha_j^n = \sum_{i=1}^{M} \omega_i \alpha_i^e \tag{13}$$

where $\omega_i$ denotes the weight value of the i-th element, and is defined as follows:

$$\omega_i = \frac{1}{M-1}\left(1 - \frac{r_{ij}}{\sum_{i=1}^{M} r_{ij}}\right) \tag{14}$$

where $r_{ij}$ is the distance from the center of the i-th element to the j-th node.

Then, the nodes affecting the sensitivity of the i-th element may be found through $r_{min}$. The nodes in this domain contribute to the calculation of the i-th element's improved sensitivity number, which is calculated as follows:

$$\alpha_i = \frac{\sum_{j=1}^{K} \omega(r_{ij})\alpha_j^n}{\sum_{j=1}^{K} \omega(r_{ij})} \tag{15}$$

where K is the total number of nodes in the filter domain of $r_{min}$, and $\omega(r_{ij})$ is defined as a linear weight factor as follows:

$$\omega(r_{ij}) = r_{min} - r_{ij} \qquad (j = 1,2,\dots,K) \tag{16}$$

In addition to the solid element, the sensitivity number of the void element is calculated, and elements with a sensitivity number less than $\alpha_{del}^{th}$ among the solid elements are removed, and elements with a sensitivity number greater than $\alpha_{add}^{th}$ among the void elements are added.

### A.3. Level Set Method

The level set method is used to capture the free boundaries of a structure in a linear elastic structure, resulting in a crisp design. Therefore, the results require less postprocessing. This method is fundamentally different from shape optimization because it not only shifts structural boundaries but also includes formation, extinction, and merging of void regions that define phase designs (Sethian & Wiegmann, 2000).

In the level set method, the boundary is expressed as a zero-level curve of the scalar function $\Phi$. The movements and merging of boundaries are operated by this scalar function and regulated by physical problems and optimization conditions. The level set method implicitly expresses the boundaries and allows a smooth boundary to be expressed in a three-dimensional space from a zero function value.

$$S = \{x : \phi(x, 0) = k\} \tag{17}$$

where $k$ is an arbitrary value containing a zero function value and $x$ is an arbitrary coordinate on the curved surface $\phi$. As the structural optimization changes over time, Eq. (17) can be expressed as level set function dynamically changing as follows:

$$S(t) = \{x(t) : \Phi(x(t), t) = k\} \tag{18}$$

Differentiating this expression with time and applying the chain rule defines the "Hamilton–Jacobi-type" equation that defines the initial value for the time-dependent function $\phi$.

$$\frac{\partial\Phi(x,t)}{\partial t} + \nabla\Phi(x,t)\frac{dx}{dt} = 0, \qquad \Phi(x, t = 0) = \Phi_0(x) \tag{19}$$

In solving this equation, let $\frac{dx}{dt}$ be the movement of a point on a surface driven by the objective of the optimization. Then, as a solution of a PDE on $\phi$, the optimal structural boundary can be expressed as follows (Wang et al., 2003):

$$\frac{\partial\Phi(x,t)}{\partial t} \equiv -\nabla\Phi(x)\Gamma(x,\Phi) \tag{20}$$

where $\Gamma(x,\Phi)$ denotes the "speed vector" of the level set and it depends on the objective of the optimization. Proper vector obtains an appropriate value from the descent direction of the objective through sensitivity analysis.

$$
\begin{aligned}
\text{min} \quad & C(u,\Phi) = \int_\Omega E\varepsilon(u)\varepsilon(u)H(\Phi)d\Omega \\
\text{s.t.} \quad & \int_\Omega H(\Omega)d\Omega \le V_f \\
& \int_\Omega E\varepsilon(u)\varepsilon(v)H(\Phi)d\Omega = \int_\Omega bvH(\Phi)d\Omega + \int_{\Gamma_s} fv\Gamma_s u|_{\Gamma_D} = 0\forall_v \in U
\end{aligned} \tag{21}
$$

**A.4. MMC Method**

In contrast to the conventional pixel-based and node point-based methods, MMC method can organize an optimal structure by determining shape characteristic parameters, such as shape, length, thickness, and direction, and the layout of these components (Guo et al., 2014).

The MMC method can incorporate numerous geometric information for defined problems, and express quite complex and diverse structure topologies with a small number of components. In the MMC method, components can overlap each other, and the overlapping regions on the other components do not have a structural effect.

The area in which the component exists may be represented by a level set function.

$$
\{\emptyset(x) > 0, if \;\; x \in \Omega \;\; \emptyset(x) = 0, if \;\; x \in \partial\Omega \;\; \emptyset(x) < 0, if \;\; x \in D\backslash \Omega \;\; , \tag{22}
$$

where

$$
\emptyset(x,y) = \left(\frac{cos\theta(x - x_0) + sin\theta(y - y_0)}{L/2}\right)^m + \left(\frac{-sin\theta(x - x_0) + cos\theta(y - y_0)}{t/2}\right)^m - 1 \tag{23}
$$

where m is a relatively large even integer number (we take m=6 in the present study). The structural component can move, dilate/shrink and rotate in the design domain by changing the values of x0, y0, L, t and $\theta$.

On the basis of the discussion above, the MMC-based TO problem is defined as follows:

$$
\begin{aligned}
\text{Find} \quad & d = (d_1, \dots, d_{nc})^T \\
\text{min} \quad & I = I(d) \\
\text{s.t.} \quad & g_i(d) \le 0, i = 1, \dots, l \;,\; d \subset \; U_d
\end{aligned} \tag{24}
$$

where $nc$ represents the total number of components involved in the optimization problem, $d = (d_1, \dots d_{nc})^T$ is the vector of design variables with $d_i = (x_{0i,} y_{0i}, L_i, t_i, \theta_i)^T$, $i = 1, \dots nc$. $U_d$ is the admissible sets that $d$ belongs to. Moreover, $g_i, i = 1, \dots, l$ are the considered constraint functionals.

# Appendix B. Background of ML

DL models can be categorized into four types: supervised learning, unsupervised learning, semi-supervised learning, and reinforcement learning. Supervised learning is a method of training by using labeled data. In the case of unsupervised learning, unlabeled data are used for training and training is done by finding undefined features or patterns from input data. Semi-supervised learning is a method of using a mixture of supervised and unsupervised learning methods, including training a larger unlabeled dataset by using small amounts of labeled data or by utilizing a proper combination of supervised and unsupervised learning models. Reinforcement learning is a different concept from supervised/unsupervised learning, which does not require data or labels separately, receives a reward for its action on the agent while the training progresses, and the model is trained to build a strategy for action in the direction of increasing this reward.

### B.1. Supervised Learning

The biggest characteristic of supervised learning is that labels exist together with the input. Based on the labeled dataset, predictions are initiated for new data. Supervised learning is largely divided into classification and regression. Classification includes binary classification, which classifies data into one of two categories, and multiclass classification, which classifies data into one of several classes. Regression is used to predict continuous values based on the features of the dataset. The representative algorithms for supervised learning include KNN, SVM, decision tree, random forest, and NN.

The KNN is a method used to perform classification, which is based on the principle that instances within a dataset typically exist close to other instances with similar properties. (Cover & Hart, 1967). SVM is a method used for classification and regression to maximize the margin of the data class and thereby create the largest distance between the separated hyperplane and the instances on both sides (Burges, 1998). A decision tree is a tree that sorts and classifies instances based on the feature, and each node in the decision tree represents the feature of the instance to be classified, and each branch represents the value that the node can predict. Instances are classified and sorted by corresponding features, starting at the root node (Murthy, 1998). Random forest is an ensemble learning method used for classification and regression, formed by a combination of several decision trees (Breiman, 2001). ANN is a concept that mathematically models the structure of human brain neurons and has since developed into a DNN that deepens the hidden layer of ANN (Liu et al., 2017). Among DNNs, the most representative CNN consists of a convolutional layer, nonlinearity layer, pooling layer, and fully connected layer, and shows good performance in image data application problems (Albawi et al., 2017).

### B.2. Unsupervised Learning

Unsupervised learning refers to a training method in which a model receives input but has no label for it. Therefore, the model clusters according to the similarity based on the input data and learns the characteristics and patterns of the data by itself.

Unsupervised learning is commonly used for clustering, dimensionality reduction, and generative models. Clustering is a method of grouping data with similar features into the same cluster. As an example, K-means is the simplest and most popular algorithm for iterative and hill climbing clustering algorithms. The K-means is an algorithm that groups given data into k clusters, which are clustered based on the center of each cluster and the average distance of the data. K-means were originally designed to find the optimal partition for dividing data, but k-means generally do not converge to the optimal partition (MacQueen, 1967).

Dimensionality reduction is the process of extracting high-dimensional data into low-dimensions, which still compresses meaningful features of the data. For example, PCA is a mathematical algorithm that reduces the dimension of data while maintaining the variations of most of the data, and the dimension is reduced by finding a basis that maximizes the variance (Jolliffe, 2002). Dimensionality reduction allows samples to be represented in relatively low dimension and these reduced samples with similar features are most likely to be gathered together in the latent space (Ringnér, 2008).

GAN, the most common generative model, consists of two NNs, a generator and a discriminator. The generator generates new data based on existing data as the reference, and the discriminator aims to distinguish between the real data and the data generated by the generator, allowing the generator to generate data similar to the actual data (Goodfellow et al., 2014). By expanding this, cGAN uses additional information as a condition for the generator and the discriminator (Mirza & Osindero, 2014). VAE is also a generative model that generates new data similar to the existing data from a latent space, which contains features of the existing data (Kingma & Welling, 2014). Therefore, for the purpose of training the decoder to be used as a generative model, training is performed by attaching an encoder.

**B.3. RL**

RL is a type of ML that makes decisions through the interaction of agents and environments (Sutton & Barto, 2018). RL is usually based on the Markov decision process, but it does not require explicit models.

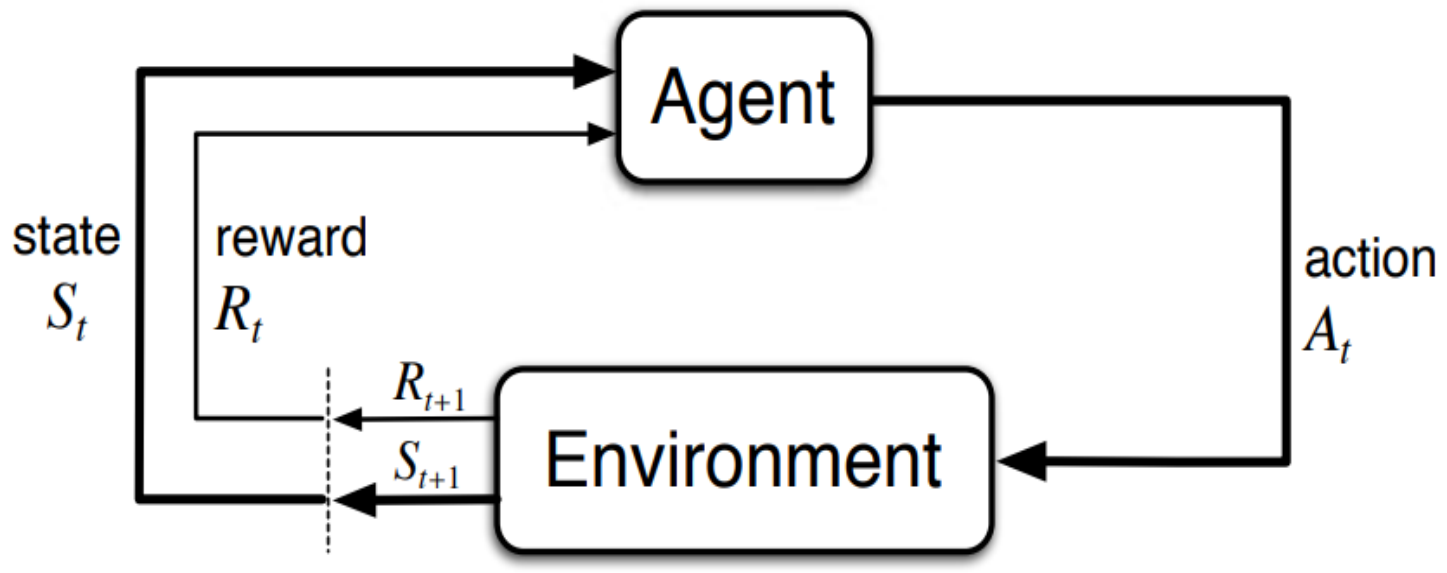

**Figure 25 RL Algorithm (Sutton & Barto, 2018)**

As shown in Figure 25, the agent takes an action $a_t$ in the current time t, which returns state $s_t$ and reward $R_t$ from the environment $\varepsilon$. The algorithm is trained through trial-and-error so that the agent determines the action in the direction of maximizing the total reward and returns $R_t = \sum_k \gamma^k \cdot R_{t+k}$ received by the agent in the state of the given environment (where $\gamma \in (0,1)$, $\gamma$ is a discount factor that reflects future rewards in the current state). The two main approaches to the RL framework are the value functions based approach and the policy search based approach. The value function method is based on the method of estimating the expected return value in a given state, and deep Q network (DQN) and double-DQN (DDQN) the representative method (Mnih et al., 2013; Mnih et al., 2015; Van Hasselt et al., 2016). The policy method finds the gradient of the policy and updates the parameters in the direction of increasing the reward. Advantage actor–critic and asynchronous advantage actor–critic are the most representative methods by using these policy updates (Mnih et al., 2016). Furthermore, as a hybrid actor critic method that applies the concept of both methods, Schulman et al. (2015) proposed trust region policy optimization, and Schulman et al. (2017) proposed PPO.